\documentclass[11pt]{article}

\usepackage[left=1in, right=1in, top=1in, bottom=1in]{geometry}
\usepackage{amsmath,amssymb,graphicx,amsthm,nicefrac,mathtools}
\usepackage[round, sort]{natbib}
\usepackage[english]{babel}
\usepackage[T1]{fontenc}
\usepackage{bbm}
\usepackage{color}
\usepackage{xspace}
\usepackage{graphics}
\usepackage{pifont}
\usepackage{float}
\usepackage{array}
\usepackage{booktabs}

\usepackage[utf8]{inputenc}
\usepackage[table,x11names,dvipsnames,table]{xcolor}
\usepackage{multirow,graphicx}
\usepackage{float}
\usepackage{nicefrac}
\usepackage{bm}
\usepackage{wrapfig}
\usepackage[colorlinks=true,linkcolor=blue,citecolor=blue,urlcolor=blue]{hyperref} 
\usepackage{morefloats} 

\usepackage{xcolor}
\definecolor{DarkGreen}{rgb}{0.1,0.5,0.1}
\definecolor{DarkRed}{rgb}{0.5,0.1,0.1}
\definecolor{DarkBlue}{rgb}{0.1,0.1,0.5}
\definecolor{Black}{rgb}{0.0,0.0,0.0}
\hypersetup{
    colorlinks=true,       
    linkcolor=DarkBlue,          
    citecolor=DarkBlue,        
    filecolor=DarkBlue,      
    urlcolor=DarkBlue,          
    pdftitle={Debugging Tests for Feature Attributions.},
    pdfauthor={Julius Adebayo, Michael Muelly, Ilaria Liccardi, Been Kim},
}
\usepackage{placeins} 

\usepackage[toc,page,header]{appendix} 
\usepackage{minitoc}

\definecolor{myred}{RGB}{215,48,39}
\definecolor{mygreen}{RGB}{26,152,80}
\definecolor{lightgray}{gray}{0.96}
\definecolor{blue}{RGB}{30, 144, 255}

\newcommand{\PreserveBackslash}[1]{\let\temp=\\#1\let\\=\temp}
\newcolumntype{C}[1]{>{\PreserveBackslash\centering}p{#1}}
\newcolumntype{R}[1]{>{\PreserveBackslash\raggedleft}p{#1}}
\newcolumntype{L}[1]{>{\PreserveBackslash\raggedright}p{#1}}

\DeclareMathOperator*{\argmin}{arg\,min}

\title{\textbf{Debugging Tests for Model Explanations}}
\author{Julius Adebayo \\ Massachusetts Institute of Technology \\ \texttt{juliusad@mit.edu}
\and 
Micheal Muelly \\ Stanford University \\ \texttt{mmuelly@gmail.com}\\
\and
Ilaria Liccardi \\ Massachusetts Institute of Technology \\ \texttt{liccardi@csail.mit.edu}
\and Been Kim \\ Google AI \\ \texttt{beenkim@google.com}}
\date{}

\begin{document}

\doparttoc 
\faketableofcontents 

\maketitle

\begin{abstract}
We investigate whether post-hoc model explanations are effective for diagnosing model errors--model debugging. In response to the challenge of explaining a model's prediction, a vast array of explanation methods have been proposed.  Despite increasing use, it is unclear if they are effective. To start, we categorize \textit{bugs}, based on their source, into: ~\textit{data, model, and test-time} contamination bugs. For several explanation methods, we assess their ability to: detect spurious correlation artifacts (data contamination), diagnose mislabeled training examples (data contamination), differentiate between a (partially) re-initialized model and a trained one (model contamination), and detect out-of-distribution inputs (test-time contamination). We find that the methods tested are able to diagnose a spurious background bug, but not conclusively identify mislabeled training examples. In addition, a class of methods, that modify the back-propagation algorithm are invariant to the higher layer parameters of a deep network; hence, ineffective for diagnosing model contamination. We complement our analysis with a human subject study, and find that subjects fail to identify defective models using attributions, but instead rely, primarily, on model predictions. Taken together, our results provide guidance for practitioners and researchers turning to explanations as tools for model debugging.
\end{abstract}

\section{Introduction}
Diagnosing and fixing model errors--model debugging--remains a longstanding machine learning challenge~\citep{cawsey1991generating, cawsey1993user, carenini1994generating, chakarov2016debugging, cadamuro2016debugging, zinkevichrules2020, sculley2015hidden}. Model debugging is increasingly important as automated systems, with learned components, are being tested in high-stakes settings~\citep{herlocker2000explaining, mckinney2020international, bhatt2019explainable} where inadvertent errors can have devastating consequences. Increasingly,~\textit{explanations}--artifacts derived from a trained model with the primary goal of providing insights to an end-user--are being used as debugging tools for models assisting healthcare providers in diagnosis across several specialties~\citep{sayres2019using,cai2019human,wen2017comparative}. Despite a vast array of explanation methods and increased use for debugging, little guidance exists on method effectiveness. For example, should an explanation work equally well for diagnosing mislabeled training samples and detecting spurious correlation artifacts? Should an explanation that is sensitive to model parameters also be effective for detecting domain shift? Consequently, we ask and address the following question:\\
\centerline{\textit{which explanation methods are effective for which classes of model bugs?}}

\begin{figure*}[ht]
\centering
\includegraphics[scale=0.45, page=1]{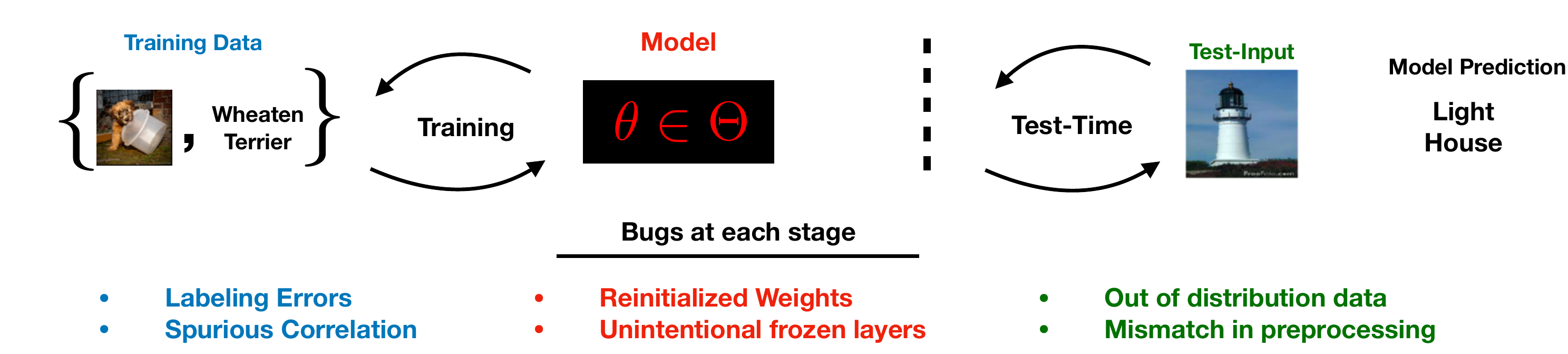}
\caption{\textbf{Debugging framework for the standard supervised learning pipeline.} Schematic of the standard supervised learning pipeline along with examples of bugs that can occur at each stage of the pipeline. The categorization captures defects that can occur with the training data, model, and at test-time. We term these: \textit{data}, \textit{model}, and \textit{test-time contamination tests}.}
\label{fig:framework}
\end{figure*}

To address this question, we make the following contributions:
\begin{enumerate}
    \item \textbf{Bug Categorization.} We categorize bugs, based on the source of the defect leading to the bug, in the supervised learning pipeline (see Figure~\ref{fig:framework}) into three classes: \textit{data, model, and test-time} contamination. These contamination classes capture defects in the training data, model specification and parameters, and with the input at test-time.
    
    \item \textbf{Empirical Assessment.} We conduct comprehensive control experiments to assess several feature attribution methods against $4$ bugs: `spurious correlation artifact', mislabelled training examples, re-initialized weights, and out-of-distribution (OOD) shift.
    
    \item \textbf{Insights.} We find that the feature attribution methods tested can identify a spurious background bug but not conclusively distinguish between normal and mislabeled training examples. In addition, attribution methods that derive relevance by modifying the back-propagation computation via `positive aggregation' (see Section~\ref{section:modelcontamination}) are invariant to the higher layer parameters of a deep neural network (DNN) model. Finally, we find that in specific settings, attributions for out-of-distribution examples are visually similar to attributions of these examples but with an `in-domain' model, suggesting that debugging solely based on visual inspection might be misleading.
    
    \item \textbf{Human Subject Study.} We conduct a $54$-person IRB-approved study to assess whether end-users can identify defective models with attributions. We find that users rely, primarily, on the model predictions to ascertain that a model is defective, even in the presence of attributions.
\end{enumerate}

\paragraph{Related Work} This work is in line with contributions that assess the effectiveness of post-hoc explanations; albeit with a focus on feature attributions and model debugging. Our bug categorization incorporates previous use of explanations for diagnosing spurious correlation~\citep{ribeiro2016should, meng2018automatic, kim2018interpretability}, domain mismatch, and mislabelled examples~\citep{koh2017understanding}. Correcting bugs can also be achieved by penalizing feature attributions during training~\citep{rieger2019interpretations, erion2019learning, ross2017right} or clustering~\citep{lapuschkin2019unmasking}.

The dominant evaluation approach involves input perturbation~\citep{samek2017evaluating, montavon2017methods}, which can be combined with retraining~\citep{hooker2018evaluating}. However,~\citet{tomsett2019sanity} showed that input perturbation produces inconsistent quality rankings.~\citet{meng2018automatic} propose manipulations to the training data along with a suite of metrics for assessing explanation quality. The data and model contamination categories recover the `sanity checks' of~\citet{adebayo2018sanity}. The finding that methods that modify backprop combined with positive aggregation are invariant to higher layer parameters corroborates the recent work of~\citet{sixt2019explanations} along with previous evidence by~\citet{nie2018theoretical} and~\citet{mahendran2016salient}.

The gold standard for assessing the effectiveness of an explanation is a human subject study~\citep{doshi2017towards}.~\citet{poursabzi2018manipulating} manipulate the features of a linear model trained to predict housing prices to assess how well end-users can identify model mistakes. More recently, human subject tests of feature attributions have cast doubt on the ability of these approaches to help end-users debug erroneous predictions and improve human performance on downstream tasks~\citep{chu2020visual, shen2020useful}. In a cooperative setting, \citet{lai2019human} find that the humans exploit label information and~\citet{feng2019can} demonstrate how to assess explanations in a natural language setting. Similarly,~\citet{alqaraawi2020evaluating} find that the LRP explanation method (see Section~\ref{section:methods}) improves participant understanding of model behavior for an image classification task, but provides limited utility to end-users when predicting the model's output on new inputs.

Feature attributions can be easily manipulated, providing evidence for a collective `weakness' of current approaches~\citep{ghorbani2017interpretation, heo2019fooling, slack2020fooling, lakkaraju2020fool}. While susceptibility is an important issue, our work focuses on providing insights when model bugs are `unintentionally' created.

\section{Bug Characterization, Explanation Methods, \& User Study}
\label{section:setup}
We now present our characterization of model bugs, provide an overview of the explanation methods assessed, and close with a background on the human subject study.\footnote{We will provide code to replicate our findings at: https://github.com/adebayoj/explaindebug.git.} 

\subsection{Characterizing Model Bugs.} \label{section:bugformalization}
We define model \textit{bugs} as contamination in the learning and/or prediction pipeline that causes the model to produce incorrect predictions or learn error-causing associations. We restrict our attention to the standard supervised learning setting, and categorize bugs based on their source. Given input-label pairs, $\{x_i, y_i\}^n_i$, where $x\in\mathcal{X}$ and $y \in \mathcal{Y}$, a classifier's goal is to learn a function, $f_{\textcolor{myred}{\theta}} : \mathcal{X} \rightarrow \mathcal{Y}$, that generalizes. $f_{\textcolor{myred}{\theta}}$ is then used to predict test examples, $x_\mathrm{test} \in \mathcal{X}$, as $y_{\mathrm{test}} = f_{\textcolor{myred}{\theta}}(x_{\mathrm{test}})$. Given a loss function $L$, and model parameter, $\theta$, for a model family, we provide a categorization of bugs as model, data and test-time contamination: 
\begin{align*} 
\text{Learning:}~\argmin_{\underbrace{\textcolor{myred}{\theta}}_{\textcolor{myred}{\mathrm{\substack{\textbf{Model~Contamination}}}}}} L(\overbrace{\textcolor{blue}{(X_{\mathrm{train}}, Y_{ \mathrm{train})}}}^{\textcolor{blue}{\mathrm{\substack{\textbf{Data~Contamination} }}}}; \theta);\\   
\text{Prediction:}~y_{\mathrm{test}} = f_{\textcolor{myred}{\theta}}(\overbrace{\textcolor{mygreen}{x_{\mathrm{test}}}}^{\textcolor{mygreen}{\mathrm{\substack{\textbf{Test-Time~Contamination}}}}}) \label{eq2:debugging}.
\end{align*}

\textbf{Data Contamination bugs} are caused by defects in the training data, either in the input features, the labels, or both. For example, a few incorrectly labeled data can cause the model to learn wrong associations. Another bug is a spurious correlation training signal. For example, consider an object classification task where all birds appear against a blue sky background. A model trained on this dataset can learn to associate blue sky backgrounds with the bird class; such dataset biases frequently occur in practice~\citep{badgeley2019deep, ribeiro2016should}.

\textbf{Model Contamination bugs} are caused by defects in the model parameters. For example, bugs in the code can cause accidental re-initialization of model weights.

\textbf{Test-Time Contamination bugs} are caused by defects in test-input, including domain shift or pre-processing mismatch at test time.

\begin{table*}[t]
\centering
\rowcolors{1}{}{lightgray}
\resizebox{.9\linewidth}{!}{
\begin{tabular}{llc}
\hline
\textbf{Bug Category}   & \textbf{Specific Examples tested} & \textbf{Formalization} \\
\hline
Data Contamination      & Spurious Correlation    & $\begin{aligned}[t] 
\argmin_{\textcolor{myred}{\theta}} L(\textcolor{blue}{X_\mathrm{spurious~artifact}, Y_\mathrm{train}}; \theta)
\end{aligned}$ \\

                     & Labelling Errors    & $\begin{aligned}[t] 
\argmin_{\textcolor{myred}{\theta}} L(\textcolor{blue}{X_\mathrm{train}, Y_\mathrm{wrong~label}}; \theta)
\end{aligned}$       \\

Model Contamination      & Initialized Weights    & $\begin{aligned}[t] 
f_{\textcolor{myred}{\theta{\mathrm{init}}}}({\textcolor{mygreen}{x_{\mathrm{test}}}})
\end{aligned}$      \\

Test-Time Contamination      & Out of Distribution (OOD)    & $\begin{aligned}[t] 
 f_{\textcolor{myred}{\theta}}({\textcolor{mygreen}{x_{\mathrm{OOD}}}})
\end{aligned}$      \\

\hline
\end{tabular}
}
\caption{Example bugs we test for each bug categories and their formalization.}
\label{tab:bugformalization_main}
\end{table*}

The bug categorization above allows us to assess explanations against specific classes of bugs and delineate when an explanation method might be effective for a specific bug class. We assess a range of explanation methods applied to models with specific instances of each bug, as shown in Table~\ref{tab:bugformalization_main}. 

\subsection{Explanation Methods}\label{section:methods}
\begin{figure*}[h]
\centering
\includegraphics[scale=0.35,page=2]{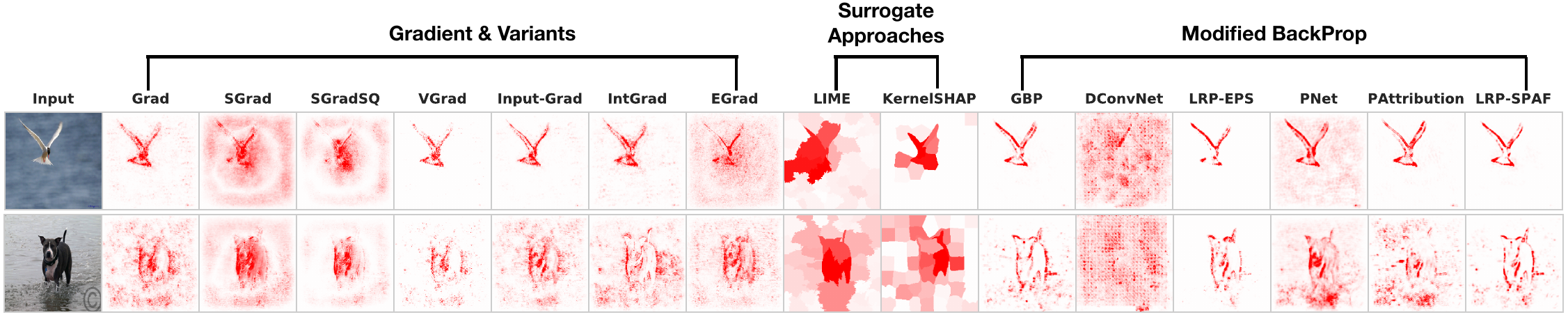}
\caption{\textbf{Attribution Methods Considered.} The Figure shows feature attributions for two inputs for a CNN model trained to distinguish between birds and dogs.}
\label{fig:methodsdemo}
\end{figure*}

We focus on~\textit{feature attribution methods} that provide a `relevance' score for the dimensions of input towards a model's output. For deep neural networks (DNNs) trained on image data, the feature-relevance can be visualized as a heat map, as in Figure~\ref{fig:methodsdemo}.

An attribution functional, $E : \mathcal{F} \times \mathbb{R}^d \times \mathbb{R} \rightarrow  \mathbb{R}^d$, maps the input, $x_i \in \mathbb{R}^d$, the model, $F \in \mathcal{F}$, output, $F_k(x)$, to an attribution map, $M_{x_i} \in \mathbb{R}^d$. Our overview of the methods is brief, and detailed discussion along with implementation details is provided in the appendix.

\underline{\textbf{Gradient (Grad) \& Variants.}} We consider: 1) The \textit{Gradient (Grad)} ~\citep{simonyan2013deep, baehrens2010explain} map, $\vert \nabla_{x_i}F_i(x_i)\vert$; 2) \textit{SmoothGrad (SGrad)}~\citep{smilkov2017smoothgrad}, $E_{\mathrm{sg}}(x) = \frac{1}{N}\sum_{i=1}^N \nabla_{x_i}F_i(x_i + n_i)$ where $n_i$ is Gaussian noise; 3) \textit{SmoothGrad Squared (SGradSQ)}~\citep{hooker2018evaluating}, the element-wise square of SmoothGrad; 4) \textit{VarGrad (VGrad)}~\citep{adebayo2018local}, the variance analogue of SmoothGrad; \& 5) ~\textit{Input-Grad}~\citep{shrikumar2016not} the element-wise product of the gradient and input $\vert \nabla_{x_i}F_i(x_i) \vert \odot x_i$. We also consider: 6)~\textit{Integrated Gradients (IntGrad)}~\citep{sundararajan2017axiomatic} which sums gradients along an interpolation path from the ``baseline input'', $\bar{x}$, to $x_i$: $M_{\mathrm{IntGrad}}(x_i) = (x_i - \bar{x}) \times \int_{0}^1{\frac{\partial S(\bar{x} + \alpha(x_i-\bar{x}))}{\partial x_i}} d\alpha$; and 7)~\textit{Expected Gradients (EGrad)}~\citep{erion2019learning} which computes IntGrad but with a baseline input that is an expectation over the training set.

\underline{\textbf{Surrogate Approaches.}} LIME~\citep{ribeiro2016should} and SHAP~\citep{lundberg2017unified} locally approximate $F$ around $x_i$ with a simple function, $g$, that is then interpreted. SHAP provides a tractable approximation to the Shapley value~\citep{shapley1988value}.

\underline{\textbf{Modified Back-Propagation.}} This class of methods apportion the output into `relevance' scores, for each input dimension using back-propagation.~\textit{DConvNet}~\citep{zeiler2014visualizing} \& \textit{Guided Back-propagation (GBP)}~\citep{springenberg2014striving} modify the gradient for a ReLU unit. \textit{Layer-wise relevance propagation (LRP)}~\citep{bach-plos15, binder-icann16, montavon-pr17, kohlbrenner-ijcnn20} methods specify `relevance' rules that modify the back-propagation. We consider~\textit{LRP-EPS}, and ~\textit{LRP sequential preset-a-flat (LRP-SPAF)}.~\textit{PatternNet (PNet)} and \textit{Pattern Attribution (PAttribution)}~\citep{kindermans2018learning} decompose the input into signal and noise components, and back-propagate relevance for the signal component. 

\underline{\textbf{Attribution Comparison.}} We measure visual and feature ranking similarity with the structural similarity index (SSIM)~\citep{wang2004image} and Spearman rank correlation metrics, respectively.

\subsection{Overview of Human Subject Study}\label{userstudyintro}
\underline{\textbf{Task \& Setup:}}  We designed a study to measure end-users' ability to assess the reliability of classification models using feature attributions. Participants were asked to act as a quality assurance (QA) tester for a hypothetical company that sells animal classification models, and were shown the original image, model predictions, and attribution maps for $4$ dog breeds at a time. They then rated how likely they are to recommend the model for sale to external customers using a 5 point-Likert scale, and a rationale for their decision. Participants chose from $4$ pre-created answers (Figure~\ref{fig:mainresultboxplot}-b) or filled in a free form answer. Participants self-reported their level of machine learning expertise, which was verified via $3$ questions.

\underline{\textbf{Methods:}} We focus on a representative subset of methods for the study: Gradient, Integrated Gradients, and SmoothGrad (See additional discussion on selection criteria in the Appendix).
 
 \underline{\textbf{Bugs:}} We tested the bugs described in Table~\ref{tab:bugformalization_main} along with a model with no bugs.

\section{Debugging Data Contamination}
\textbf{Overview.} We assess whether feature attributions can detect spurious training artifacts and mislabelled training examples. Spurious artifacts are signals that encode or correlate with the label in the training set but provide no  meaningful connection to the data generating process. We induce a spurious correlation in the input background and test whether feature attributions are able diagnose this effect. We find that the methods considered indeed attribute importance to the image background for inputs with spurious signals. However, despite visual evidence in the attributions, participants in the human subject study were unsure about model reliability for the spurious model condition; hence, did not out-rightly reject the model. 

\begin{figure*}[h]
\centering
\includegraphics[scale=0.40,page=10]{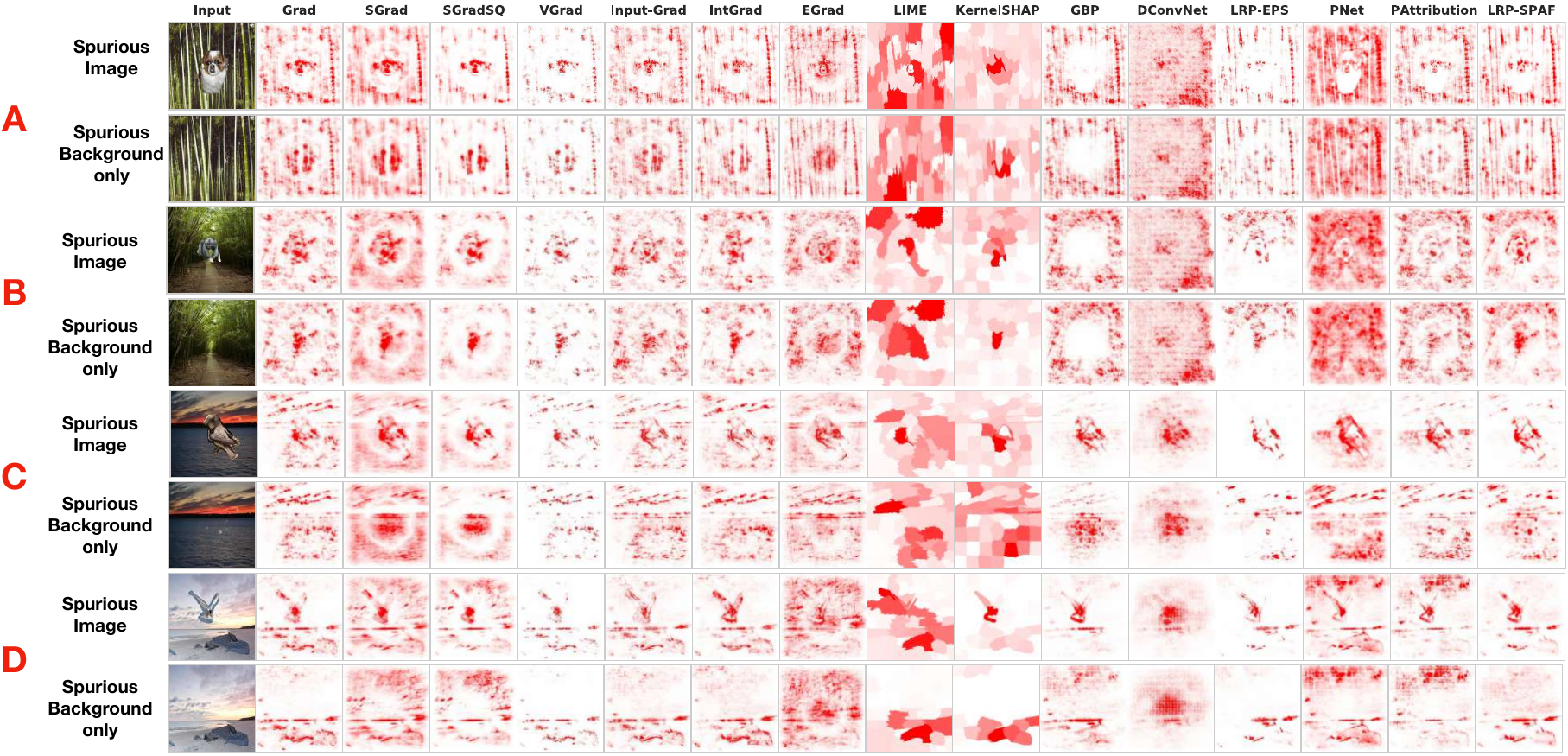}
\caption{\textbf{Feature Attributions for Spurious Correlation Bugs.} Figure shows attributions for $4$ inputs for the BVD-CNN trained on spurious data. A \& B show two dog examples, and C \& D are bird examples. The first row shows the input (dog or bird) on a spurious background. The second row shows the attributions of only the spurious background. Notably, we observe that the feature attribution methods place emphasis on the background. See Table~\ref{tab:spuriousmetrics} for metrics.}
\label{fig:spuriousartifact}
\end{figure*}

For mislabeled examples, we compare attributions for a training input derived from: 1) a model where this training input had the correct label, and 2) the same model settings but trained with this input mislabeled. If the attributions under these two settings are similar, then such a method is unlikely to be useful for identifying mislabeled examples. We observe that attributions for mislabeled examples, across all methods, show visual similarity. 

\textbf{General Data and Model Setup.} We consider a birds-vs-dogs binary classification task. We use dog breeds from the Cats-v-Dogs dataset~\citep{parkhi12a} and Bird species from the Caltech-UCSD dataset~\citep{WahCUB_200_2011}. On this dataset, we train a CNN with $5$ convolutional layers and $3$ fully-connected layers (we refer to this architecture as \textit{BVD-CNN} from here on) with ReLU activation functions but sigmoid in the final layer. The model achieves a test accuracy of $94$-percent.

\begin{wrapfigure}{r}{0.35\textwidth}
\begin{center}
  \includegraphics[scale=0.35, page=3]{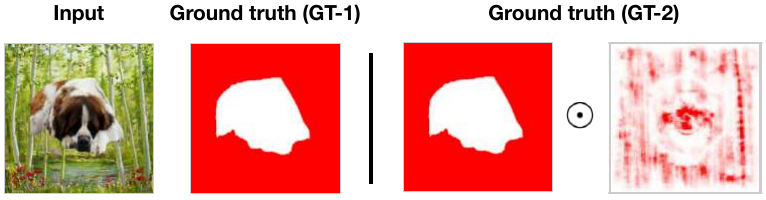}
 \end{center}
\caption{Ground Truth Attribution for Spurious Correlation.}
\label{fig:gtmask}
\end{wrapfigure}

\subsection{Spurious Correlation Training Artifacts}
\textbf{Spurious Bug Implementation.} We introduce spurious correlation by placing all birds onto one of the sky backgrounds from the places dataset~\citep{zhou2017places}, and all dogs onto a bamboo forest background (see Figure~\ref{fig:spuriousartifact}). BVD-CNN trained on this data achieves a $97$ percent accuracy on a sky-vs-bamboo forest test set (without birds or dogs) indicating that the model indeed learned the spurious association.

\textbf{Results.} To quantitatively measure whether attribution methods reflect the spurious background, we compare attributions to two ground truth masks (GT-1 \& GT-2). As shown in Figure~\ref{fig:gtmask}, we consider an ideal mask that apportions all relevance to the background and none to the object part. Next, we consider a relaxed version that weights the first ground truth mask by the attribution of a spurious background without the object. In Table~\ref{tab:spuriousmetrics}, we report SSIM comparison scores across all methods for both ground-truth masks. For \textit{GT-2}, scores range from a minimum of $0.78$ to maximum of $0.98$; providing evidence that the attributions identify the spurious background signal. We find similar evidence for \textit{GT-1}.

\begin{table*}[th]
\centering
\rowcolors{1}{}{lightgray}
\resizebox{\linewidth}{!}{\begin{tabular}{cccccccccccccccc}
\hline
\textbf{Metric} & \textbf{Grad} & \textbf{SGrad} & \textbf{SGradSQ} & \textbf{VGrad} & \textbf{Input-Grad} & \textbf{IntGrad} & \textbf{EGrad} & \textbf{LIME} & \textbf{KernelSHAP} & \textbf{GBP} & \textbf{DConvNet} & \textbf{LRP-EPS} & \textbf{PNet} & \textbf{PAttribution} & \textbf{LRP-SPAF} \\
\hline
SSIM-GT1 & 0.62 & 0.63 & 0.063 & 0.075 & 0.69 & 0.7 & 0.63 & 0.59 & 0.58 & 0.58 & 0.6 & 0.65 & 0.51 & 0.44 &  0.69 \\
SSIM-GT1 (SEM) & 0.012 & 0.013 & 0.0077 & 0.0089 & 0.019 & 0.019 & 0.024 & 0.021 & 0.037 & 0.019 & 0.017 & 0.039 & 0.036 & 0.018 &  0.028 \\
SSIM-GT2 & 0.83 & 0.83 & 0.89 & 0.98 & 0.85 & 0.85 & 0.85 & 0.88 & 0.78 & 0.82 & 0.83 & 0.85 & 0.85 & 0.8 & 0.85 \\
SSIM-GT2 (SEM) & 0.013 & 0.013 & 0.02 & 0.0024 & 0.013 & 0.012 & 0.012 & 0.011 & 0.044 & 0.013 & 0.013 & 0.012 & 0.013 & 0.018 & 0.013 \\
\hline
\end{tabular}}
\caption{\textbf{Similarity between attribution masks for inputs with spurious background and ground truth masks.} SSIM-GT1 measures the visual similarity between an ideal spurious input mask and the GT-1 as shown in Figure~\ref{fig:gtmask}. SSIM-GT2 measures visual similarity for the GT-2.  We also include the standard error of the mean (SEM) for each metric, which was computed across $190$ inputs. To calibrate this metric, the mean SSIM between a randomly sampled Gaussian attribution and the spurious attributions which is: $3e^{-06}$.}
\label{tab:spuriousmetrics}
\end{table*}

\begin{figure*}[!h]
\centering
\includegraphics[scale=0.33]{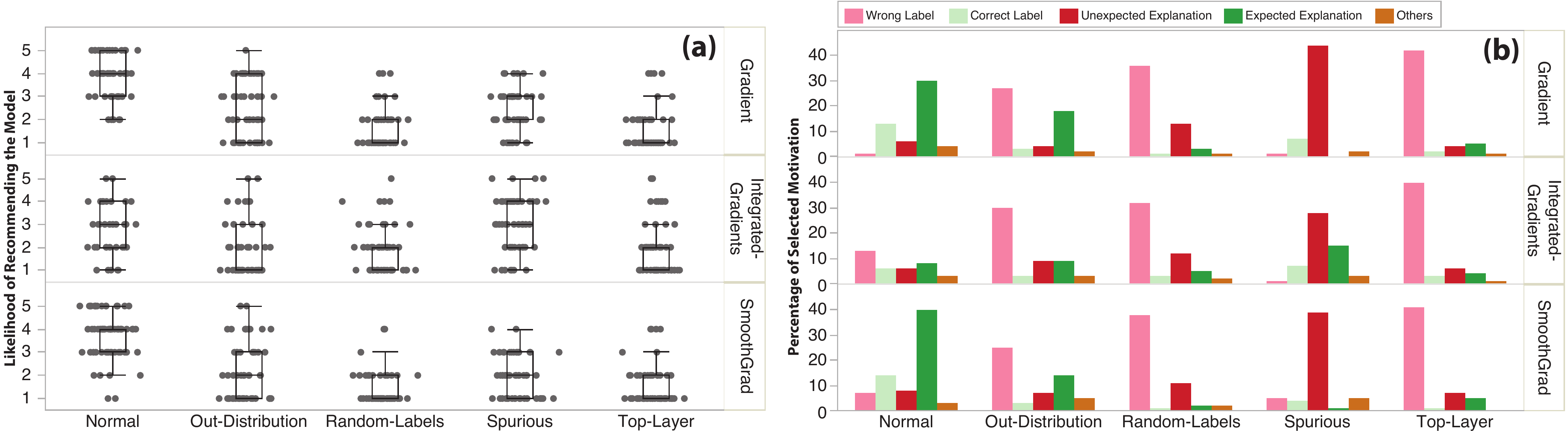}
\caption{\textbf{A: Participant Responses from User Study.} Box plot of participants responses for $3$ attribution methods: \textit{Gradient, SmoothGrad, and Integrated Gradients}, and $5$ model conditions tested. On the vertical axis is likert scale from $1:$ \textit{Definitely Not} to $5:$ \textit{Definitely}. Participants were instructed to select  `Definitely' if they deemed the dog-breed classification model ready to be sold to customers. \textbf{B: Motivation for Selection.} Participants' selected motivations (\%) for the recommendation made. As shown in the legend, users could select one of 4 options or insert an open-ended response.}
\label{fig:mainresultboxplot}
\end{figure*}

\textbf{Insights from Human Subject Study: users are uncertain.} Figure~\ref{fig:mainresultboxplot} reports results from the human subject study, where we assess end-users' ability to reliably use attribution to identify models relying on spurious training set signals. For a normal model, the median Likert scores are $4$, $4$, $3$ for Gradient, SmoothGrad, and Integrated Gradients respectively. Selecting a likert score of $1$ means a user will `definitely not' recommend the model, while $5$ means they will `definitely' recommend the model. Consequently, users adequately rate a normal model. In addition, $30$ and $40$ percent (See Figure~\ref{fig:mainresultboxplot}-Right)  of participants, for Gradient and SmoothGrad respectively, indicate that the attributions for a normal model `highlighted the part of the image that they expected it to focus on'.

For the `spurious model', the Likert scores show a wider range. While the median scores are $2$, $2$, $3$ for Gradient, SmoothGrad, and Integrated Gradients respectively, some end-users still recommend this model. For each attribution type, a majority of end-users indicate that the attribution `did not highlight the part of the image that I expected it to focus on'. Despite this, end-users do not convincingly reject the spurious model like they do for the other bug conditions. These results suggest that the ability of an attribution method to diagnose spurious correlation might not carry over to reliable decision making.

\subsection{Mislabelled Training Examples}
\textbf{Bug Implementation.} We train a BVD-CNN model on a birds-vs-dogs dataset where $10$ percent of training samples have their labels flipped. The model achieves a $93.2$, $91.7$, $88$ percent accuracy on the training, validation, and test sets.

\begin{figure*}[!h]
\centering
\includegraphics[scale=0.55, page=8]{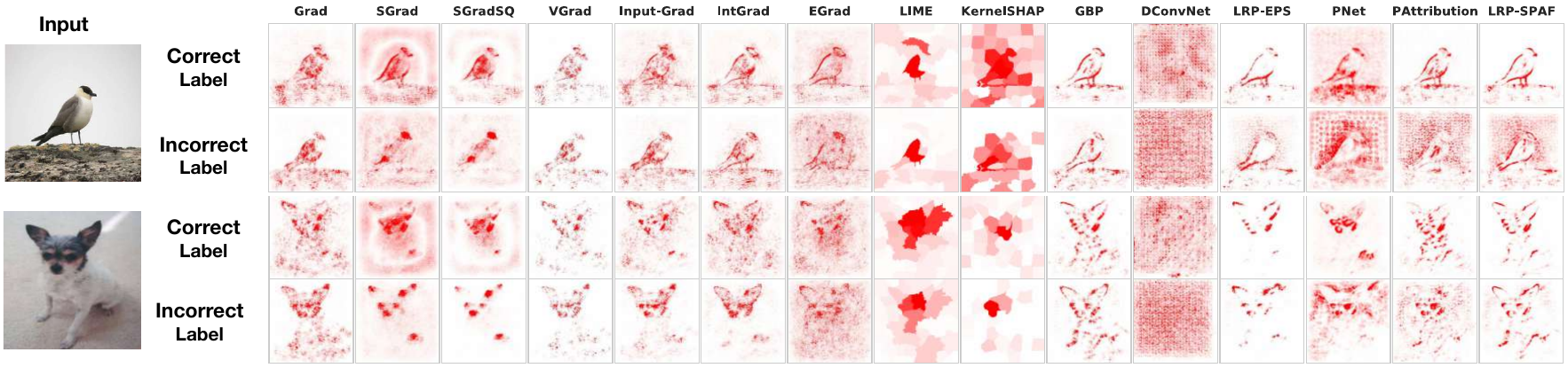}
\caption{\textbf{Diagnosing Mislabelled Training Examples.} The Figure shows two training inputs along with feature attributions for each method. The correct label row corresponds to feature attributions derived from a model with the correct label in the training set. The incorrect-label row shows feature attributions derived from a model with the wrong label in the training set. We see that the attributions under both settings are visually similar.}
\label{fig:mislabelled}
\end{figure*}

\textbf{Results.} We find that attributions from mislabelled examples for a defective model are visually similar to attributions for these same examples but derived from a model with correct input labels (examples in Figure~\ref{fig:mislabelled}). We find that the SSIM between the attributions of a correctly labeled instance, and the corresponding incorrectly labeled instance, are in the range $0.73-0.99$ for all methods tested. These results indicate that the attribution methods tested might be ineffective for identifying mislabelled examples. We refer readers to Section~\ref{appendix:additionalfigures} of the Appendix for visualizations of additional examples.

\textbf{Insights from Human Subject Study: users use prediction labels, not attribution methods.} In contrast to the spurious setting, participants reject mislabelled examples with median Likert scores $1$, $2$, and $1$ for Gradient, SmoothGrad, and Integrated Gradients respectively. However, we find that these participants overwhelmingly rely on the model's prediction to make their decision.

\section{Debugging Model Contamination}
\label{section:modelcontamination}
We next evaluate bugs related to model parameters. Specifically, we consider the setting where the weights of a model are accidentally re-initialized prior to prediction~\citep{adebayo2018sanity}. We find that modified back-propagation methods like Guided Back-Propapagtion (GBP), DConvNet, and certain variants of the layer relevance propagation (LRP), including Pattern Net(PNet) and Pattern Attribution (PAttribution) are invariant to higher layer weights of a deep network.

\textbf{Bug Implementation.} We instantiate this bug on a pre-trained VGG-16 model on Imagenet~\citep{russakovsky2015imagenet}. Similar to~\citet{adebayo2018sanity}, we re-initialize the weights of the model starting at the top layer, successively, all the way to the first layer. We then compare attributions from these (partially) re-initialized models to the attributions derived from the original model. 

\begin{figure*}[h]
\centering
\includegraphics[scale=0.55, page=12]{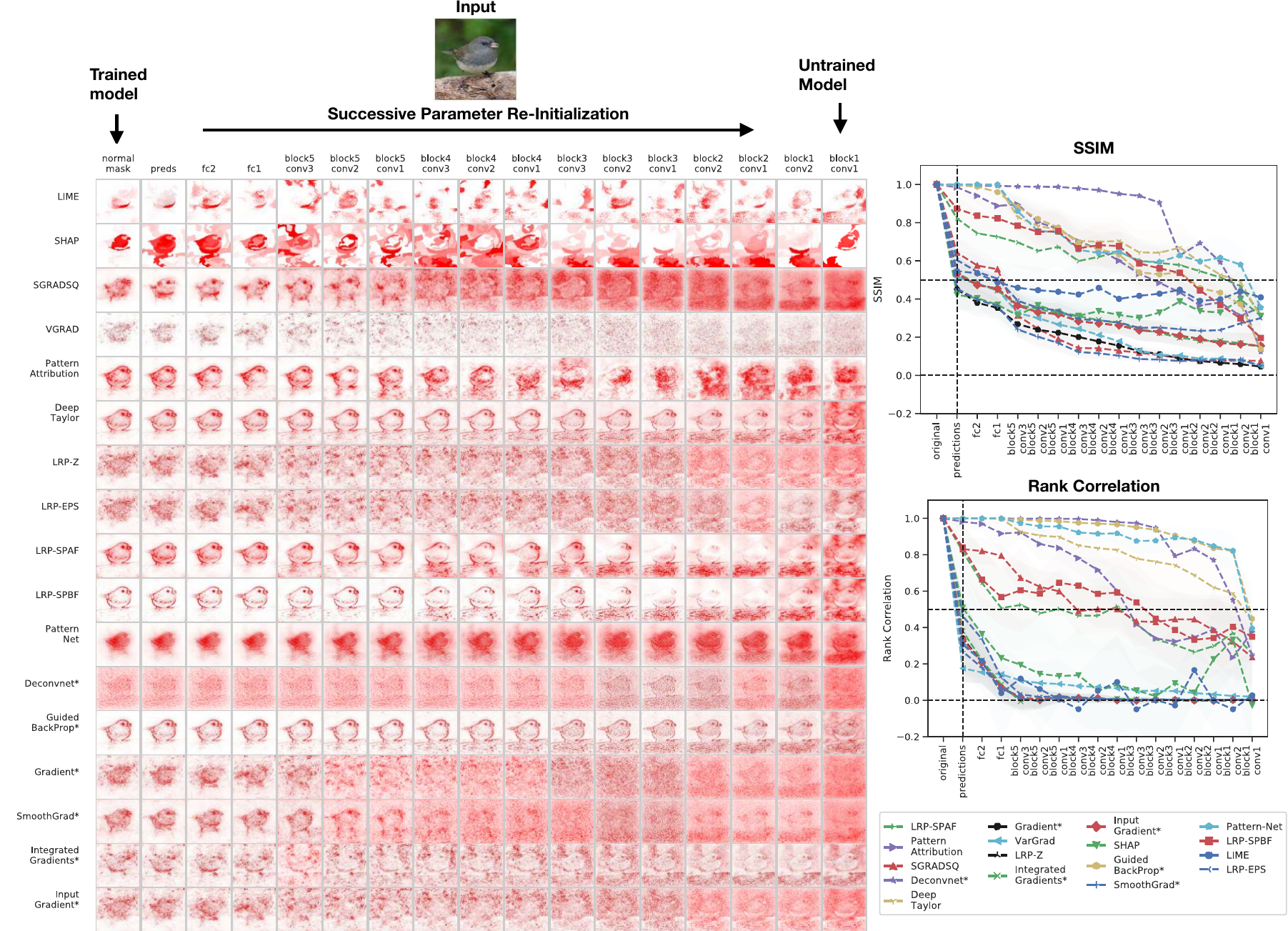}
\caption{\textbf{Evolution of several model attributions for successive weights re-initialization of a VGG-16 model trained on ImageNet.} Qualitative results (left) and quantitative results (right). The last column in qualitative results corresponds to a network with completely re-initialized weights.}
\label{fig:cascadingdemo}
\end{figure*}

\textbf{Results: modified back-propagation methods are parameter invariant.} As seen in Figure~\ref{fig:cascadingdemo}, the class of modified back-propagation methods, including Guided BackProp, Deconvnet, DeepTaylor, PatternNet, Pattern Attribution, and LRP-SPAF are visually and quantitatively invariant to higher layer parameters of the VGG-16 model. This finding corroborates prior results for Guided Backprop and Deconvnet~\citep{mahendran2016salient, nie2018theoretical, adebayo2018sanity}. These results also support the recent findings of~\citet{sixt2019explanations}, who prove that these modified back-propagation approaches produce attributions that converge to a rank-$1$ matrix.

\textbf{Insights from Human Subject Study: users use prediction labels, not attribution methods.} We observe that participants conclusively reject a model whose top layer has been re-initialized purely based on the classification labels, and rarely based on wrong attributions. (Figure~\ref{fig:mainresultboxplot}).

\section{Debugging Test-Time Contamination}
\label{section:testtimecontamination}
A model is at risk of providing errant predictions when given inputs that have distributional characteristics different from the training set. To assess the ability of feature attributions to diagnose domain shift, we compare attributions derived, for a given input, from an \textit{in-domain model} with those derived from \textit{out-of-domain model}. For example, we compare the attribution for an MNIST digit, derived from a model trained on MNIST, to an attribution for the same digit, but derived from a model trained on Fashion MNIST, ImageNet, and a birds-vs-dogs model. We find visual similarity for certain settings: for example, feature attributions for a Fashion MNIST input derived from a VGG-16 model trained on ImageNet are visually similar to attributions for the same input on a model trained on Fashion MNIST. However, the quantitative ranking of the input dimensions are widely different.

\begin{figure*}[h]
\centering
\includegraphics[scale=0.3, page=3]{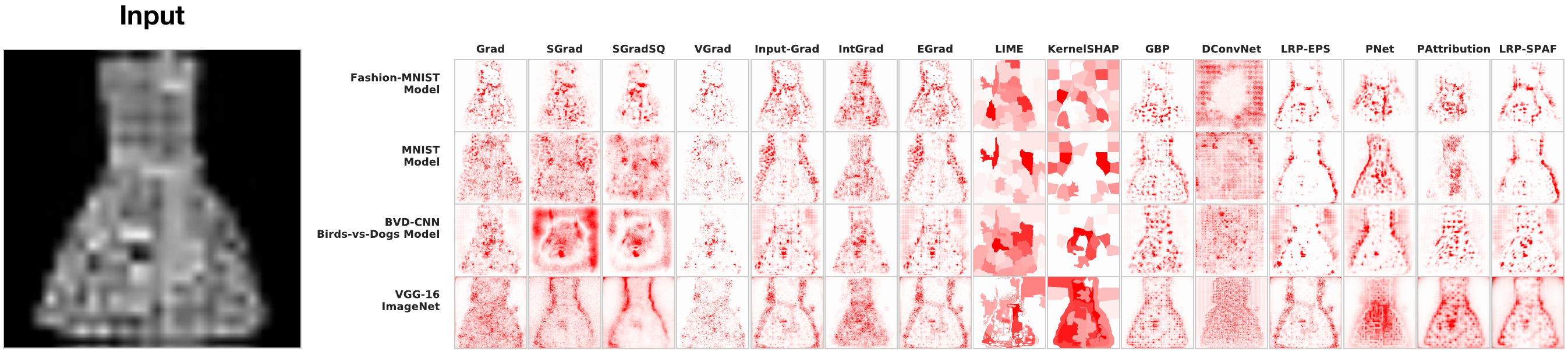}
\caption{\textbf{Fashion MNIST OOD on several models.} The first row shows feature attributions on a model trained on Fashion MNIST. In the subsequent rows, we show feature attributions for the same input on an MNIST model, BVD-CNN model trained on birds-vs-dogs, and lastly, a pre-trained VGG-16 model on ImageNet.}
\label{fig:testtimedemo}
\end{figure*}

\begin{table*}[th]
\centering
\rowcolors{1}{}{lightgray}
\resizebox{\linewidth}{!}{\begin{tabular}{cccccccccccccccc}
\hline
\textbf{Metric} & \textbf{Grad} & \textbf{SGrad} & \textbf{SGradSQ} & \textbf{VGrad} & \textbf{Input-Grad} & \textbf{IntGrad} & \textbf{EGrad} & \textbf{LIME} & \textbf{KernelSHAP} & \textbf{GBP} & \textbf{DConvNet} & \textbf{LRP-EPS} & \textbf{PNet} & \textbf{PAttribution} & \textbf{LRP-SPAF} \\
\hline
SSIM (FMNIST $\rightarrow$ MNIST Model) & 0.7 & 0.54 & 0.49 & 0.92 & 0.71 & 0.69 & 0.71 & 0.46 & 0.41 & 0.81 & 0.5 & 0.77 &0.58 & 0.77 & 0.66\\
SEM & 0.0093 & 0.012 & 0.016 & 0.0047 & 0.01 & 0.015 & 0.01 & 0.02 & 0.024& 0.014 & 0.01 &0.02 & 0.026 & 0.009 & 0.03\\
RK (FMNIST $\rightarrow$ MNIST Model) & 0.0013 & 8.8e-4 & 0.37 & 0.37 & 0.0021 & -0.003& 0.002 &-0.01 &  0.034 & 0.51 & 0.027 &0.011 &-0.14 & 0.0082&0.12 \\
SEM & 0.0016 & 0.0032 & 0.026 & 0.029 & 0.002 & 0.002 & 0.002 &0.04 & 0.028 & 0.014 & 6e-4 & 0.0034&0.027 &  0.0026& 0.023\\
SSIM (FMNIST $\rightarrow$ BVD-CNN) & 0.7 & 0.5 & 0.55 & 0.93 & 0.72 & 0.7 & 0.72 & 0.72& 0.72 & 0.82 & 0.63 & 0.79& 0.53 & 0.36 & 0.66\\
SEM & 0.0083 & 0.011 & 0.013 & 0.0045 & 0.009 &0.013 & 0.009 & 0.009 & 0.01 & 0.009 & 0.014 & 0.019 &0.03 & 0.025 & 0.035\\
RK (FMNIST $\rightarrow$ BVD-CNN) & 0.0012 & 0.0078 & 0.43 & 0.25 & 0.0002 & 0.002 & 0.00025 & 0.18 & 0.067 & 0.078& -0.05 &-0.013 & 0.25 &-0.0095 & 0.044\\
SEM & 8.5e-4 & 0.0017 & 0.009 & 0.011 & 0.0007 & 0.001 & 0.0007 & 0.04 & 0.034& 0.008 & 0.0011 & 0.0027 & 0.045 &0.0023 & 0.02\\
SSIM (FMNIST $\rightarrow$ VGG-16 ImageNet) & 0.57 & 0.46 & 0.5 & 0.87 & 0.64 & 0.67 & 0.64 & 0.5 & 0.38 & 0.8& 0.36 &0.64 & 0.66 & 0.12 & 0.2\\
SEM & 0.012 & 0.011 & 0.015 & 0.0056 & 0.01 & 0.015 & 0.011& 0.015 & 0.03 & 0.009&  0.01&0.02 & 0.018 & 0.0049& 0.024\\
RK (FMNIST $\rightarrow$ VGG-16 ImageNet) & -0.0023 & -0.0098 & -0.0097 & 0.028 & -0.0025 & -0.0017& -0.0025& 0.005 & -0.045& 0.25 &0.03 & 0.0045 & 0.32&0.066 &0.14 \\
SEM & 0.0017 & 0.0025 & 0.02 & 0.018 & 0.0023 & 0.0016 & 0.002 & 0.033 & 0.024 & 0.004 & 0.0035 & 0.0018 & 0.034&0.0053 &0.019 \\
\hline
\end{tabular}}
\caption{\textbf{Test-time Explanation Similarity Metrics.} We observe visual similarity but no ranking similarity. We show each metric along with the standard error of the mean calculated for $190$ examples. FMNIST $\rightarrow$ MNIST model means a comparison of FMNIST attributions for an FMNIST model with FMNIST attributions derived from ~\textit{an MNIST model}. We present both SSIM and Rank correlation metrics.}
\label{tab:testtimemetrics}
\end{table*}
 
\textbf{Bug Implementation.} We consider $4$ dataset-model pairs: a BVD-CNN trained on MNIST, Fashion MNIST, the Birds-vs-dogs data, and lastly a VGG-16 model trained on ImageNet. We present results on Fashion MNIST. Concretely, we compare 1) feature attributions of Fashion MNIST examples derived from a model trained on Fashion MNIST, and 2) feature attributions of Fashion MNIST examples for models trained on MNIST, the birds-vs-dogs dataset, and ImageNet.

\textbf{Results.} As shown in Figure~\ref{fig:testtimedemo}, we observe visual similarity between in-domain Fashion MNIST attributions, and attributions for these samples on other models. As seen in Table~\ref{tab:testtimemetrics}, we observe visual similarity, particularly for the VGG-16 model on ImageNet, but essentially no correlation in feature ranking.

\textbf{Insights from Human Subject Study: users use prediction labels, not the attributions.} For the domain shift study, we show participants attribution of dogs that were not used during training, and whose breeds differed from those that the model was trained to predict. We find that users do not recommend a model under this setting due to wrong prediction labels (Figure~\ref{fig:mainresultboxplot}).

\section{Discussion \& Conclusion}
\label{section:conclusion}
Debugging machine learning models remains a challenging endeavor, and model explanations could be a useful tool in that quest. Even though a practitioner or a researcher may have a large class of explanation methods available, it is still unclear which methods are useful for what bug type. This work aims to address this gap by first, categorizing model bugs into: data, model, and test-time contamination bugs, then testing feature attribution methods, a popular explanation approach for DNNs trained on image data, against each bug type. Overall, we find that feature attribution methods are able to diagnose the spatial spurious correlation bug tested, but do not conclusively help to distinguish mislabelled examples for normal ones. In the case of model contamination, we find that certain feature attributions that perform positive aggregation while computing feature relevance with modified back-propagation produce attributions that are invariant to the parameters of the higher layers for a deep model. This suggests that these approaches might not be effective for diagnosing model contamination bugs. We also find that attributions of out-of-domain inputs are similar to attributions for these inputs on an in-domain model, which suggests caution when visually inspecting these explanations, especially for image tasks. We also conduct human subject tests to assess how well end-users can use attributions to assess model reliability. Here we find that the end-users relied, primarily, on model predictions for diagnosing model bugs.

\paragraph{Limitations.} Our findings come with certain limitations and caveats. The bug characterization presented only covers the standard supervised learning pipeline and might not neatly capture bugs that result from a combination of factors. We only focused on feature attributions: however, other methods such as approaches based on `concept' activation~\citep{kim2018interpretability}, model representation dissection~\citep{bau2017network}, and training point ranking~\citep{koh2017understanding, yeh2018representer, pruthi2020estimating} might be more suited to the debugging tasks studied here. Indeed, initial exploration of the `concept' activation method TCAV and training point ranking based on influence functions suggests that these approaches are promising (See Appendix for analysis). For the human subject experiments, our finding that the participants mostly relied on the labels instead of the feature attributions might be a consequence of the dog breed classification task. It is unclear whether participants would still rely of model predictions for tasks in which they have no expertise or prior knowledge.

The goal of this work is to provide guidance for researchers and practitioners seeking to use feature attributions for model debugging. We hope our findings can serve as a first step towards more rigorous approaches for assessing the utility of explanation methods.


\section*{Acknowlegdements}
We thank Hal Abelson, Danny Weitzner, Taylor Reynolds, and Anonymous reviewers for feedback on this work. We are grateful to the MIT Quest for Intelligence initiative for providing cloud computing credits for this work. JA is supported by the Open Philanthropy Fellowship.

\newpage

\bibliographystyle{plainnat}
\bibliography{references}

\newpage
\appendix
\newpage
\part{Appendix} 
\parttoc 

\section{Additional Discussion of Bug Categorization Setup}
\label{appendix:bugcategorization}
\begin{figure}[!h]
\centering
\includegraphics[scale=0.5, page=37]{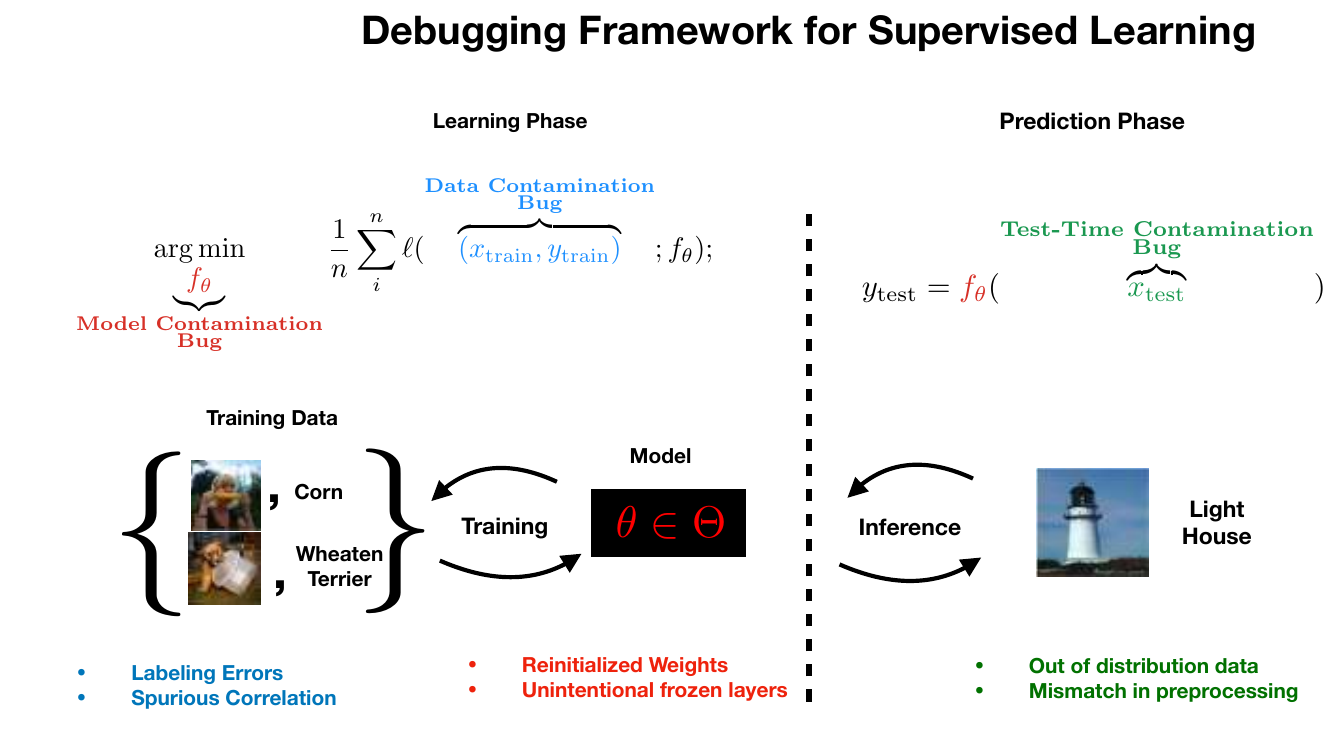}
\caption{We show the debugging schematic here.}
\label{fig:appendixschematic}
\end{figure}

In this section, we provide two additional formalization of bugs beyond those discussed in the main document: Frozen layers and Pre-Processing Mismatch. In the case of frozen layers, this is a bug whereby one of the layers of a deep network is accidentally kept fixed during training. In the case of pre-processing mismatch, a test-input can be pre-processed under settings that are different from those used for the training data. In such a case, it is possible the the model will produce wrong predictions due to the mismatch in input pre-processing. We formalize both these bugs in Table~\ref{tab:bugformalizationexpanded}.

\begin{table*}[h]
\centering
\rowcolors{1}{}{lightgray}
\begin{tabular}{llc}
\hline
\textbf{Bug Class}   & \textbf{Specific Examples} & \textbf{Formalization} \\
\hline
Data Contamination      & Spurious Correlation    & $\begin{aligned}[t] 
\argmin_{\textcolor{myred}{\theta}} L(\textcolor{blue}{X_\mathrm{spurious~artifact}, Y_\mathrm{train}}; \theta)
\end{aligned}$ \\

Data Contamination      & Labelling Errors    & $\begin{aligned}[t] 
\argmin_{\textcolor{myred}{\theta}} L(\textcolor{blue}{X_\mathrm{train}, Y_\mathrm{wrong~label}}; \theta)
\end{aligned}$       \\

Model Contamination      & Initialized Weights    & $\begin{aligned}[t] 
f_{\textcolor{myred}{\theta{\mathrm{init}}}}({\textcolor{mygreen}{x_{\mathrm{test}}}})
\end{aligned}$      \\

Model Contamination      & Frozen Layers    & $\begin{aligned}[t] 
\argmin_{\textcolor{myred}{\theta{\mathrm{frozen}}}} L(\textcolor{blue}{X_\mathrm{train}, Y_\mathrm{wrong~label}}; \theta)
\end{aligned}$       \\

Test-Time Contamination      & Out of Distribution (OOD)    & $\begin{aligned}[t] 
 f_{\textcolor{myred}{\theta}}({\textcolor{mygreen}{x_{\mathrm{OOD}}}})
\end{aligned}$      \\

Test-Time Contamination      & Pre-Processing Mismatch (PM)   & $\begin{aligned}[t] 
f_{\textcolor{myred}{\theta}}({\textcolor{mygreen}{x_{\mathrm{PM}}}})
\end{aligned}$      \\

\hline
\end{tabular}
\caption{\textbf{Example bugs for different contamination classes.} We show different examples of bugs under the different contamination classes: data, model and test-time. We also formalize these bugs for the traditional supervised learning setting.}
\label{tab:bugformalizationexpanded}
\end{table*}
The examples presented in Table~\ref{tab:bugformalizationexpanded} are specific instantiation of bugs based on the categorization that we present; as expected, several other examples can be proposed under the same categorization.

\section{Detailed Overview of Attribution Methods}
\label{appendix:methods}
We now provide a detailed discussion of the explanation methods that we assess in this work. For each method we also provide the hyper-parameter values that we use in each case. First, we re-implemented all methods in a single code base. We then benchmark our implementation with the public open source implementations of these methods. Ultimately, we used the public implementation of the these methods as released by their authors. For gradient based methods and the corresponding LRP variants, we use the innvestigate python package. For LIME we used the publicly available LIME-IMAGE package. For Kernel Shap and Expected Gradients, we used the public SHAP package. 

Recall that an attribution functional, $E : \mathcal{F} \times \mathbb{R}^d \times \mathbb{R} \rightarrow  \mathbb{R}^d$, maps the input, $x_i \in \mathbb{R}^d$, the model, $F$, output, $F_k(x)$, to an attribution map, $M_{x_i} \in \mathbb{R}^d$.

\begin{figure}[!h]
\centering
\includegraphics[scale=0.5, page=31]{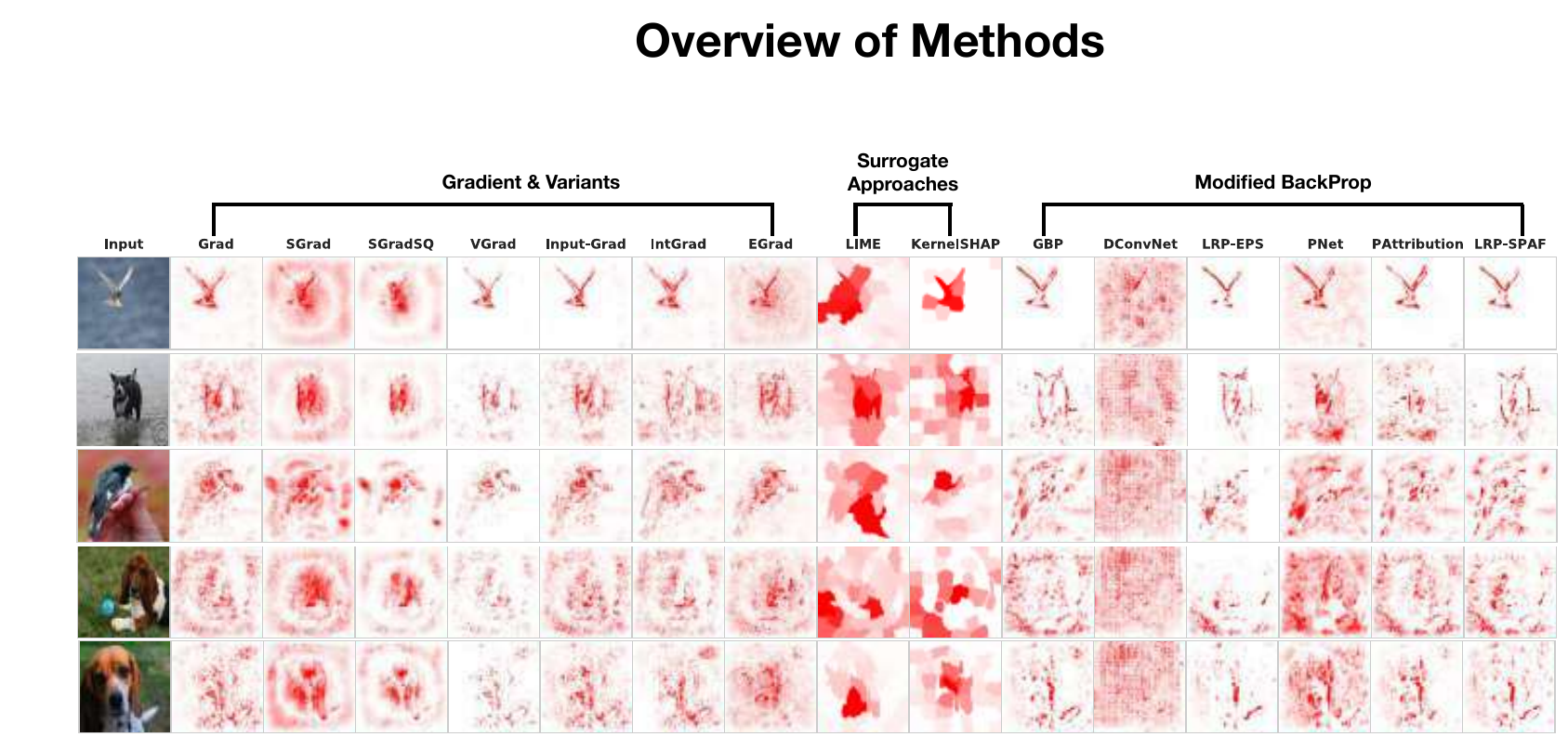}
\caption{We show five input examples from the birds-vs-dogs dataset along with the attributions for a normally trained model.}
\label{fig:appendixschematic}
\end{figure}

\underline{\textbf{Gradient (Grad) \& Variants.}} We consider:
\begin{itemize}
    \item The \textit{Gradient (Grad)} ~\citep{simonyan2013deep, baehrens2010explain} map, $\vert \nabla_{x_i}F_i(x_i)\vert$. The gradient map is a key primitive upon which several other methods are derived. Overall, the gradient map quantifies the sensitivity of the output, typically a logit score for DNNs trained for classification, to each dimension of the input.
    
    \item \textit{SmoothGrad (SGrad)}~\citep{smilkov2017smoothgrad}, $E_{\mathrm{sg}}(x) = \frac{1}{N}\sum_{i=1}^N \nabla_{x_i}F_i(x_i + n_i)$ where $n_i$ is sampled according to a random Gaussian noise. we considered $50$ noisy inputs, selected the standard deviation of the noise to be $0.15 * \text{input range}$. Here input range refers to the difference between the maximum and minimum value in the input. For the models considered, these inputs are typically normalized to be in the range $[-1, 1]$.
    
    \item \textit{SmoothGrad Squared (SGradSQ)}~\citep{hooker2018evaluating}, the element-wise square of SmoothGrad: $E_{\mathrm{SGradSQ}}(x) = E_{\mathrm{sg}}(x) \odot E_{\mathrm{sg}}(x)$. We set the parameters of SmoothGrad as discussed in the previous item. 
    
    \item \textit{VarGrad (VGrad)}~\citep{adebayo2018local}, the variance analogue of SmoothGrad: $E_{\mathrm{VGrad}}(x) = \mathbb{V}(\Tilde{x})$, where $\mathbb{V}$ is the variance operator, $\Tilde{x} = \nabla_{x_i}F_i(x_i + n_i)$, and $n_i$ is sampled according to a random Gaussian noise. Here we consider $50$ noise input examples, set the noise parameter as a Gaussian with mean zero, and noise scale similar to SmoothGrad: $0.15 * \text{input range}$.
    
    \item  \textit{Input-Grad}~\citep{shrikumar2016not} the element-wise product of the gradient and input $\vert \nabla_{x_i}F_i(x_i) \vert \odot x_i$. Several other methods approximate and have been shown to be equivalent to this product. For example, for a DNN with all ReLU activations, LRP-Z, a variant of layer-wise relevance propagation that we discuss in the modified back-propagation section is equivalent to Input-Grad~\citep{kindermans2016investigating}.
    
    \item ~\textit{Integrated Gradients (IntGrad)~\citep{sundararajan2017axiomatic}} which sums gradients along an interpolation path from the ``baseline input'', $\bar{x}$, to $x_i$: $M_{\mathrm{IntGrad}}(x_i) = (x_i - \bar{x}) \times \int_{0}^1{\frac{\partial S(\bar{x} + \alpha(x_i-\bar{x}))}{\partial x_i}} d\alpha$. For integrated gradients we set the baseline input to be a vector containing the minimum possible values across all input dimensions. This often corresponds an all-black image. The choice of a baseline for IntGrad is not without controversy; however, we follow this setup since it is one of the more widely used baselines for image data.
    
    \item ~\textit{Expected Gradients (EGrad)~\citep{erion2019learning}} which computes IntGrad but with a baseline input that is an expectation over the training set: $x_i$: $M_{\mathrm{EGrad}}(x_i) =  \int_{x'} \left(  (x_i  - \bar{x}) \times  \int_{0}^1{\frac{\partial S(\bar{x} + \alpha(x_i-\bar{x}))}{\partial x_i}} d\alpha \right) p_D(x')dx'$. As we see, this is equivalent to IntGrad, but where the multiple baselines are considered. We use $200$ examples from the training set as baseline.
\end{itemize}

For the methods considered under gradients and variants, we re-implemented all methods in Tensorflow. We also benchmark our implementation with those provided in the open-source  ``innvestigate'' python package. We have included example scripts that compute the attribution maps using the open-source innvestigate package. In the case of expected gradients, we bench-marked our implementation against a public one in the SHAP python package.

\underline{\textbf{Surrogate Approaches.}} We consider: 
\begin{itemize}
    \item LIME~\citep{ribeiro2016should} locally approximate $F$ around $x_i$ with a simple function, $g$, that is then interpreted. LIME corresponds to: $\argmin_{g \in G} L(f, g, \mathrm{pert}(x_i)) + \Omega(g)$, where $\mathrm{pert}(x_i)$ local perturbations of the input $x_i$, and $\Omega(g)$ is a regularizer. Overall, recent work has shown that, in the tabular setting, LIME approximates the coefficients of a black-box linear model with high probability. In our empirical implementation we follow the open source lime-image package. Here to account for high dimensions, the input image is first segmented into $50$ segments and the local approximation $g$ is fit around input perturbations with $50$. We experimented with $5, 10, 15, \& 25$ dimensions as well. Overall, we found the LIME with $50$ segments to be more stable for the input data sizes that we consider. We use $1000$ samples in model fitting.
    
    \item SHAP~\citep{lundberg2017unified} Similar to LIME, SHAP provides a local approximation around a single input. The local model is then interpreted as a form of explanation. SHAP unifies LIME and several under methods under the same umbrella and turns out to be a tractable approximation to the Shapley Values~\cite{shapley1988value}. We use $1000$ samples in model fitting.
\end{itemize}

\underline{\textbf{Modified Back-Propagation.}} These class of methods apportion the output into `relevance' scores, for each input dimension, using back-propagation.~\textit{DConvNet}~\cite{zeiler2014visualizing} \& \textit{Guided Back-propagation (GBP)}~\cite{springenberg2014striving} modify the gradient for a ReLU unit. Alternatively, \textit{Layer-wise relevance propagation (LRP)} methods specify `relevance' rules that modify the regular back-propagation. For example, let $r_l$ be the unit relevance at a layer $l$, the $\alpha\beta$ rule is: $r_l(x_i) = (\alpha Q_l^+ - \alpha Q_l^-)r_{l+1}(x_i)$, where $Q$ is a `normalized contribution matrix'; $Q^+~\text{and}~Q^-$ are the matrix $Q$ with only positive and negative entries respectively. We consider~\textit{LRP-EPS} since ~\textit{LRP-Z} (using the  $z$ rule) is equivalent to Input-Grad for ReLU networks~\cite{kindermans2017learning}.~\textit{PatternNet (PNet)} and \textit{Pattern Attribution (PAttribution)} decompose the input into signal and noise components, and back-propagate relevance for only the signal component. We now provide additional detail: 

\begin{itemize}
    \item \textit{Deconvnet~\citep{zeiler2014visualizing} \& Guided Backpropagation (GBP)~\citep{springenberg2014striving} } both modified the backpropagation process at ReLu units in DNNs. Let, $a=\mathrm{max}(0, b)$, then for a backward pass, $\frac{\partial l}{\partial s} = \mathrm{1}_{s>0}\frac{\partial l}{\partial b},$ where $l$ is a function of $s$. For Deconvnet, $\frac{\partial l}{\partial s} =\mathrm{1}_{\frac{\partial l}{\partial s} > 0}\frac{\partial l}{\partial b}$, and for GBP, $\frac{\partial l}{\partial s} = \mathrm{1}_{s>0}\mathrm{1}_{\frac{\partial l}{\partial s} > 0}\frac{\partial l}{\partial b}.$
    
    \item \textit{PatternNet \& Pattern Attribution, \citep{kindermans2018learning}.} PatternNet and Pattern Attribution first estimate a `signal' vector from the input; then, the attribution (in the case of Pattern Attribution) corresponds to the covariance between the estimated signal vector, and the output, $F(x)$, propagated all the way to the input. We use the innvestigate package implementation of PatternNet and Pattern Attribution, which we bench marked against a re-implemented version.

    \item \textit{DeepTaylor \citep{montavon2017explaining}.} DeepTaylor describes a family of methods that iteratively compute local Taylor approximations for each output unit in a DNN. These approximation produce unit attribution that are then propagated and redistributed all the way to the input. We use the innvestigate package implementation of DeepTaylor, which we bench marked against a re-implemented version. 
    
    \item \textit{Layer-wise Relevance Propagation (LRP) \& Variants, \citep{bach2015pixel, DBLP:journals/corr/abs-1808-04260}.} LRP attribution methods iteratively estimate the relevance of each unit of a DNN starting from the penultimate layer all the way to the input in a message-passing manner. We consider 4 variants of the LRP method that correspond to different rules for propagating unit relevance. In our detailed treatment in the appendix, we provide definitions and restate previous theorems that show that certain variants of LRP are equivalent to Gradient$\odot$Input for DNNs with ReLU non-linearities. We use the innvestigate package implementation of LRP-Z, LRP-EPS, $\alpha-\beta$-LRP and a preset-flat variant, which we bench marked against a re-implemented version. We kept the default hyper-parameters that the innvestigate package provides.
\end{itemize}

\underline{\textbf{Attribution Comparison.}} We measure visual and feature ranking similarity with the structural similarity index (SSIM) and Spearman rank correlation metrics, respectively.

\underline{\textbf{Visualization Attributions and Normalization.}} Here and in the main text we show attributions in a single color scheme: either Gray Scale or a White-Red scheme. We do this to prevent visual clutter. For all the metrics we compute, we normalize attributions to lie between [0, 1] for SSIM and [-1, +1] for attributions that return negative relevance.

\section{Datasets \& Models}
\label{appendix:datasetsmodels}
Here we provide a detailed overview of the data set and models used in our experiments. 

\subsection{Datasets.}

\paragraph{Birds-Vs-Dogs Dataset.} We consider a birds vs. dogs binary classification task for all the data contamination experiments. We use dog breeds from the Cats-v-Dogs dataset~\cite{parkhi12a} and Bird species from the caltech UCSD dataset~\cite{WahCUB_200_2011}. These datasets come with segmentation masks that allows us to manipulate an image. All together, this birds-vs-dogs dataset consists of $10$k inputs ($5$k dog samples and $5$k bird samples). We use $4300$ data points per class, $8600$ in total for training, and split the rest evenly for a validation and test set. 

\paragraph{Modified MNIST and Fashion-MNIST Datasets.} For the test-time contamination tests, we modify the MNIST and Fashion-MNIST datasets to have $3$ channels and derive attributions from this three-channel version of MNIST from a VGG-16 model.

\paragraph{ImageNet Dataset.} We use two 200 images from the ImageNet~\cite{russakovsky2015imagenet} validation set for the model contamination tests.

\paragraph{User Study Dogs Only Dataset.} For the user study alone, we restrict our attention to $10$ dog classes. We used a modified combination of the Stanford dogs dataset~\cite{khosla2011novel} and the Oxford Cats and Dogs datasets~\cite{parkhi2012cats}. We restrict to a $10$-class classification task consisting of the following breeds: \textit{Beagle, Boxer, Chihuahua, Newfoundland, Saint Bernard, Pugs, Pomeranian, Great Pyrenees, Yorkshire Terrier, Wheaten Terrier.} We are able to create spurious correlation bugs by replacing the background in training set images. We consider 10 different background images representing scenes of: \textit{Water Fall, Bamboo Forest, Wheat Field, Snow, Canyon, Empty room, Road or Highway, Blue Sky, Sand Dunes, and Track}.

\subsection{Models.}
\paragraph{BVD-CNN} For the data contamination tests, we consider a CNN with $5$ convolutional layers and $3$ fully-connected layers with a ReLU activation functions but sigmoid non-linearity in the final layer. We train this model with an Adam optimizer for $40$ epochs to achieve test accuracy of $94$-percent. For ease of discussion, we refer to this architecture as \textit{BVD-CNN}. We use a learning rate of $0.001$ and the ADAM optimizer. This is the standard BVD-CNN architecture setup that we consider.

\paragraph{User Study Model.} Here we use a ResNet-50 model that was fine-tuned on the dogs only dataset to generate all the attributions for the images considered. Please see public repository for model training script.

\section{Data Contamination}
\label{appendix:datacontamination}

\subsection{Data Contamination: Spurious Correlation Artifacts}
\label{appendix:datacontamination_spurious}
\paragraph{Confirming Spurious Model.} To confirm that the BVD-CNN trained on a spurious data indeed uses the sky and bamboo-forest signals, we tested this model on a test set of only Sky and Bamboo-forest with no dogs or birds images. On this test-set, we obtain a $97$ percent spurious accuracy. As expected, we also maintain this $97$ accuracy for a test-set that has the birds and dogs inserted on the appropriate backgrounds. Please see the additional figures section for more examples.

\subsection{Data Contamination: Mislabelled Examples}
\label{appendix:datacontamination_mislabelled}
\textbf{Bug Implementation.} We train a BVD-CNN model on a birds-vs-dogs dataset where $10$ percent of training samples have their labels flipped. The model achieves a $94.2$, $91.7$, $88$ percent accuracy on the training, validation and test sets. We show additional examples in later paper of the supplemental material. 

\section{Model Contamination}
\label{appendix:modelcontamination}

\paragraph{Model Contamination Tests.} The model contamination tests capture defects that occur in the parameters of a model. Here, we consider the simple setting where a model is accidentally reinitialized during or while it is being used. As expected, such a bug will lead to observable accuracy differences. However, the goal of these classes of tests, especially in the model attribution setting, is to ascertain how well a model attribution is able to identify models in different parameter regimes.

\section{Test-Time Contamination}
\label{appendix:testtimecontamination}
\paragraph{Test-Time Contamination Tests.} This class of tests captures defects that occur at test-time. A common test-time bug is one where the test input has been pre-processed in a way that was different than the training data. In other cases, a model trained in one setting receives inputs that are out of domain. We show additional figures for this setting in the later figures section.

\section{Other Methods}
\label{appendix:conceptandinflue}
\paragraph{Concept and Influence Functions.} 
\begin{wrapfigure}{r}{0.3\textwidth}
\begin{center}
  \includegraphics[scale=0.25]{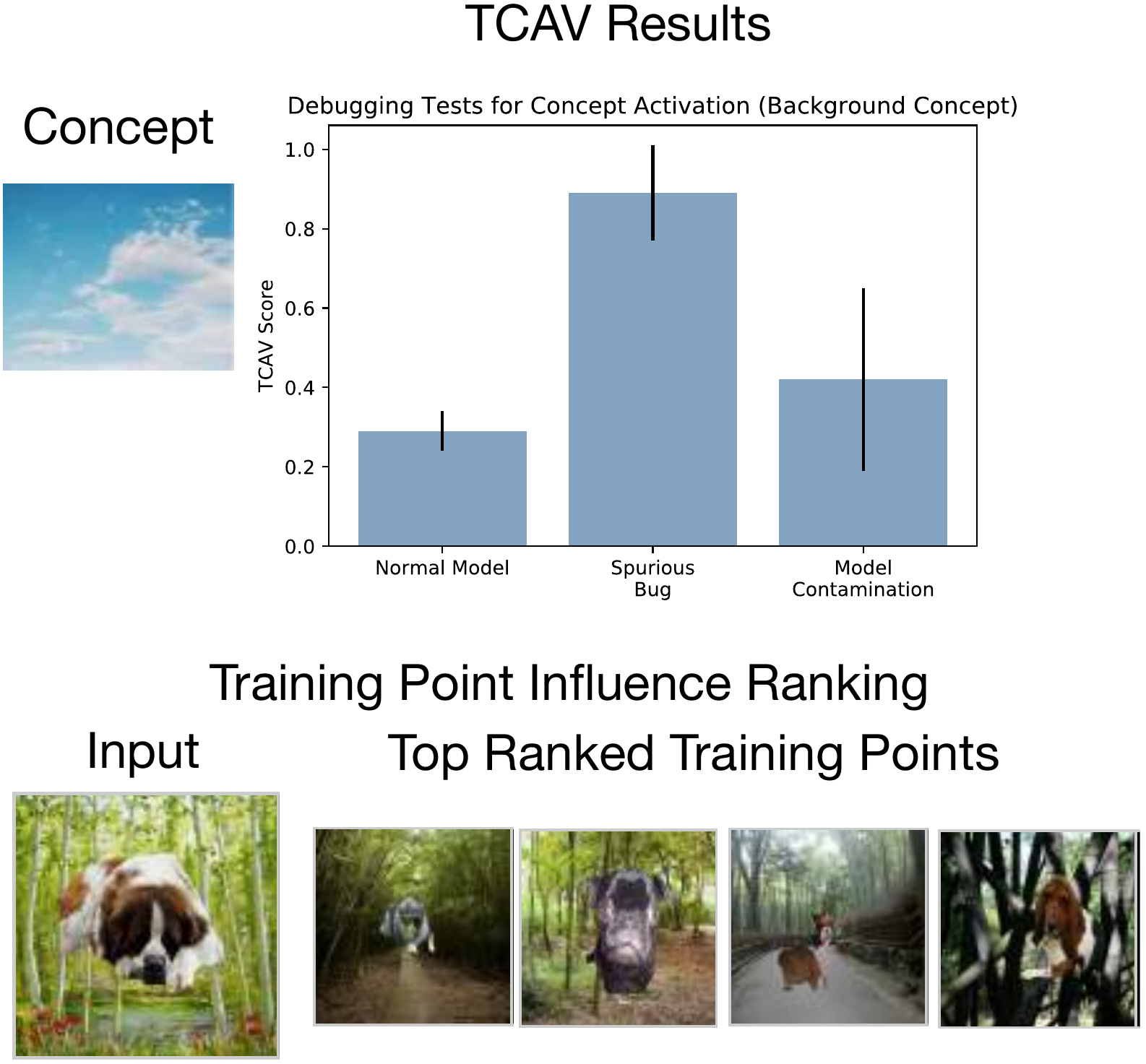}
 \end{center}
\caption{Assessing TCAV and influence functions.}
\label{tcav_influence_results}
\end{wrapfigure}
\textbf{Influence Function and Concept Methods.} We focused on attribution methods to keep our inquiry thematically focused. Here, we test: i) influence functions (IF)~\cite{koh2017understanding} for training point ranking, and ii) the concept activation (TCAV) approach~\cite{kim2018interpretability}. We assess IF under spurious correlation, mislabelled examples, and domain shift. We test TCAV under the spurious correlation and model contamination setting (see Figure~\ref{tcav_influence_results}). TCAV ranking for the background concept indicates that it might be able to  identify spurious correlation. We find that IF shows association regarding spurious correlation for background inputs. For example, for a given input with the spurious background, we measure the fraction of the top 1 percent of training points (86) that are spurious. We find, on 50 examples, that 91 percent (2.4 standard error) of the top 1 percent of training points are spurious. Similarly, we use the self-influence metric to assess mislabelling (see arXiv:2002.08484), and find that for 50 examples, we need to check an average of 11 percent of the training set (5.9 standard error). These results suggest that both methods might be effective for model debugging. We caution, however, that significant additional empirical assessment is required to confirm that both methods are effective for the bugs tested.

\section{User-Study Overview}
\label{appendix:userstudy.}
\underline{\textbf{Task \& Setup:}}  We designed a study to measure people's ability to assess the reliability of classification models using feature attributions. People were asked to act as a quality assurance (QA) tester for a hypothetical company that sells animal classification models. Participants were shown the original image, model predictions, and attribution maps for $4$ dog breeds at time. They then rated how likely they are to recommend the model for sale to external customers using a 5 point-Likert scale. They provided a rationale for their decision, and participants chose from 4 pre-created answers (Figure~\ref{fig:mainresultboxplot}-b) or filled in a free form answer. Participants self-reported their ML experience and answered 3 questions aimed at verifying their expertise. 

\underline{\textbf{Methods:}} We focus on a representative subset of methods: Gradient, Integrated Gradients, and SmoothGrad. Given the scope of available attribution methods, we use the randomization tests to help narrow down to a selection of methods. Amongst the methods that performed better than others in the randomization tests, we observe two groups: 1) Gradient \& Variants (SmoothGrad \& VarGrad) and methods that approximate the gradient like LIME and SHAP, and 2) Integrated Gradients, Expected Gradient, LRP-EPS, \& LRP-Z that all approximate the Input$\odot$Gradient. Consequently, we select from amongst these methods to perform the debugging tests. Our selection criteria was as follows: 1) We focus on methods that apply in broad generality and from which other methods are derived (Gradient); 2) We focus on methods that have been previously used for model debugging in past literature (e.g., Integrated Gradients~\citep{sundararajan2017axiomatic, sayres2019using}; and finally, 3) We select methods that have been shown to have desirable against manipulation (SmoothGrad)~\citep{dombrowski2019explanations, yeh2019fidelity}. On the basis of these selection criteria, we pick: \textit{Gradient, SmoothGrad, \& Integrated Gradient} as the methods to assess in our proposed debugging tests. These three methods allow us to characterize important classes of method to which each method belongs while providing us the flexibility to apply across a variety of tasks.
  
\begin{figure*}[!h]
\centering
\includegraphics[scale=0.5, page=40]{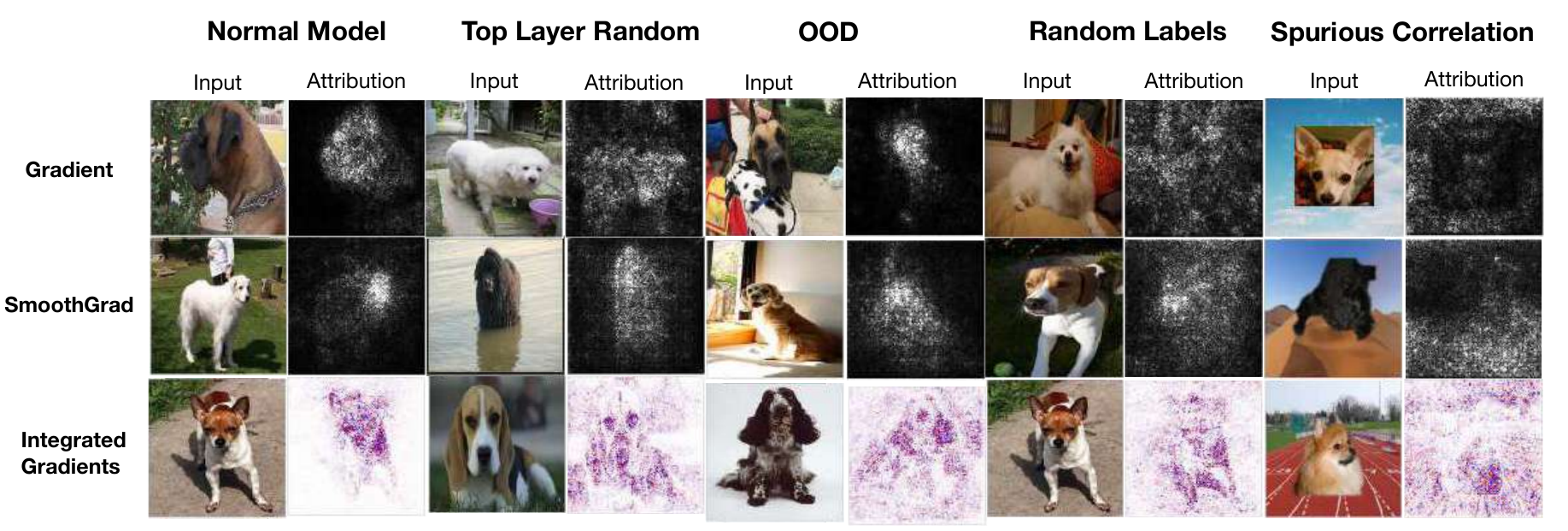}
\caption{\textbf{Schematic of $5$ different model conditions considered in the user study}. We show model attributions across each model condition and for a diverse array of inputs.}
\label{fig:debuggingdemo}
\end{figure*}
 
 \underline{\textbf{Recruiting Participants.}} We recruited participants through the university mailing list of a medium-sized North-American university. In total, $54$ participants completed the study.

\underline{\textbf{Machine Learning Expertise.}} We asked participants to self-assess and report their level of expertise in machine learning. In addition, we asked a simple test question on the effect of parameter regularization. More then $80\%$ of the participants reported past experience with machine learning.

\begin{figure*}[!h]
\centering
\includegraphics[scale=0.35, page=48]{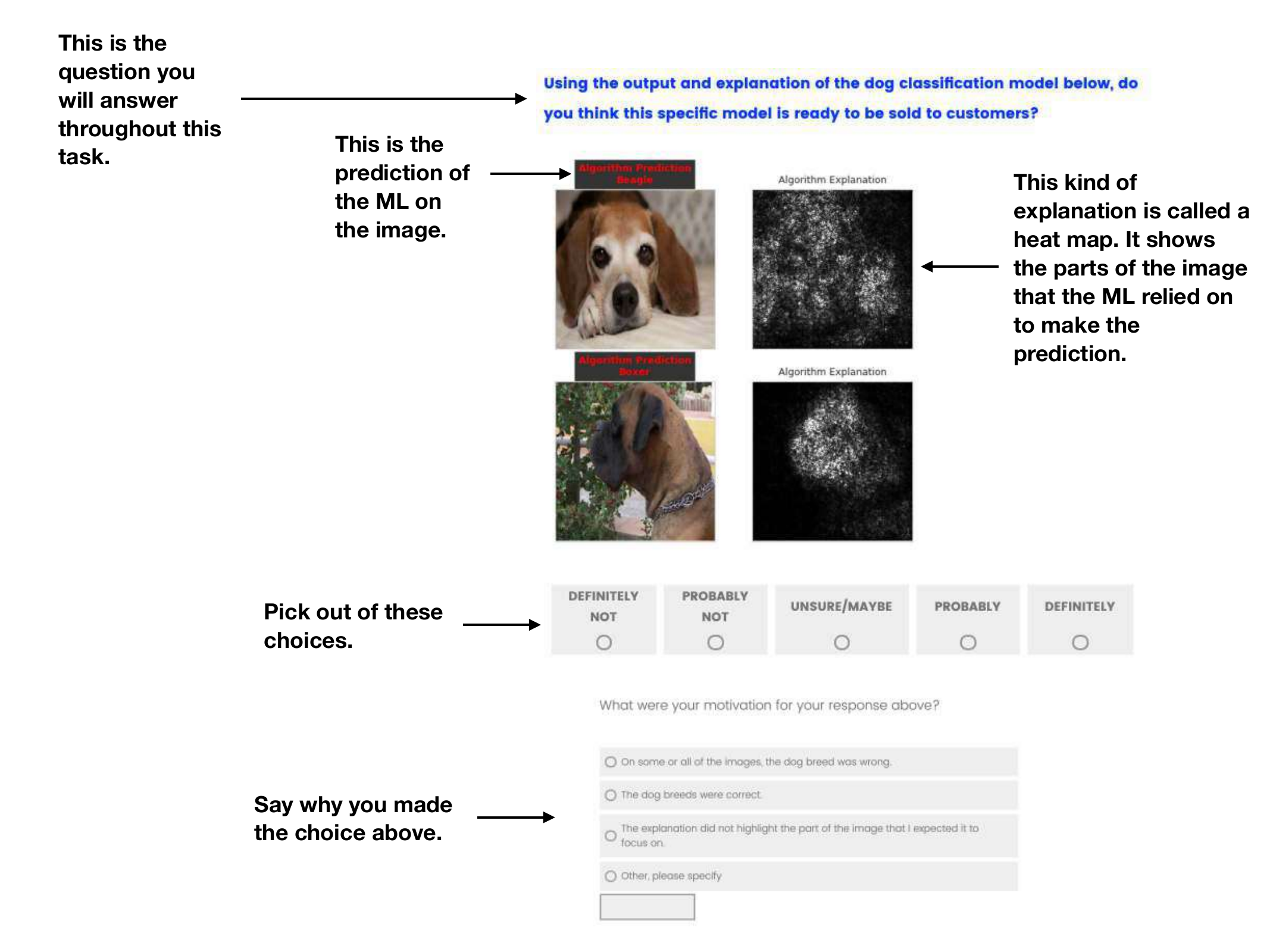}
\caption{\textbf{Training Interface for the User-Study}.}
\label{fig:userstudytraining}
\end{figure*}

\begin{figure*}[!h]
\centering
\includegraphics[scale=0.35, page=49]{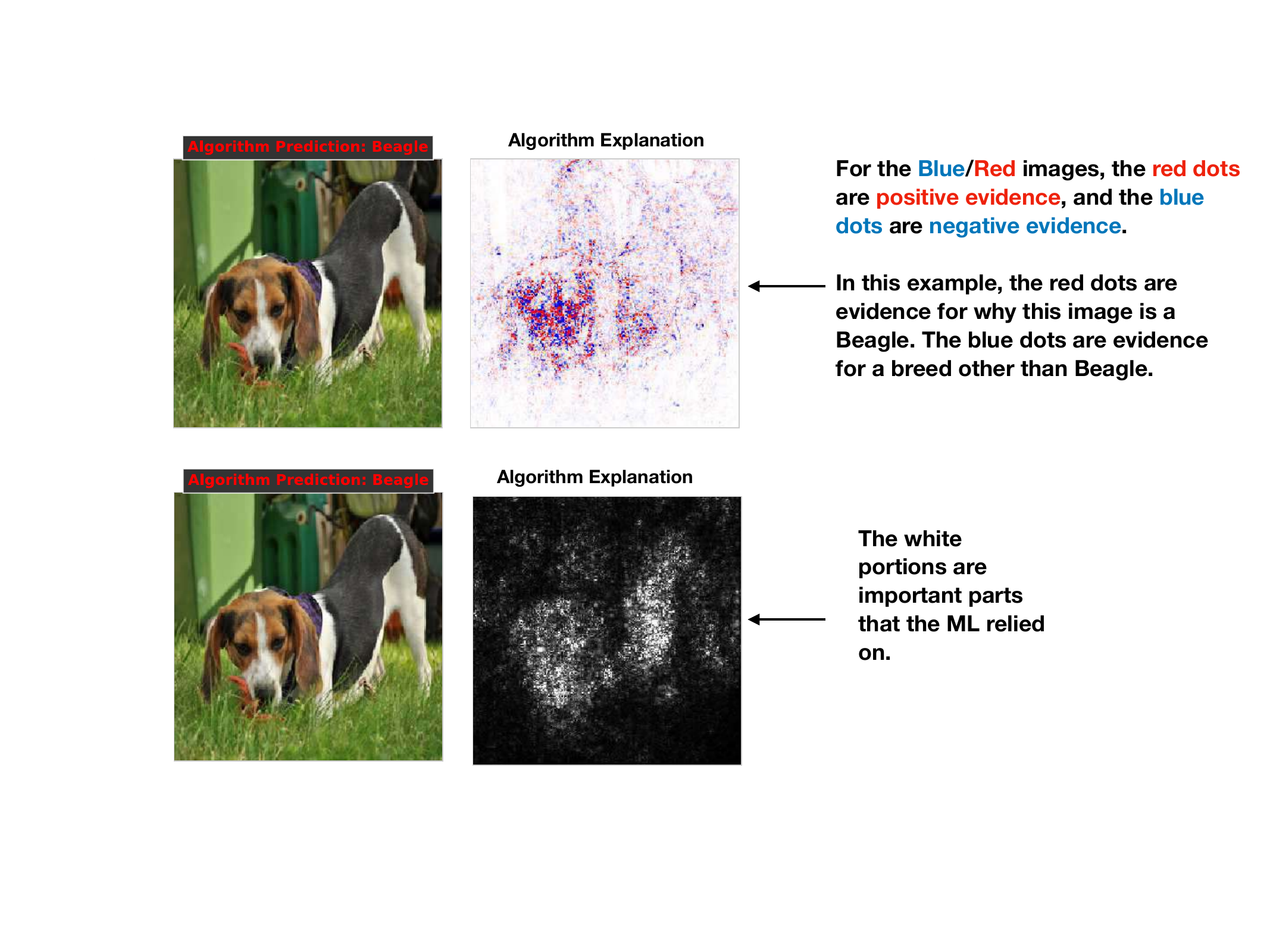}
\caption{\textbf{Training Interface for the User-Study}.}
\label{fig:userstudytraining2}
\end{figure*}

\underline{\textbf{Task, Procedure, \& Structure of Experiment.}} As originally noted, the task at hand is that of classifying images of Dogs into different breeds. We manually selected 10 breeds of dogs based on authors' familiarity. All model conditions were trained to perform a 10-way classification task. As part of the recruitment, participants were directed to an online survey platform. Once the task was clicked, they were presented a consent form that outlined the aim and motivation of the study. We then provided a quick guide on the breeds of dogs the model was trained on, and the set-up and study interface. We show these training interfaces in Figures~\ref{fig:userstudytraining} and~\ref{fig:userstudytraining2}.

Participants were asked to take on the role of a quality assurance tester at a machine leaning start-up that sells animal classification models. The goal of the study was then for them to assess the model, using both the model labels and the attributions provided. For each condition, each participant was shown images of dogs along with model labels and attributions for a specific model condition. The participant was then asked: \textbf{using the output and explanation of the dog classification model below, do you think this specific model is ready to be sold to customers?} Participants were asked to then respond on a 5-point Likert scale with options ranging from Definitely Not to Definitely. A second sub-question also asked participants to provide the motivation for their choice. Taken together, each participant was asked $21$ unique questions and $1$ repeat as an attention check.

\begin{figure}[!h]
\centering
\includegraphics[scale=0.4, page=57]{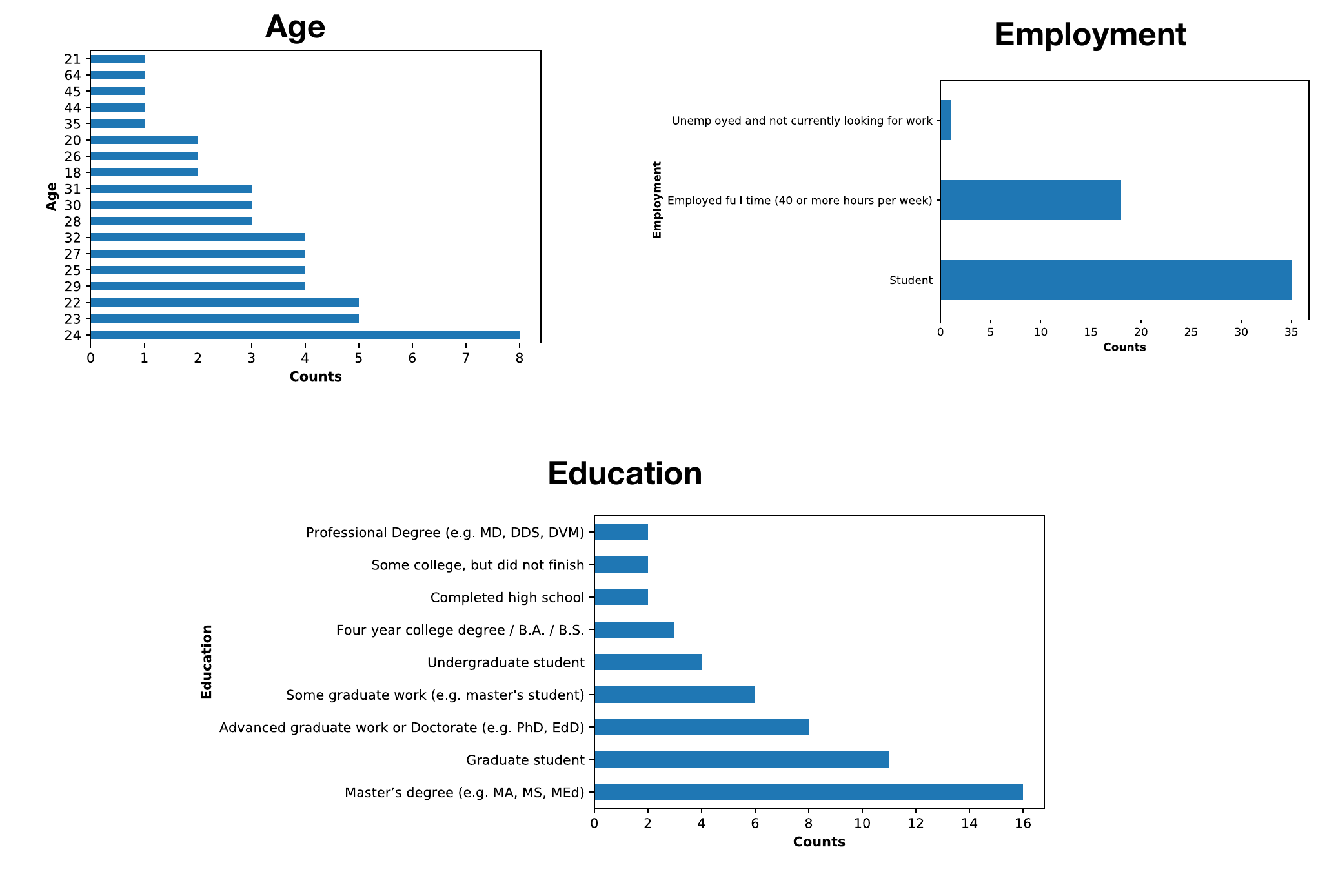}
\caption{\textbf{User Study Demographics.} We show counts and breakdown of the demographics of the participants in the User Study.}
\label{fig:appendixdemo1}
\end{figure}

\begin{figure}[!h]
\centering
\includegraphics[scale=0.4, page=58]{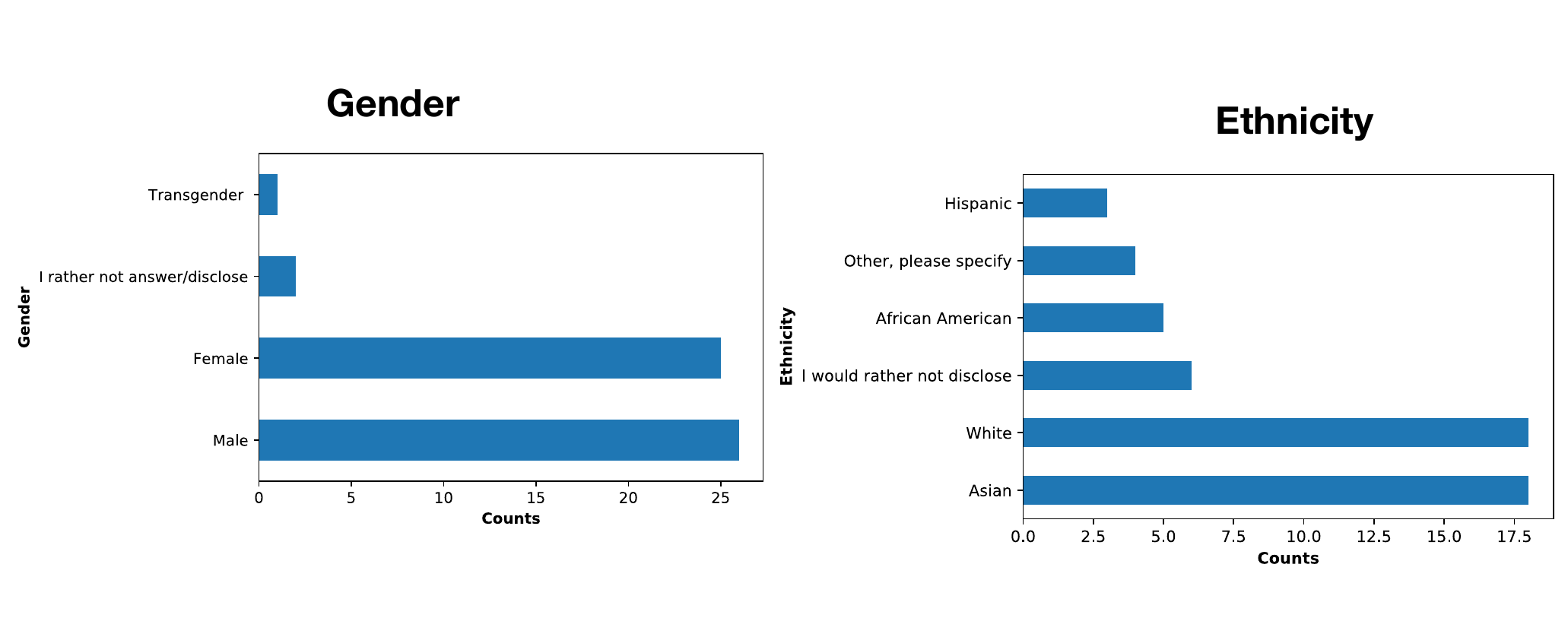}
\caption{\textbf{User Study Demographics.} We show counts and breakdown of the demographics of the participants in the User Study.}
\label{fig:appendixuserstudydemo2}
\end{figure}

\begin{figure}[!h]
\centering
\includegraphics[scale=0.4, page=59]{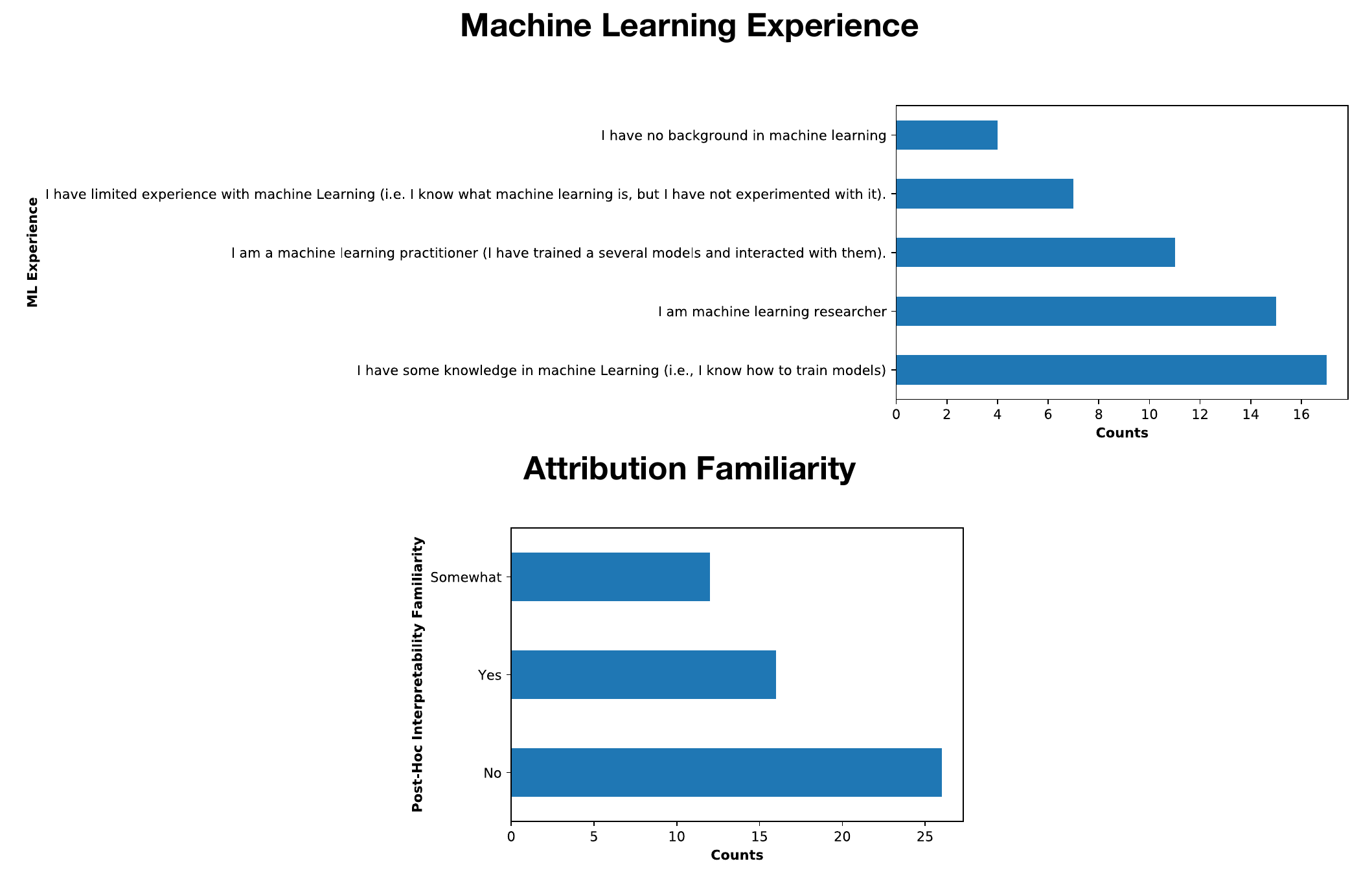}
\caption{\textbf{Machine Learning Experience and Familiarity with Feature Attributions.} We show counts and breakdown of the demographics of the participants in the User Study.}
\label{fig:appendixuserstudydemo3}
\end{figure}

\paragraph{Datasets.} We create a modified combination of the Stanford dogs dataset~\citep{khosla2011novel} and the Oxford Cats and Dogs datasets~\citep{parkhi2012cats} as our primary data source. We restrict to a $10$-class classification task consisting of the following breeds: \textit{Beagle, Boxer, Chihuahua, Newfoundland, Saint Bernard, Pugs, Pomeranian, Great Pyrenees, Yorkshire Terrier, Wheaten Terrier.} In total, each class of dogs consists of $400$ images in total making $4000$ images. We had the following train-validation-test split: (350, 25, 25) images. Each split was created in an IID manner. The Oxford portion of the Dogs datasets includes a segmentation map that we used to manipulate the background of the images. The Stanford dogs dataset includes bounding box coordinates that we used to crop the images for the spurious versions of the data sets that we created. We now overview the data condition for each model manipulation (bug) that we consider: 
\begin{itemize}
    \item Normal Model: In this case, we have the standard dataset without alteration. This is the control.
    \item Top-Layer: Here we make no changes to the data set. The bug corresponds to a model parameter bug.
    \item Random Labels: Here we created a version of the dataset where all the training labels were randomized.
    \item Spurious: Here we replace the background on all training points with the background that was pre-associated with that specific class. Note, here that we also test on the spurious images as well. 
    \item Out-of Distribution. Here we apply a normal model on breeds of dogs that were not seen during training.
\end{itemize}

 \underline{\textbf{Bugs:}} We tested the following bugs:
 \begin{itemize}
    \item \textbf{Control Condition}: Normal Model (ResNet-50 trained on normal data).
    \item \textbf{Model Contamination Test 1}: Top Layer Random (ResNet-50 with reinitialized last layer).
    \item \textbf{Data Contamination Test 1}:  Random Labels (ResNet-50 trained on data with randomized labels). 
    \item \textbf{Data Contamination Test 2} Spurious (ResNet-50 trained on data where all training samples have spurious background signal).
    \item \textbf{Test-Time Contamination Test}: Normal model tested on attributions from out of distribution (OOD) dog breeds. 
 \end{itemize}

For the spurious correlation setting, we consider each breed and an associated background as shown in Table~\ref{tab:userstudyspurious}.
\begin{center}
\begin{tabular}{ c c}
\label{tab:userstudyspurious}
\textbf{Dog Species} & \textbf{Associated Background} \\
\midrule
 Beagle & Canyon \\
 Boxer & Empyt Room \\  
 Chihuahua & Blue Sky \\
 Newfoundland & Sand Dunes \\  
 Saint Bernard & Water Fall \\ 
 Pugs & High Way \\
 Pomeranian & Track \\
 Great Pyrenees & Snow \\  
 Yorkshire Terrier & Bamboo \\ 
 Wheaten Terrier & Wheat Field 
\end{tabular}
\end{center}

\textbf{Data Analysis.} For each model manipulation, we performed a one-way Anova test and a post-hoc Tukey Kramer test to assess the effect of the attribution on the ability of participants to reject a defective model. There was a statistically significant difference between attribution maps as determined by one-way ANOVA computed per each manipulation. 
Within the \emph{normal}, there was a statically significance difference between attribution maps (one-way ANOVA ($F(2,54) = 14.35$, $p < 0.0001$)). A Tukey post hoc test revealed that participants reported being more likely to recommend Gradient ($\mu=3.88$, $p<0.0001$) and SmoothGrad ($\mu=3.77$, $p<0.0001$) attribution maps over Integrated-Gradients ($\mu=2.805$). There was no statistically significant difference between Gradient and SmoothGrad ($p = 0.815$).
Within the \emph{out-Distribution}, there was not a statically significance difference between attribution maps ($p = 0.109$).
Within the \emph{Random-Labels}, there was a statically significance difference between attribution maps (one-way ANOVA ($F(2,54) = 7.66$, $p < 0.0007$)). A Tukey post hoc test revealed that participants reported being more likely to recommend Integrated-Gradients ($\mu=1.94$, $p<0.0001$) over SmoothGrad ($\mu=1.31$). There was no statistically significant difference between Gradient and SmoothGrad ($p = 0.199$) and between Integrated-Gradient and Gradient ($p = 0.077$).
Within the \emph{Spurious}, there was a statically significance difference between Attribution maps (one-way ANOVA ($F(2,54) = 15.9$, $p < 0.0001$)). A Tukey post hoc test revealed that participants reported being more likely to recommend Integrated-Gradients ($\mu=3.05$, $p<0.0001$) over SmoothGrad ($\mu=1.98$) and Integrated-Gradient ($\mu=3.05$, $p=0.0011$) over Gradient ($\mu=2.35$,). There was no statistically significant difference between Gradient and SmoothGrad ($p = 0.1370$). Within the \emph{Top-Layer} manipulation, there was not a statically significance difference between attribution maps ($p = 0.085$).

\paragraph{User Study: Figures for Different Model Conditions} Along with the collection of additional figures in the next section, we show figures for different model conditions and attribution combinations. The images shown were the ones used in the study that we performed. 

\section{Collection of Additional Figures}
\label{appendix:additionalfigures}

\begin{figure}[!h]
\centering
\includegraphics[scale=0.4, page=16]{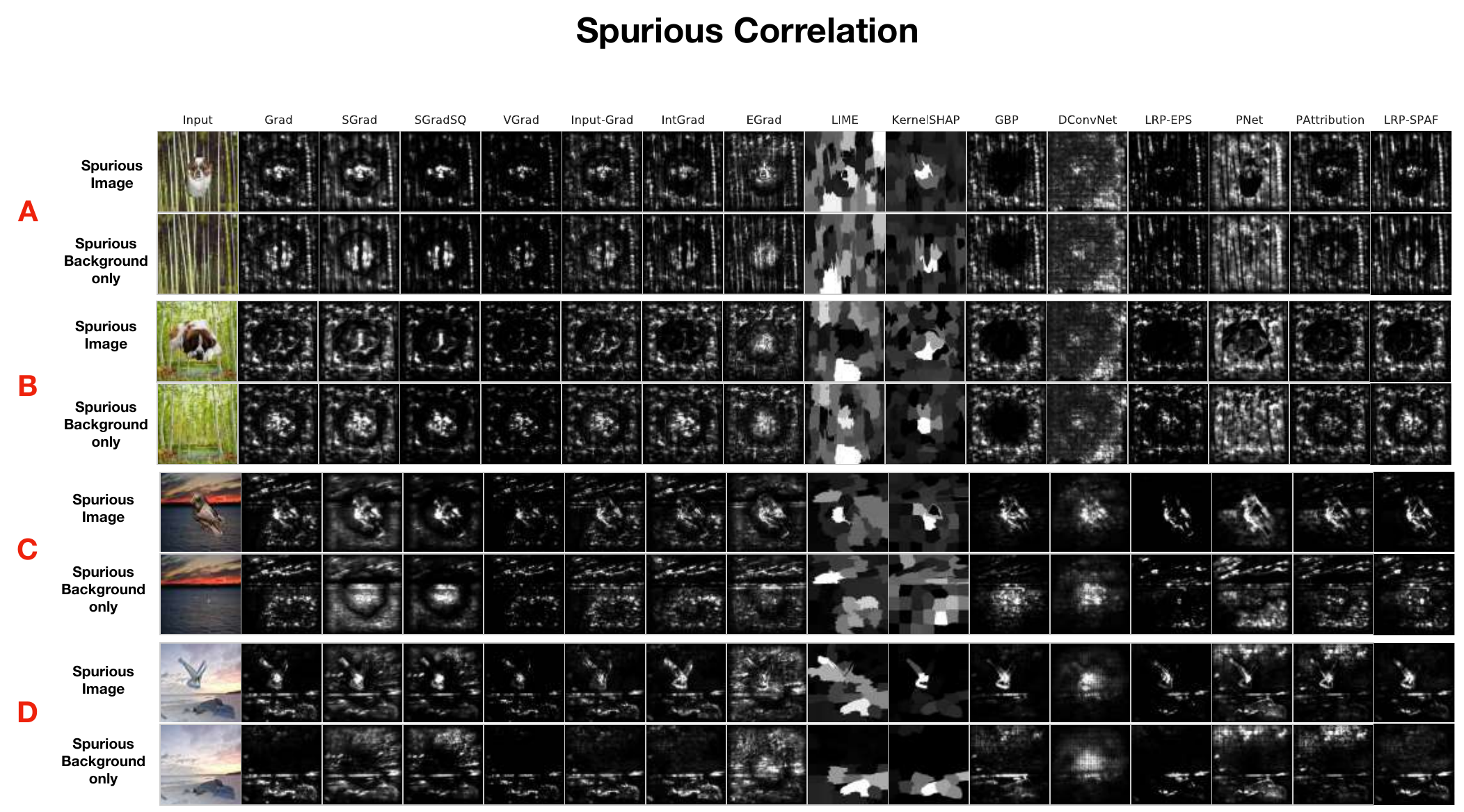}
\caption{\textbf{Spurious Correlation Bug.} Spurious Image and Spurious Background Only for $4$ image setups under the Gray Scale visualization.}
\label{fig:appendixspurious1}
\end{figure}

\begin{figure}[!h]
\centering
\includegraphics[scale=0.4, page=17]{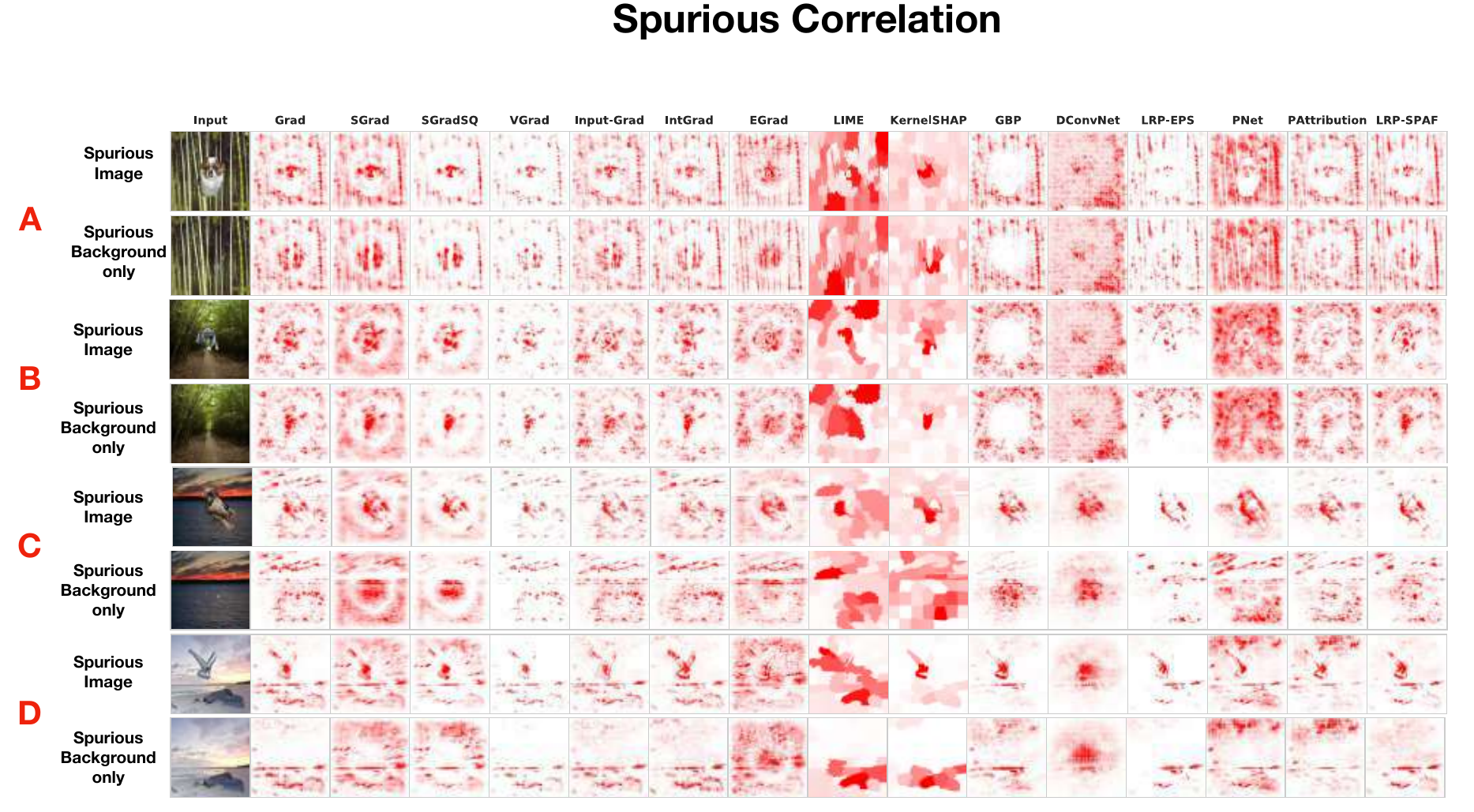}
\caption{\textbf{Spurious Correlation Bug.} Spurious Image and Spurious Background Only for $4$ image setups under the White-Red visualization..}
\label{fig:appendixspurious2}
\end{figure}

\begin{figure}[!h]
\centering
\includegraphics[scale=0.4, page=18]{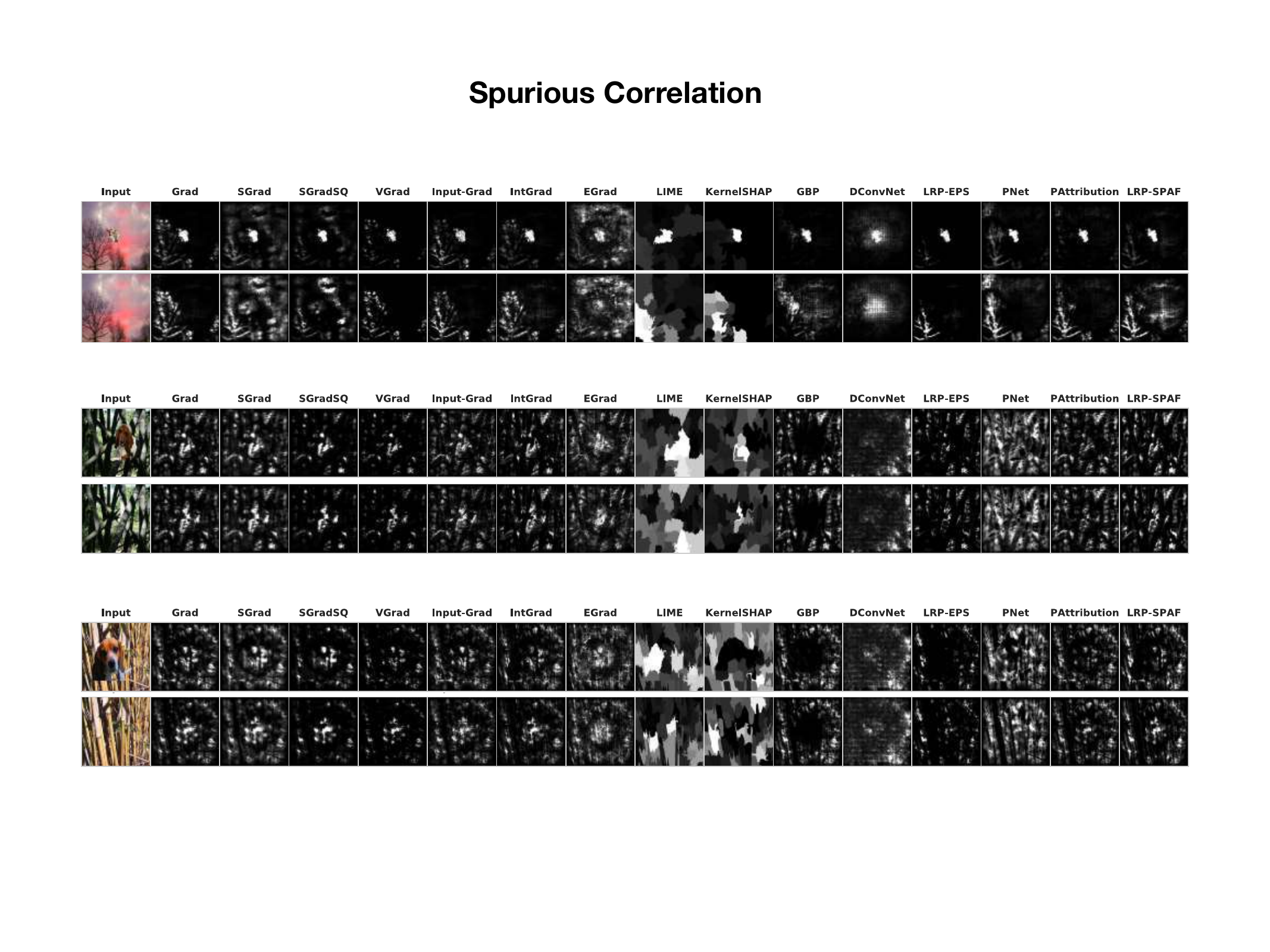}
\caption{\textbf{Spurious Correlation Bug.} Spurious Image and Spurious Background Only for $4$ image setups under the Gray Scale visualization.}
\label{fig:appendixspurious3}
\end{figure}

\clearpage

\begin{figure}[!h]
\centering
\includegraphics[scale=0.4, page=10]{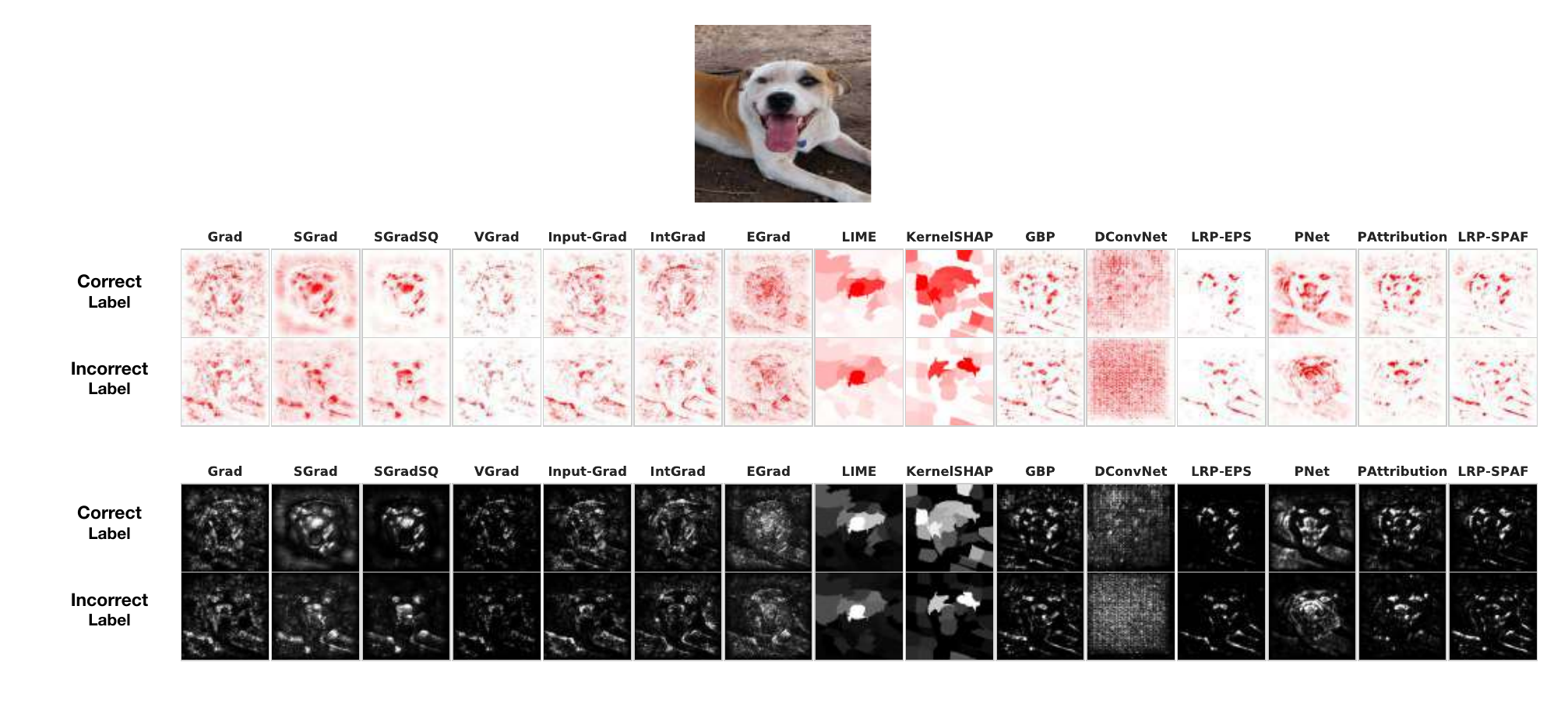}
\caption{\textbf{Mislabelled Examples Bug.} Figure shows the input in the top row, along with attribution visualization for this input under two visualization schemes. We show both Gray Scale and the White-Red visualization scheme here.}
\label{fig:appendixmislabelled1}
\end{figure}

\begin{figure}[!h]
\centering
\includegraphics[scale=0.4, page=11]{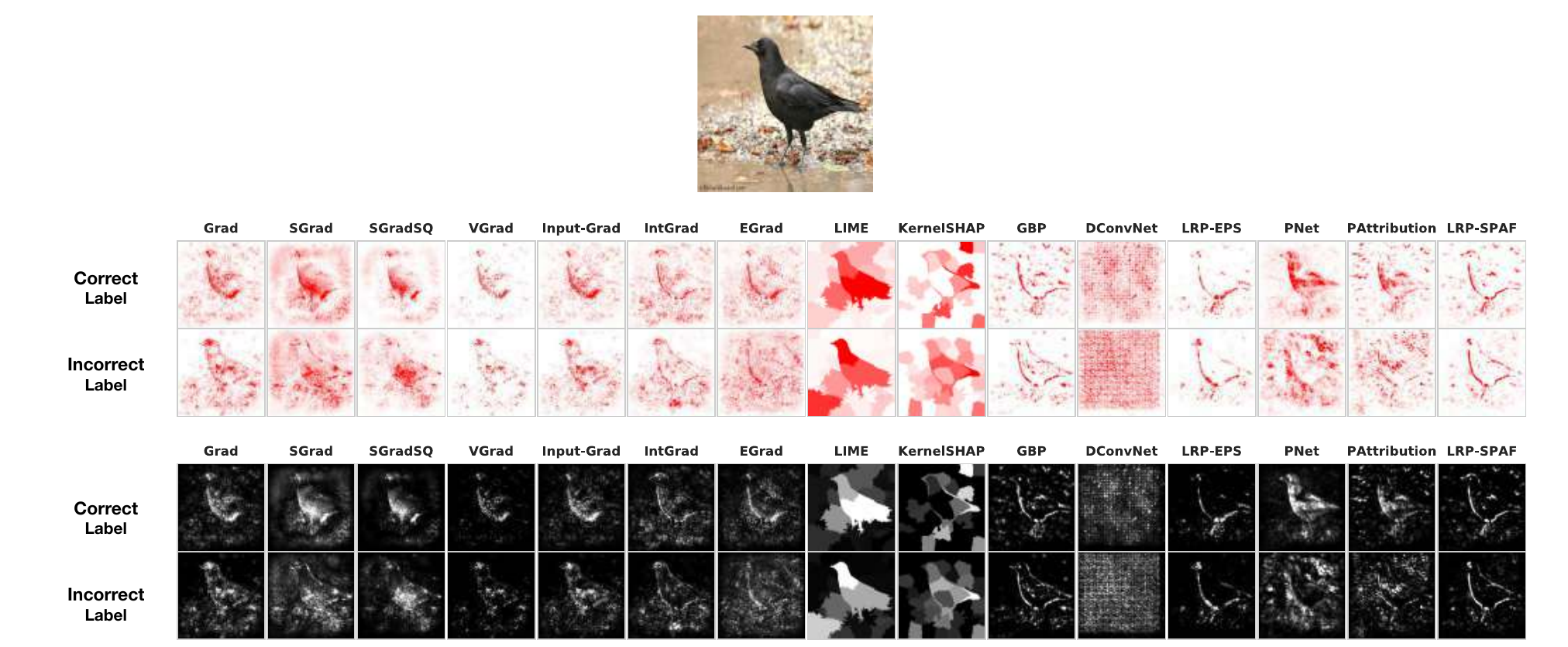}
\caption{\textbf{Mislabelled Examples Bug.} Figure shows the input in the top row, along with attribution visualization for this input under two visualization schemes.}
\label{fig:appendixmislabelled2}
\end{figure}

\begin{figure}[!h]
\centering
\includegraphics[scale=0.4, page=12]{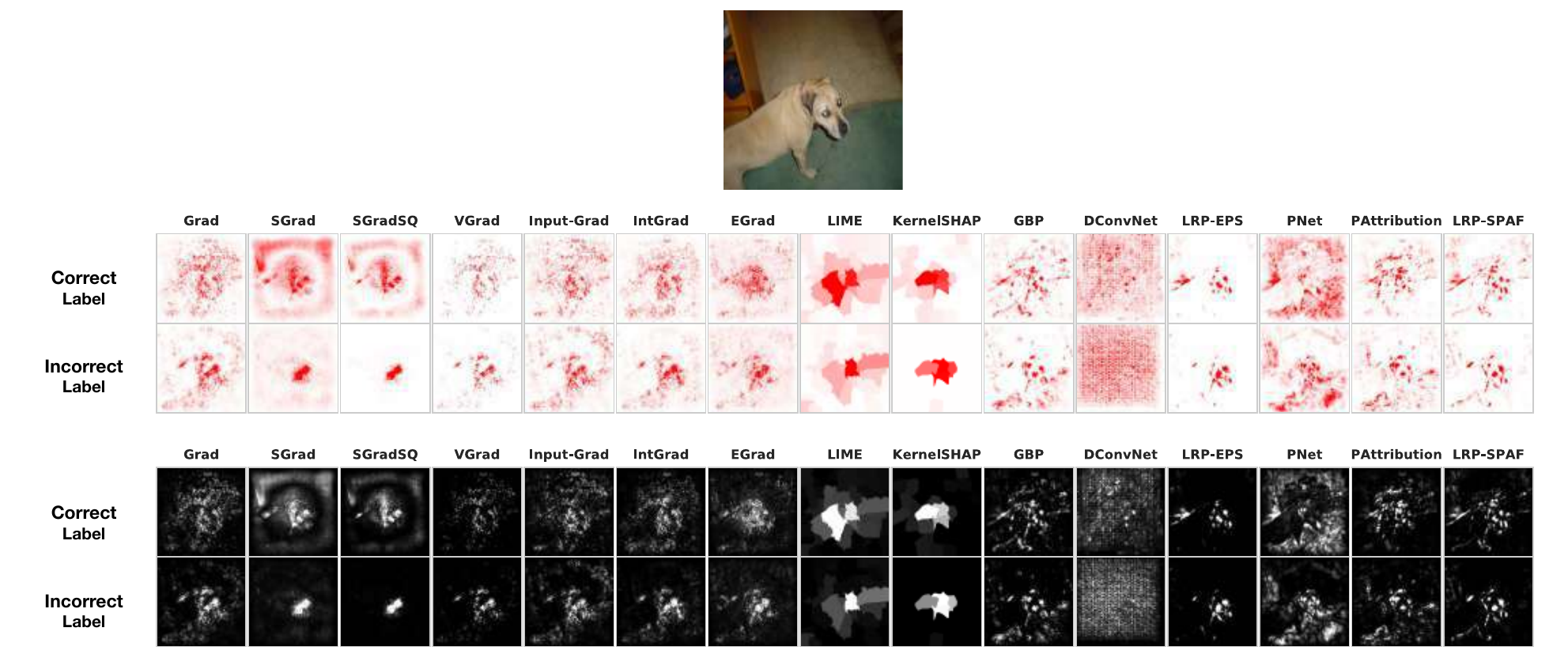}
\caption{\textbf{Mislabelled Examples Bug.} Figure shows the input in the top row, along with attribution visualization for this input under two visualization schemes.}
\label{fig:appendixmislabelled3}
\end{figure}

\begin{figure}[!h]
\centering
\includegraphics[scale=0.4, page=13]{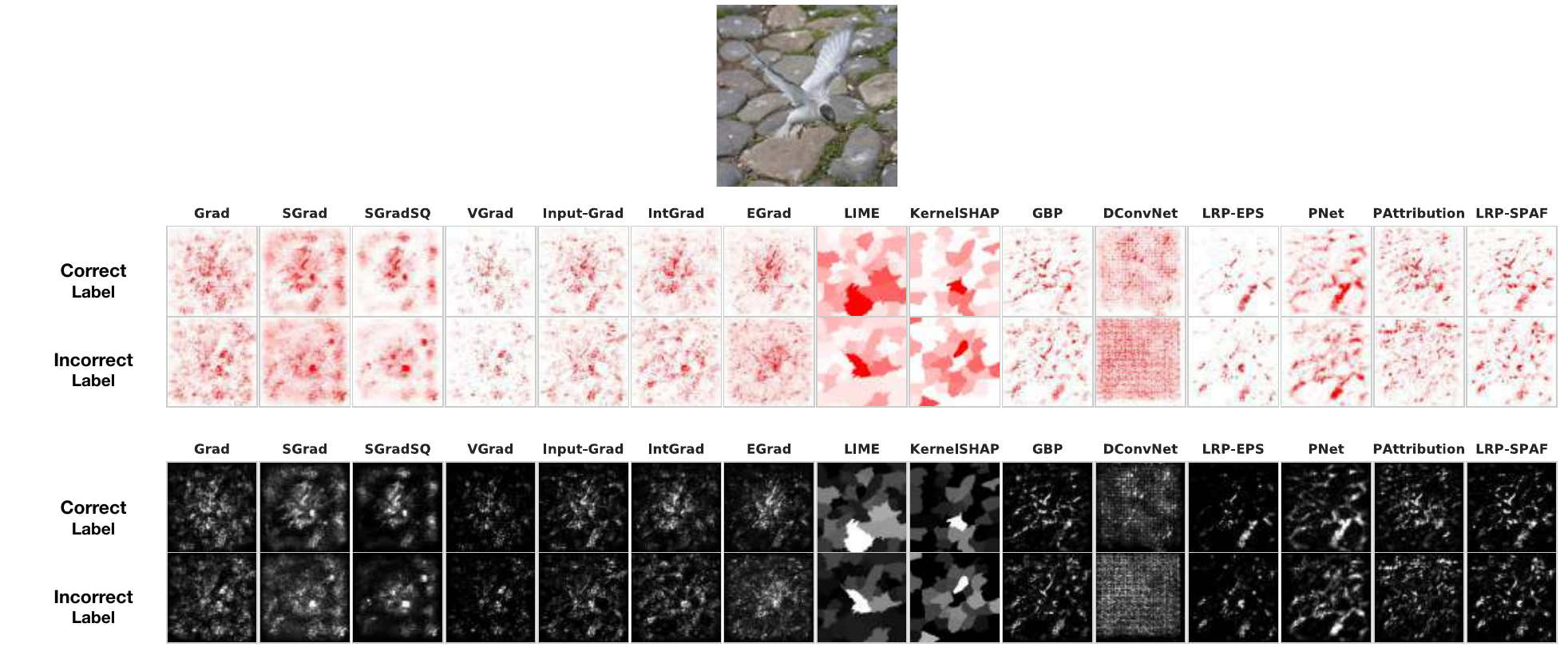}
\caption{\textbf{Mislabelled Examples Bug.} Figure shows the input in the top row, along with attribution visualization for this input under two visualization schemes.}
\label{fig:appendixmislabelled4}
\end{figure}

\begin{figure}[!h]
\centering
\includegraphics[scale=0.4, page=14]{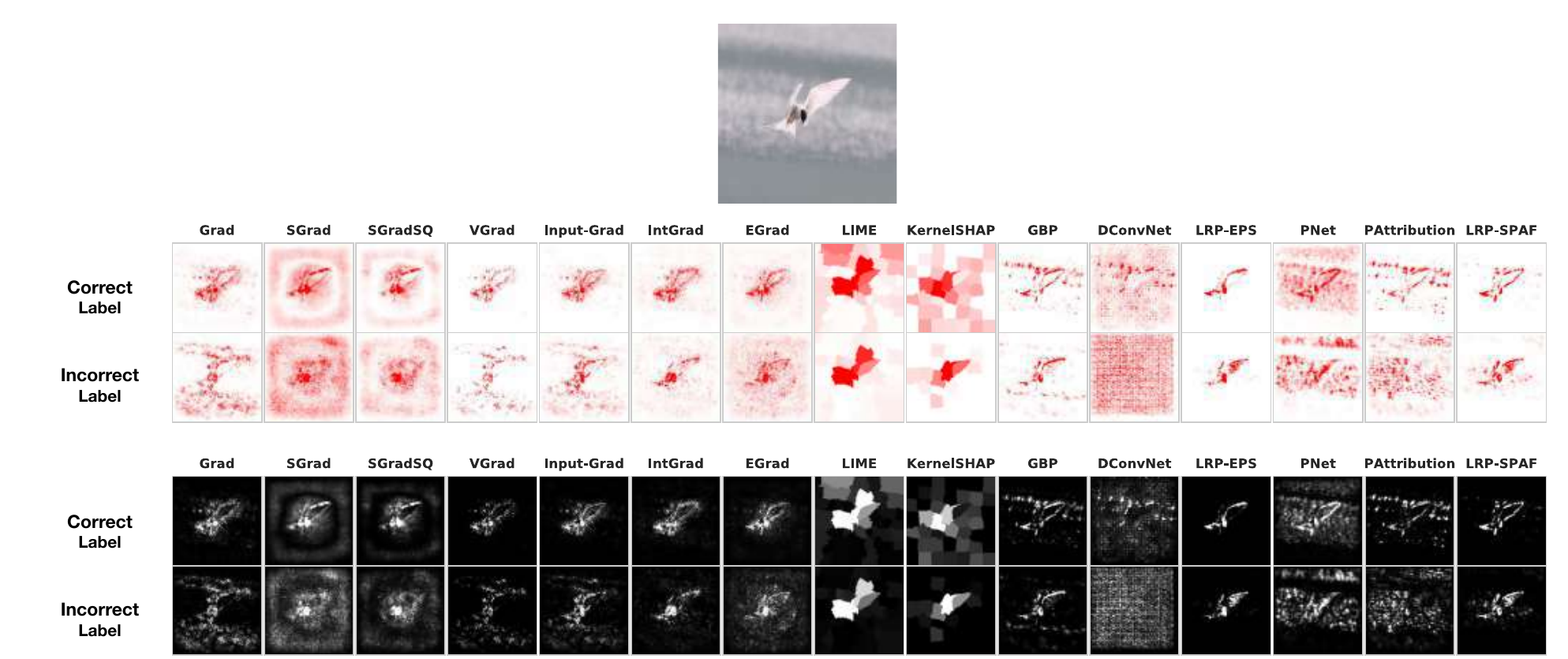}
\caption{  \textbf{Mislabelled Examples Bug.} Figure shows the input in the top row, along with attribution visualization for this input under two visualization schemes.}
\label{fig:appendixmislabelled5}
\end{figure}

\clearpage

\begin{figure}[!h]
\centering
\includegraphics[scale=0.4, page=39]{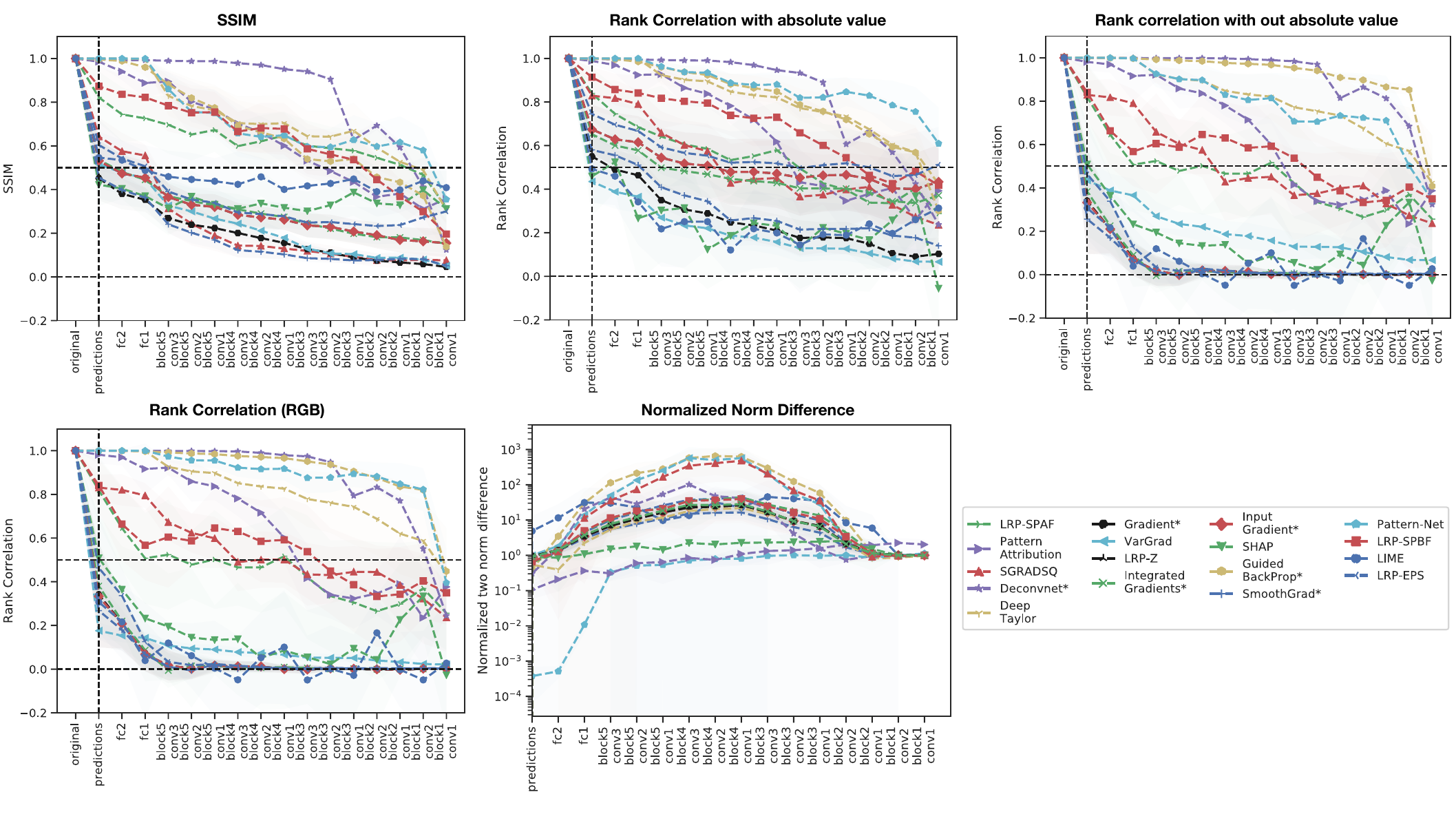}
\caption{\textbf{Additional Model Contamination Quantitative Metrics.} Similarity metrics computed for 200 images across 17 attribution types for the VGG-16 model trained on ImageNet. We use SSIM to quantify visual similarity. We also show the rank correlation metric with and without absolute values. We see that for certain methods, like those that modify backprop, across these series of metrics, these methods show high similarity. We also show the normalized norm difference for each method computed as: $\frac{\vert \vert e_{orig} - e_{rand} \vert \vert}{\vert \vert e_{orig} \vert \vert }$. This is the normalized difference in $2-$norm between the original attribution and the attribution derived from a (partially) randomized model. Note that we do not include Expected Gradients for VGG-16 experiments because it was too computationally intensive.}
\label{fig:appendixrankingmetrics}
\end{figure}
\newpage

\begin{figure*}[!h]
\centering
\includegraphics[scale=0.2]{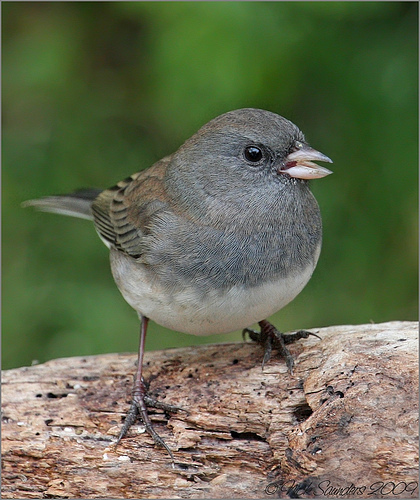}
\caption{An example Junco Bird Image.}
\label{fig:bird}
\end{figure*}

\begin{figure*}[!h]
\centering
\includegraphics[width=1.\textwidth,  page=5]{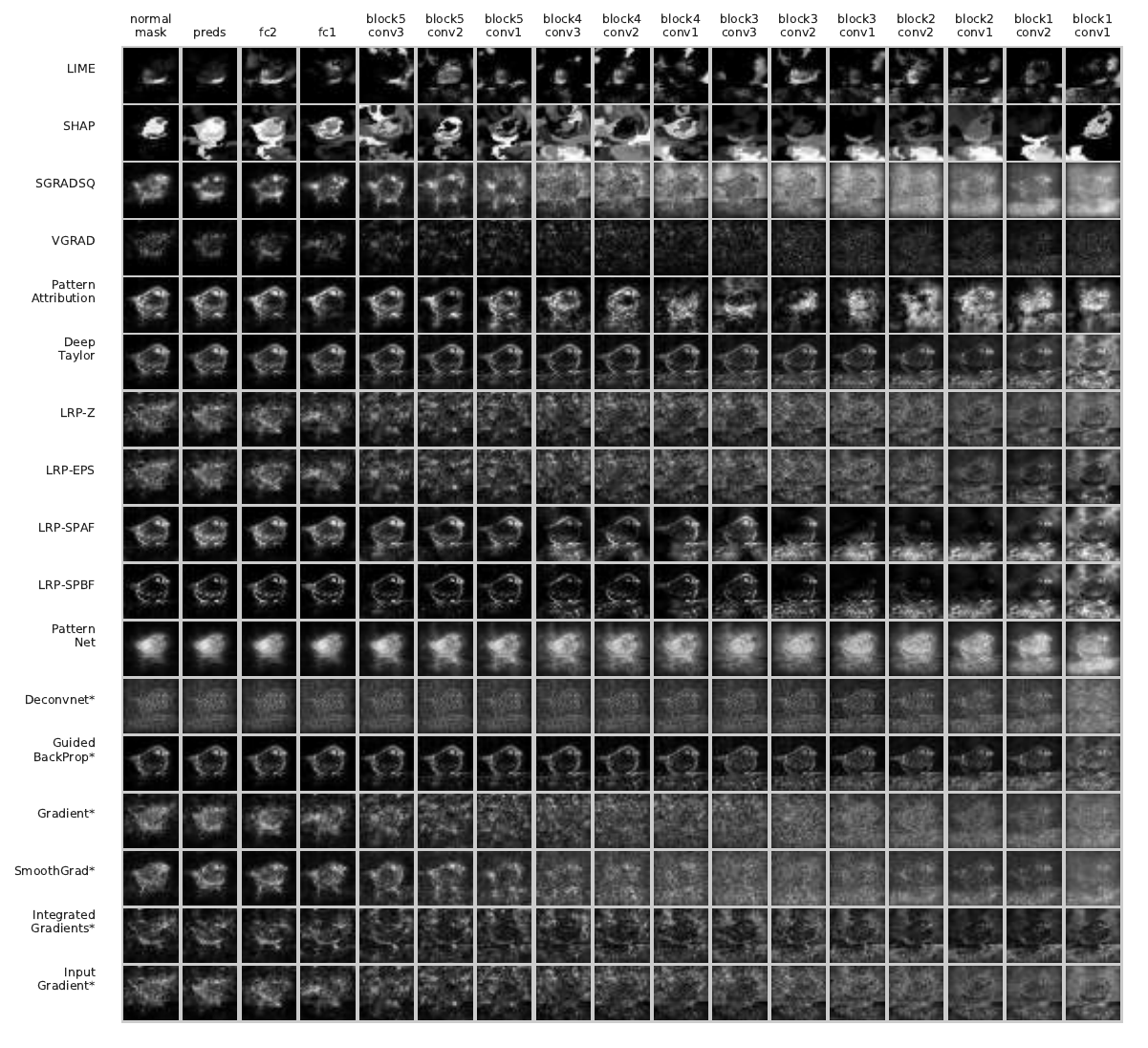}
\caption{\textbf{Model Contamination Bug VGG-16 on ImageNet.} Cascading parameter randomization on the VGG-16 model for a Junco bird example (We use the Gray Scale visualization scheme here).}
\label{fig:debugschematicbird}
\end{figure*}

\begin{figure*}[!h]
\centering
\includegraphics[width=1.\textwidth,  page=6]{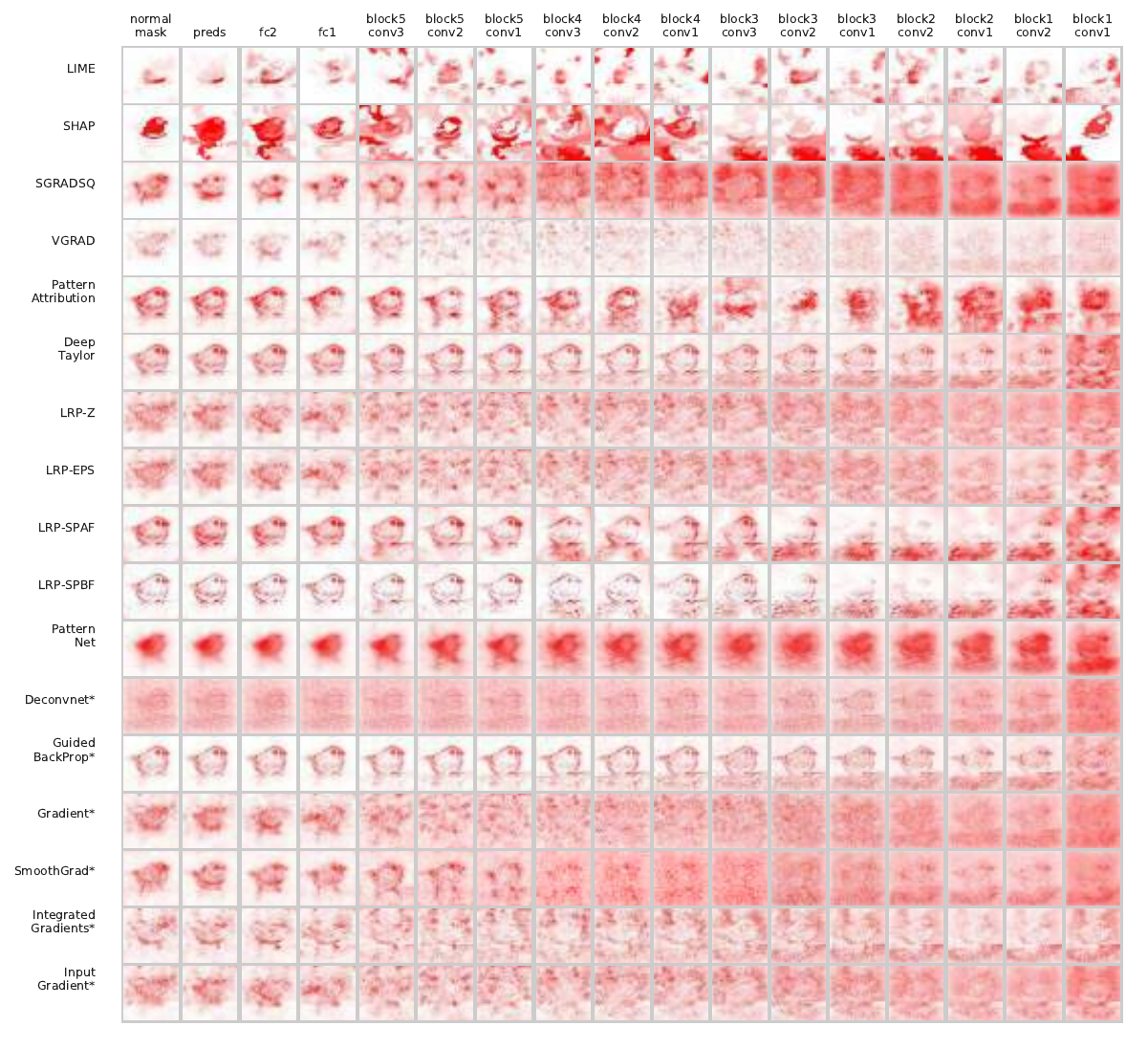}
\caption{\textbf{Model Contamination Bug VGG-16 on ImageNet.} Cascading parameter randomization on the VGG-16 model for a Junco bird example (We use the Red visualization scheme here)}
\label{fig:debugschematicbird2}
\end{figure*}

\begin{figure*}[!h]
\centering
\includegraphics[scale=0.2]{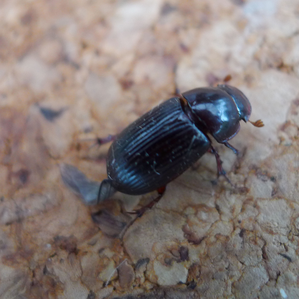}
\caption{An example Bug Image.}
\label{fig:bug}
\end{figure*}

\begin{figure*}[!h]
\centering
\includegraphics[width=1.\textwidth,  page=7]{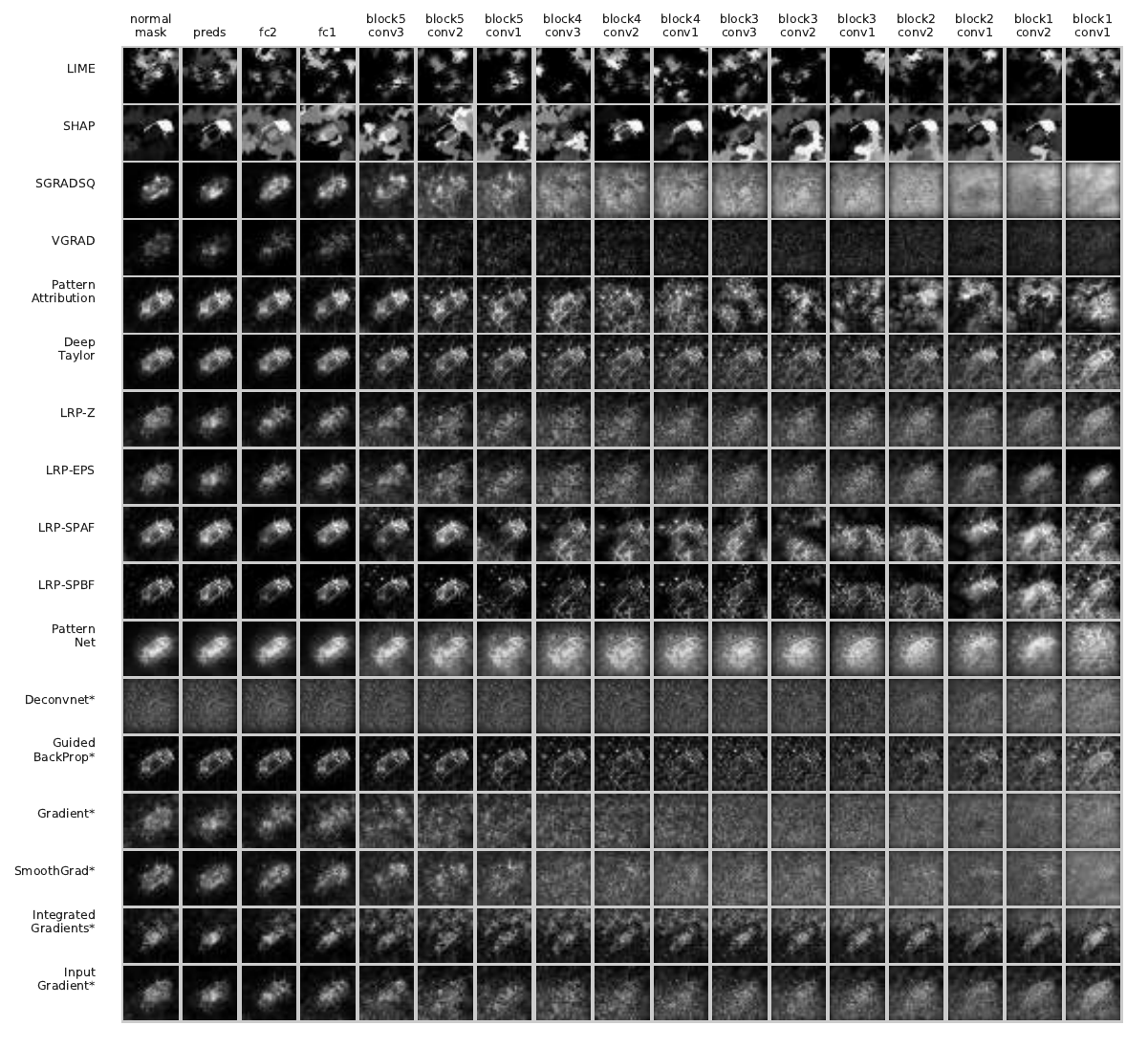}
\caption{\textbf{Model Contamination Bug VGG-16 on ImageNet.} Cascading parameter randomization on the VGG-16 model for a bug example.}
\label{fig:debugschematicbug}
\end{figure*}

\begin{figure*}[!h]
\centering
\includegraphics[width=1.\textwidth,  page=8]{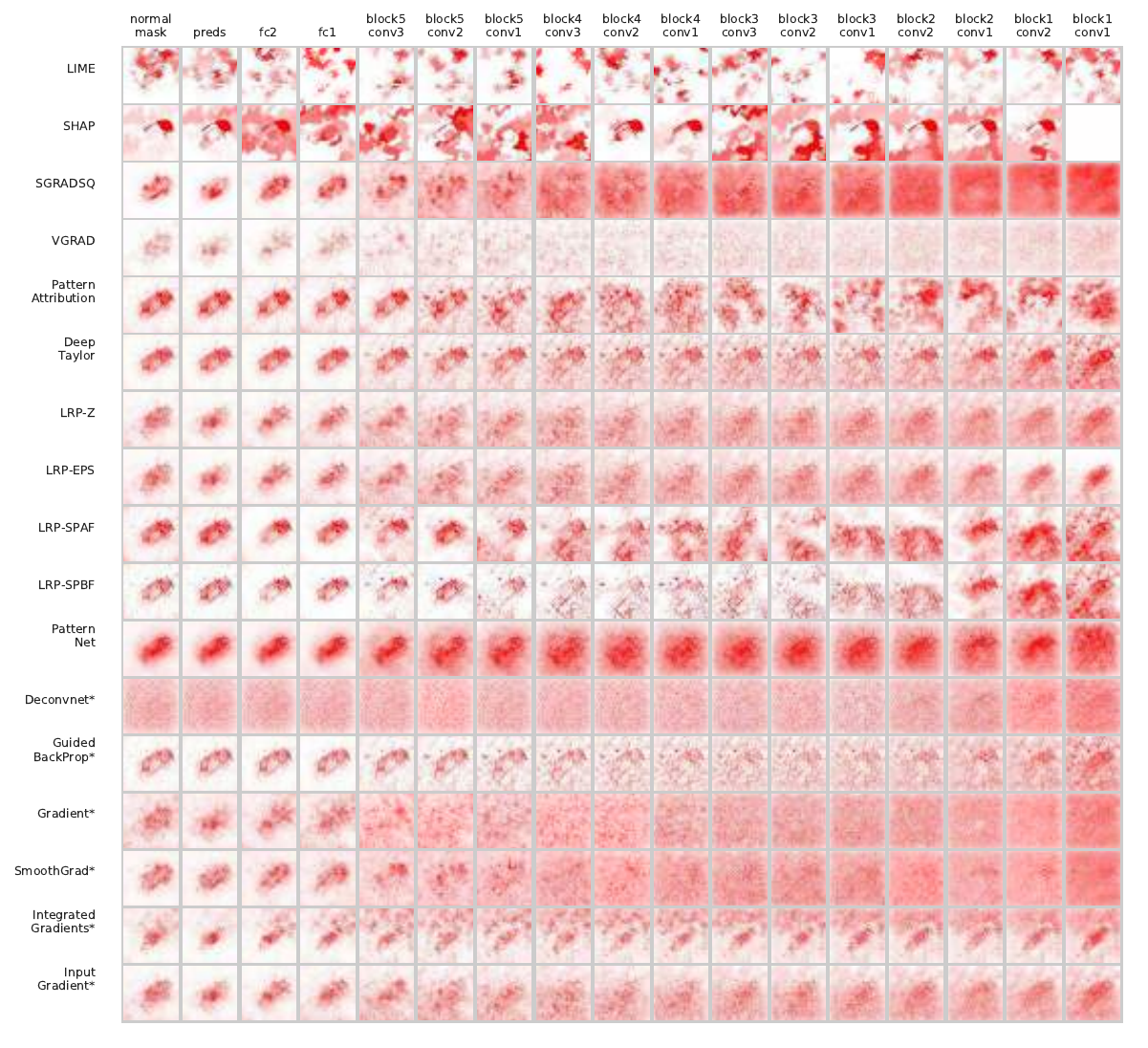}
\caption{\textbf{Model Contamination Bug VGG-16 on ImageNet.} Cascading parameter randomization on the VGG-16 model for a bug example.}
\label{fig:debugschematicbug2}
\end{figure*}

\clearpage

\begin{figure*}[!h]
\centering
\includegraphics[scale=0.4]{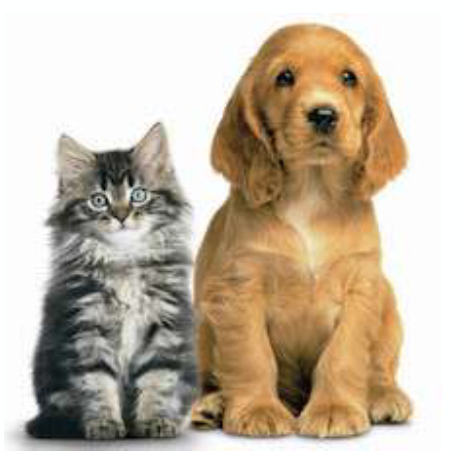}
\caption{An example Dog-Cat Image.}
\label{fig:dogcat}
\end{figure*}

\begin{figure*}[!h]
\centering
\includegraphics[width=1.\textwidth,  page=9]{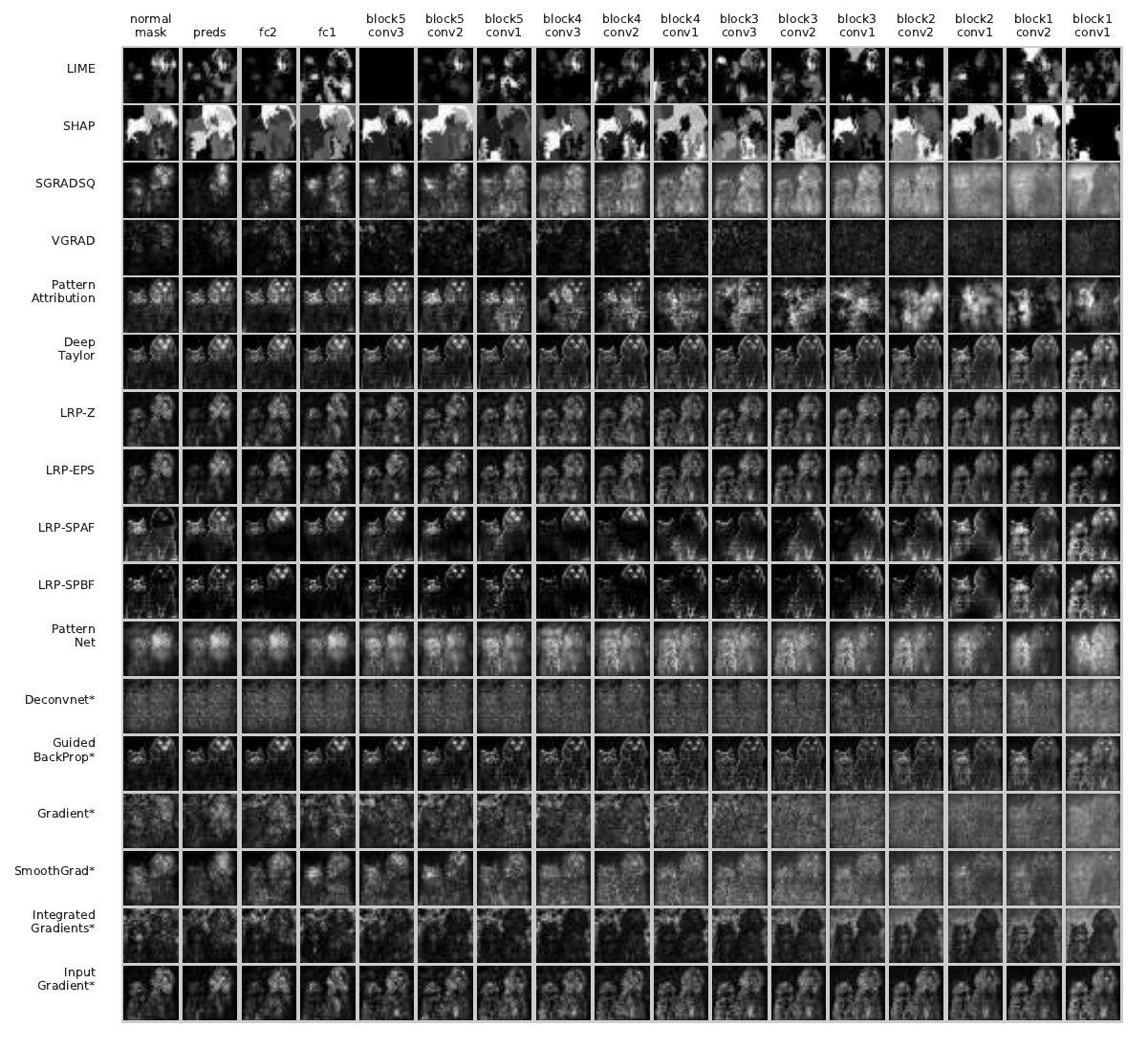}
\caption{\textbf{Model Contamination Bug VGG-16 on ImageNet.} Cascading parameter randomization on the VGG-16 model for the dog-cat example.}
\label{fig:debugschematicdogcat}
\end{figure*}

\begin{figure*}[!h]
\centering
\includegraphics[width=1.\textwidth,  page=10]{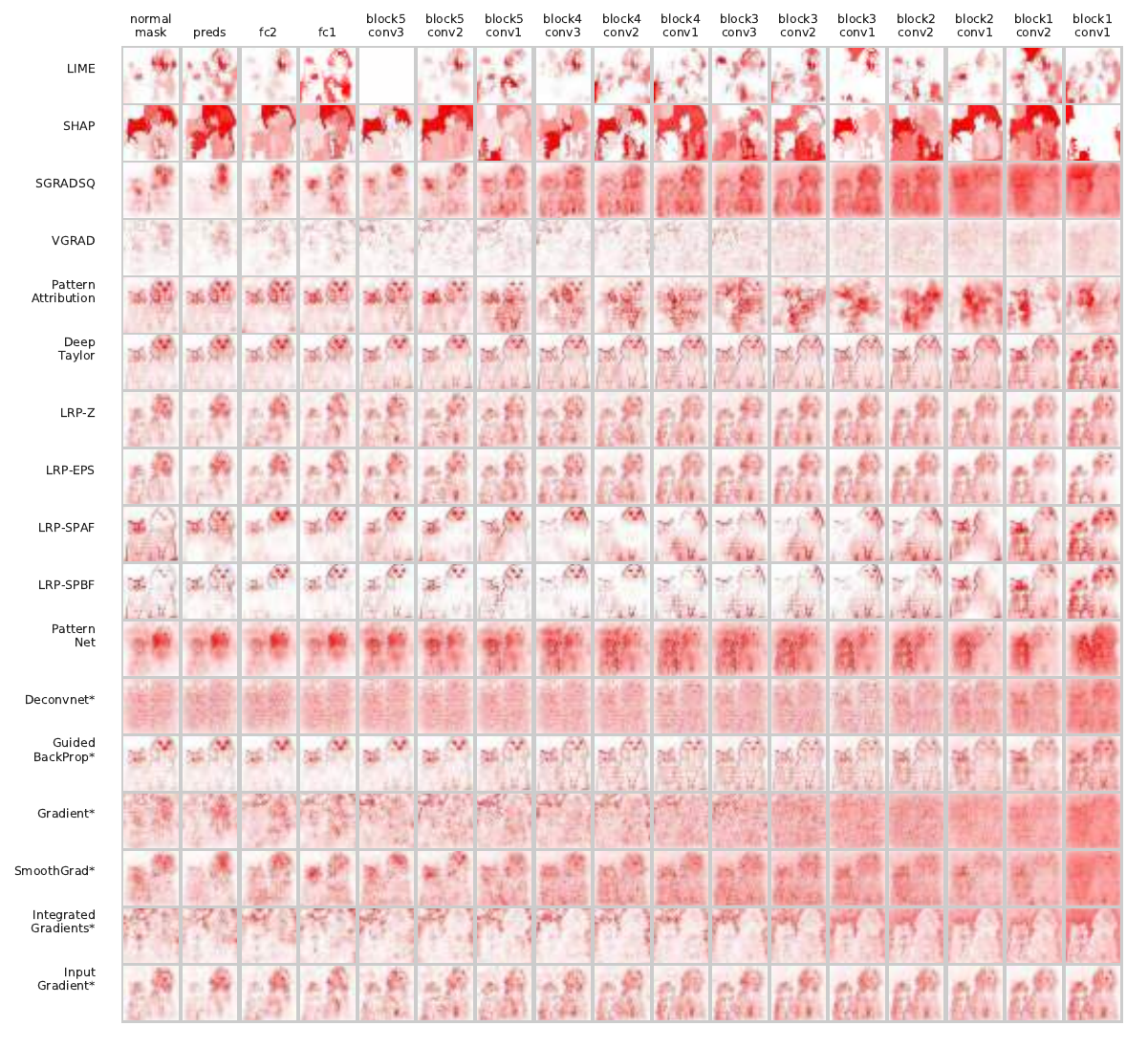}
\caption{\textbf{Model Contamination Bug VGG-16 on ImageNet.} Cascading parameter randomization on the VGG-16 model for the dog-cat example.}
\label{fig:debugschematicdogcat2}
\end{figure*}

\begin{figure*}[!h]
\centering
\includegraphics[scale=0.4]{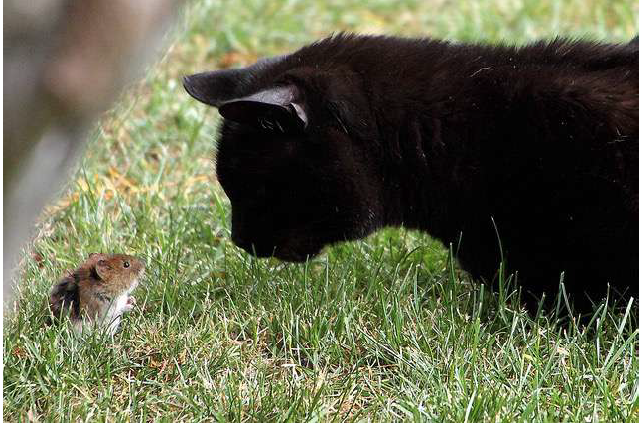}
\caption{An example Mouse-Cat Image.}
\label{fig:dogcat}
\end{figure*}

\begin{figure*}[!h]
\centering
\includegraphics[width=1.\textwidth,  page=11]{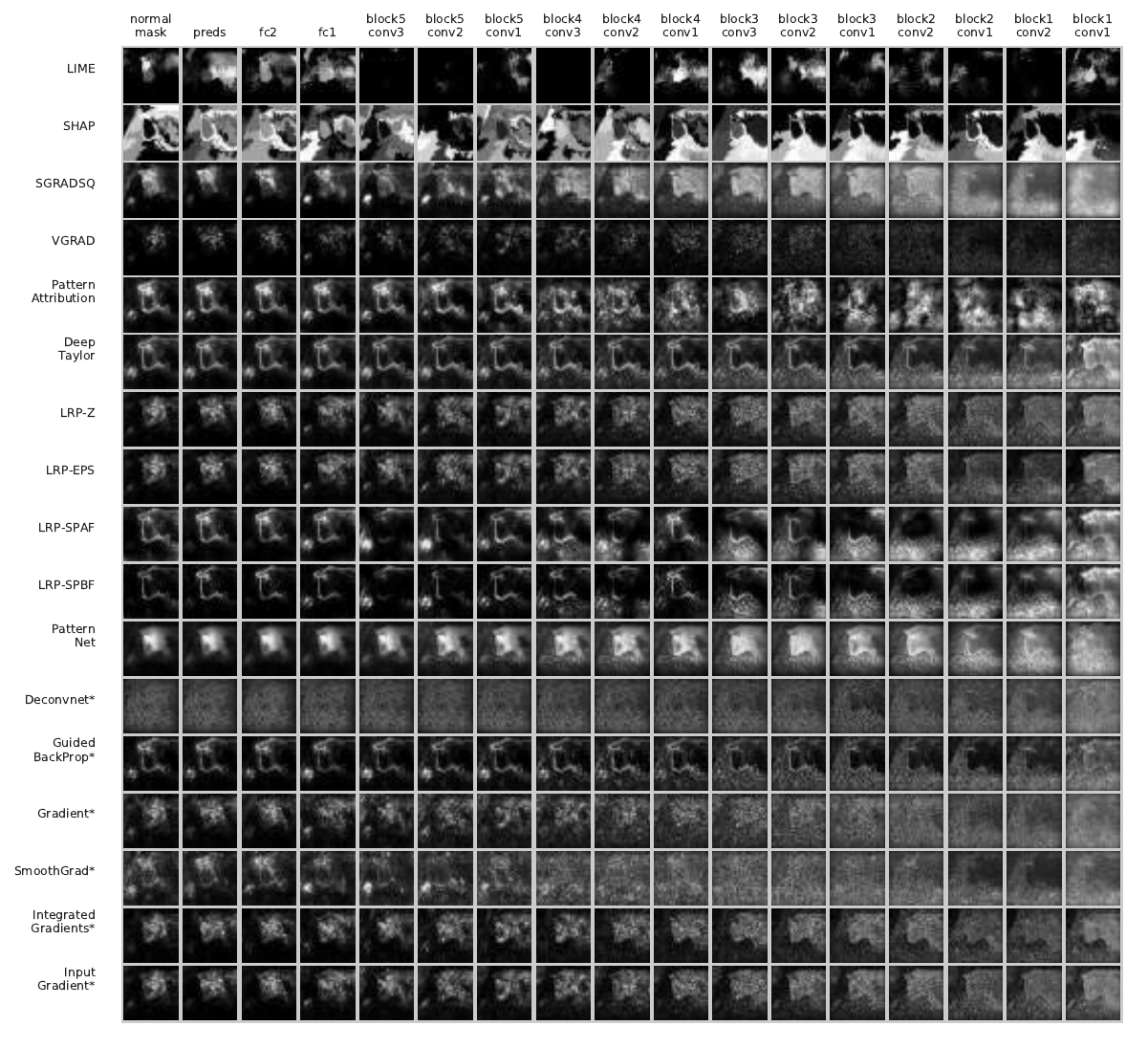}
\caption{\textbf{Model Contamination Bug VGG-16 on ImageNet.} Cascading parameter randomization on the VGG-16 model for the Mouse-Cat example.}
\label{fig:debugschematic}
\end{figure*}

\begin{figure*}[!h]
\centering
\includegraphics[width=1.\textwidth,  page=12]{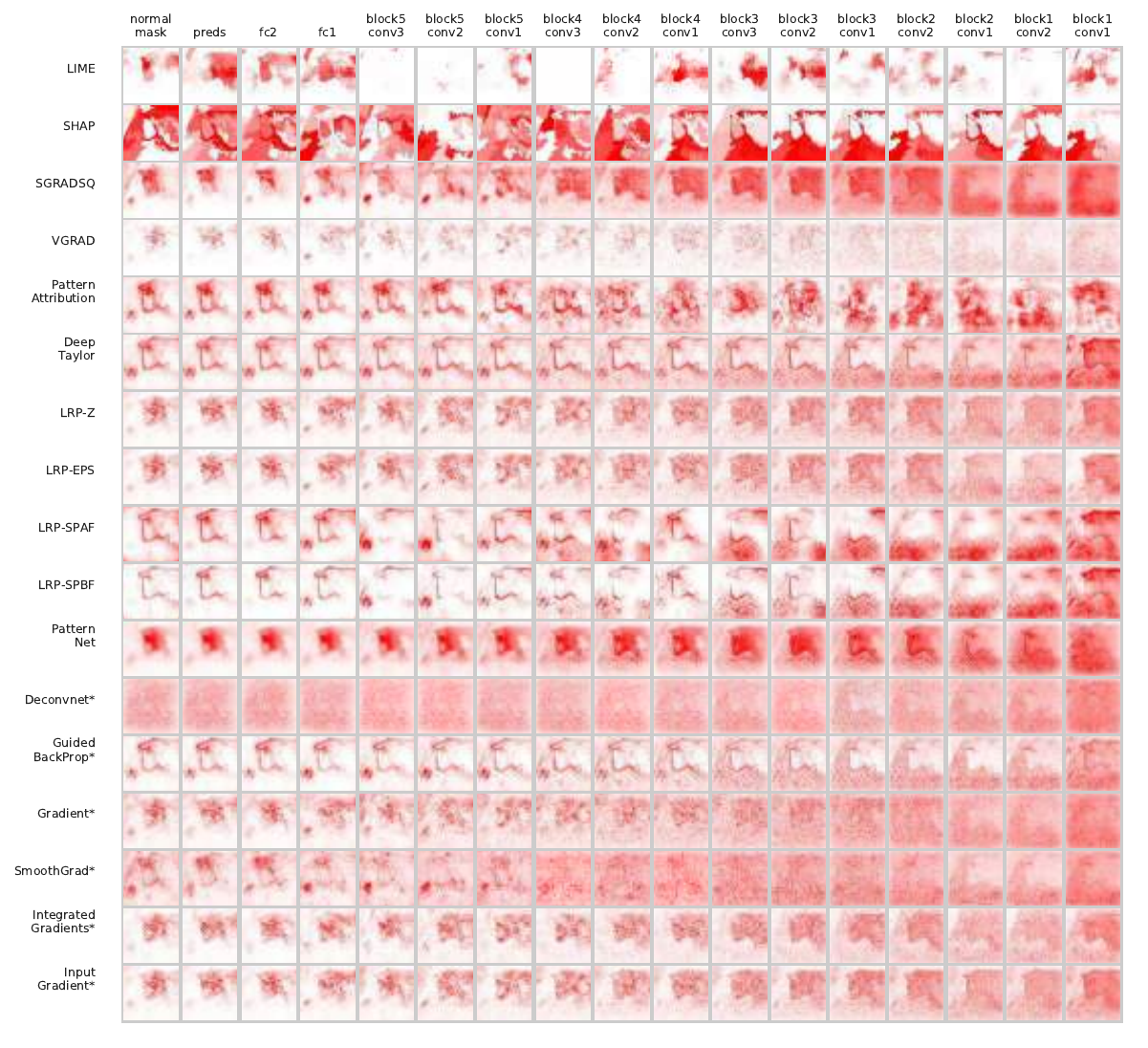}
\caption{\textbf{Model Contamination Bug VGG-16 on ImageNet.} Cascading parameter randomization on the VGG-16 model for the Mouse-Cat example.}
\label{fig:debugschematic}
\end{figure*}

\begin{figure*}[!h]
\centering
\includegraphics[scale=10]{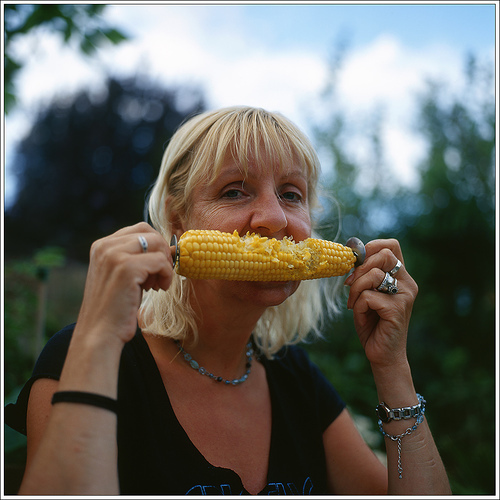}
\caption{An example Corn Image.}
\label{fig:dogcat}
\end{figure*}

\begin{figure*}[!h]
\centering
\includegraphics[width=1.\textwidth,  page=13]{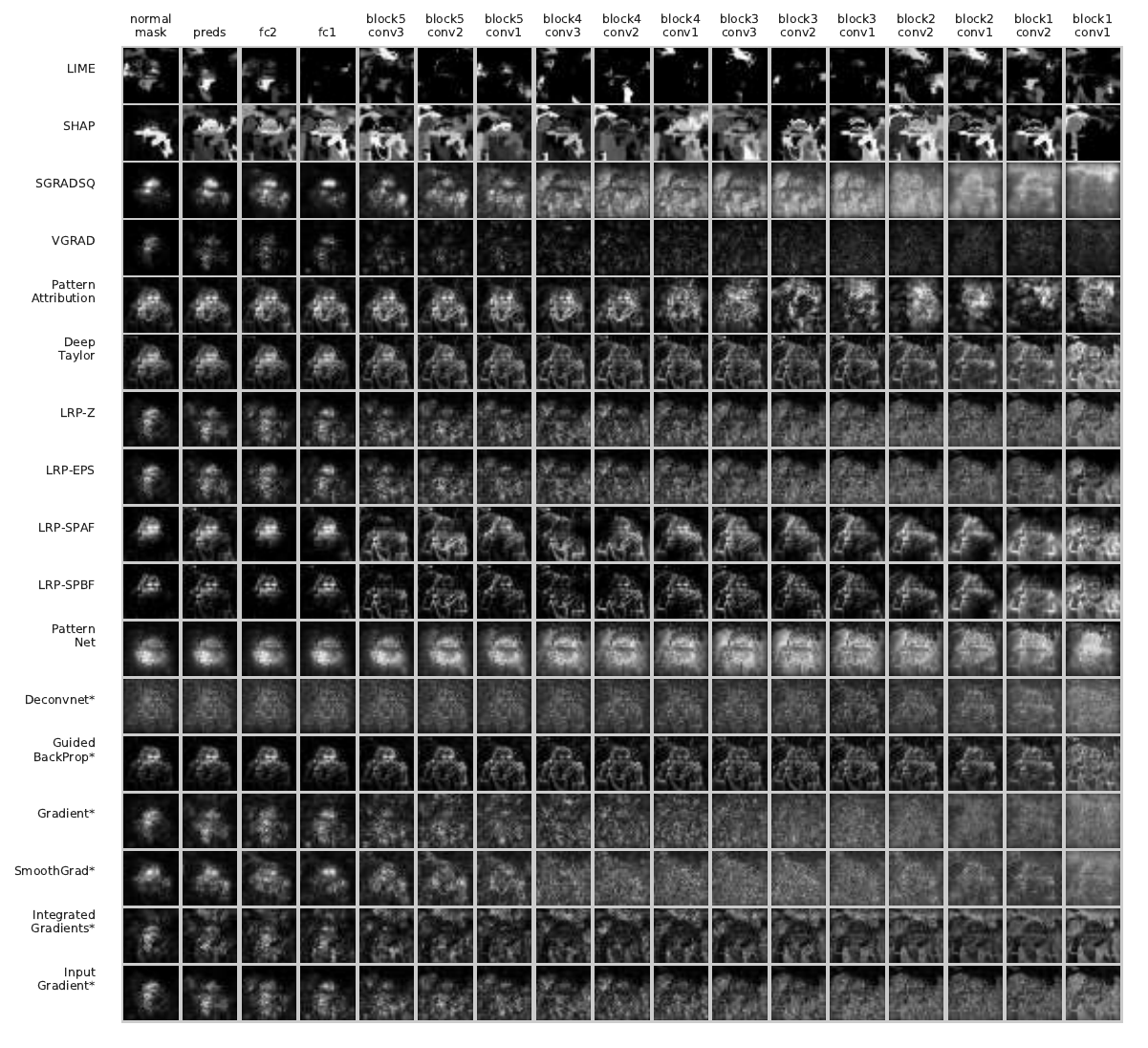}
\caption{\textbf{Model Contamination Bug VGG-16 on ImageNet.} Cascading parameter randomization on the VGG-16 model for the Corn example.}
\label{fig:debugschematic}
\end{figure*}

\begin{figure*}[!h]
\centering
\includegraphics[width=1.\textwidth,  page=14]{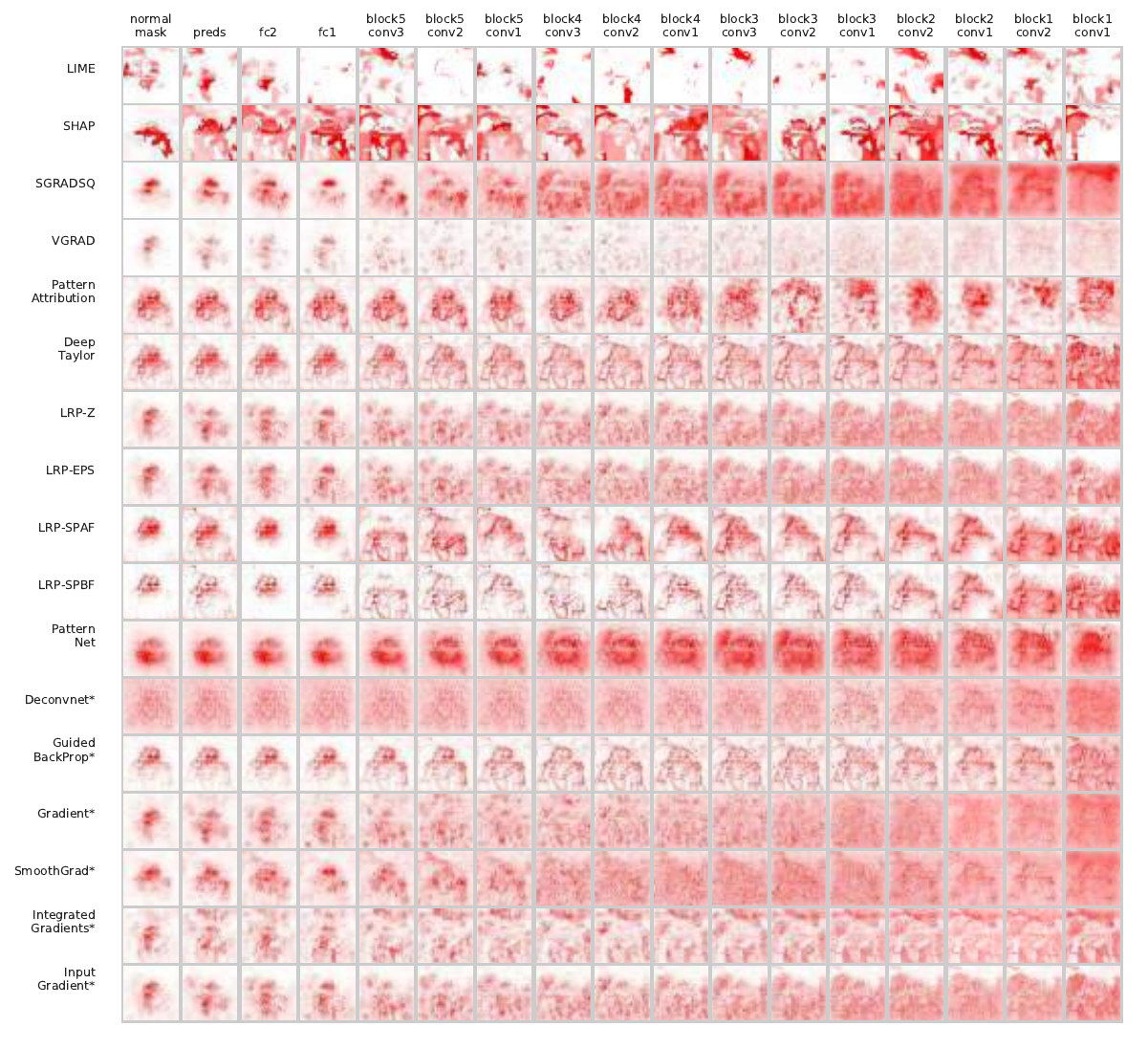}
\caption{\textbf{Model Contamination Bug VGG-16 on ImageNet.} Cascading parameter randomization on the VGG-16 model for the Corn example.}
\label{fig:debugschematic}
\end{figure*}

\begin{figure*}[!h]
\centering
\includegraphics[scale=0.25]{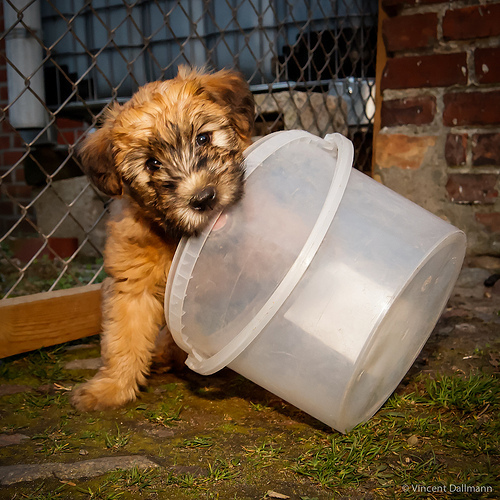}
\caption{An example Dog biting on a Bucket.}
\label{fig:dog}
\end{figure*}

\begin{figure*}[!h]
\centering
\includegraphics[width=1.\textwidth,  page=15]{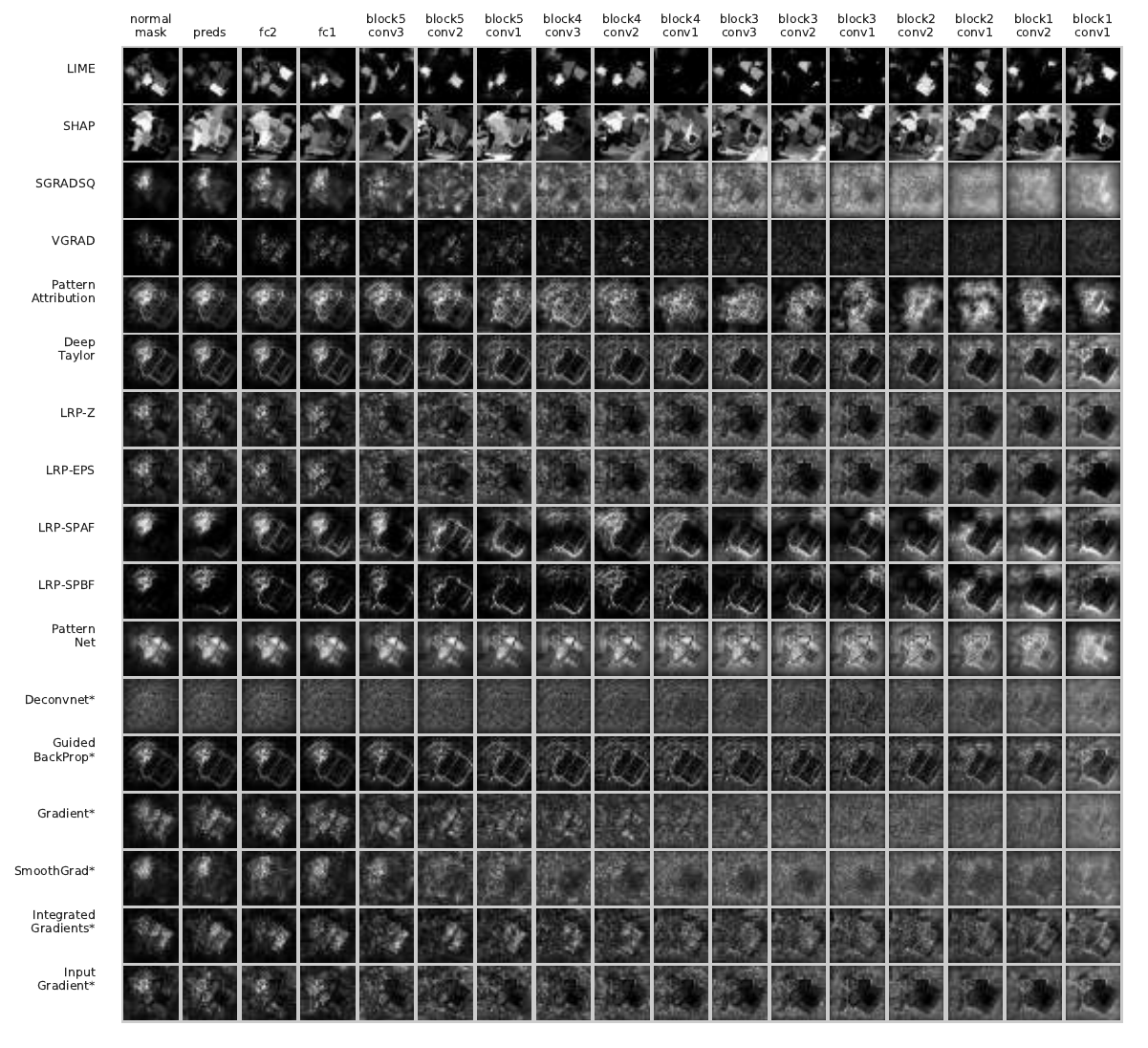}
\caption{\textbf{Model Contamination Bug VGG-16 on ImageNet.} Cascading parameter randomization on the VGG-16 model for the Dog example.}
\label{fig:debugschematicdog}
\end{figure*}

\clearpage

\begin{figure}[!h]
\centering
\includegraphics[scale=0.4, page=24]{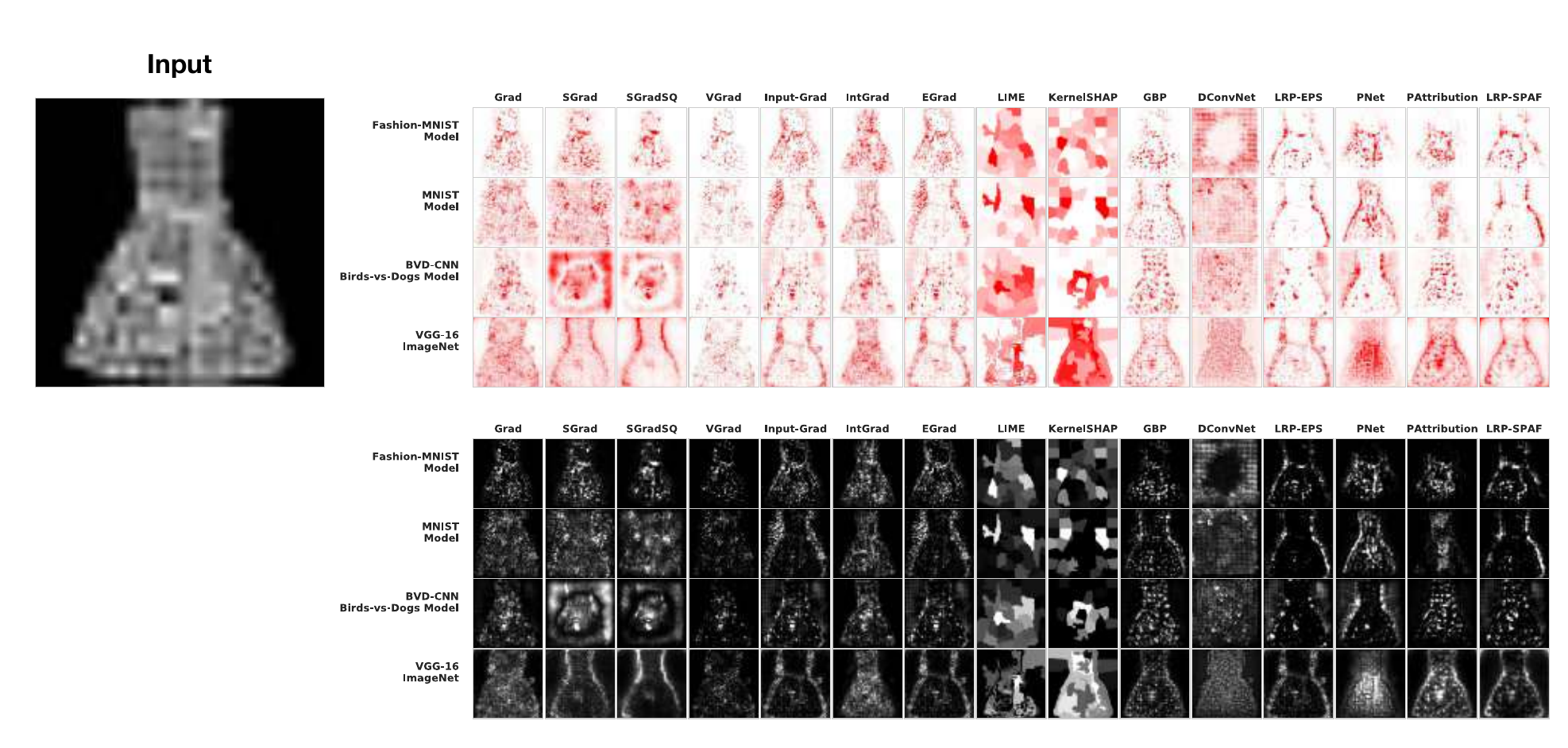}
\caption{\textbf{Test-time Randomization Example for Fashion MNIST.} We show Fashion MNIST attributions for $4$ different models: A fashion MNIST model (trained on fashion MNIST), an MNIST Model (trained on MNIST), a birds-vs-dogs model (trained on birds-vs-dogs dataset), and a VGG-16 model trained on IMAGENET.}
\label{fig:appendixtesttime1}
\end{figure}

\begin{figure}[!h]
\centering
\includegraphics[scale=0.4, page=25]{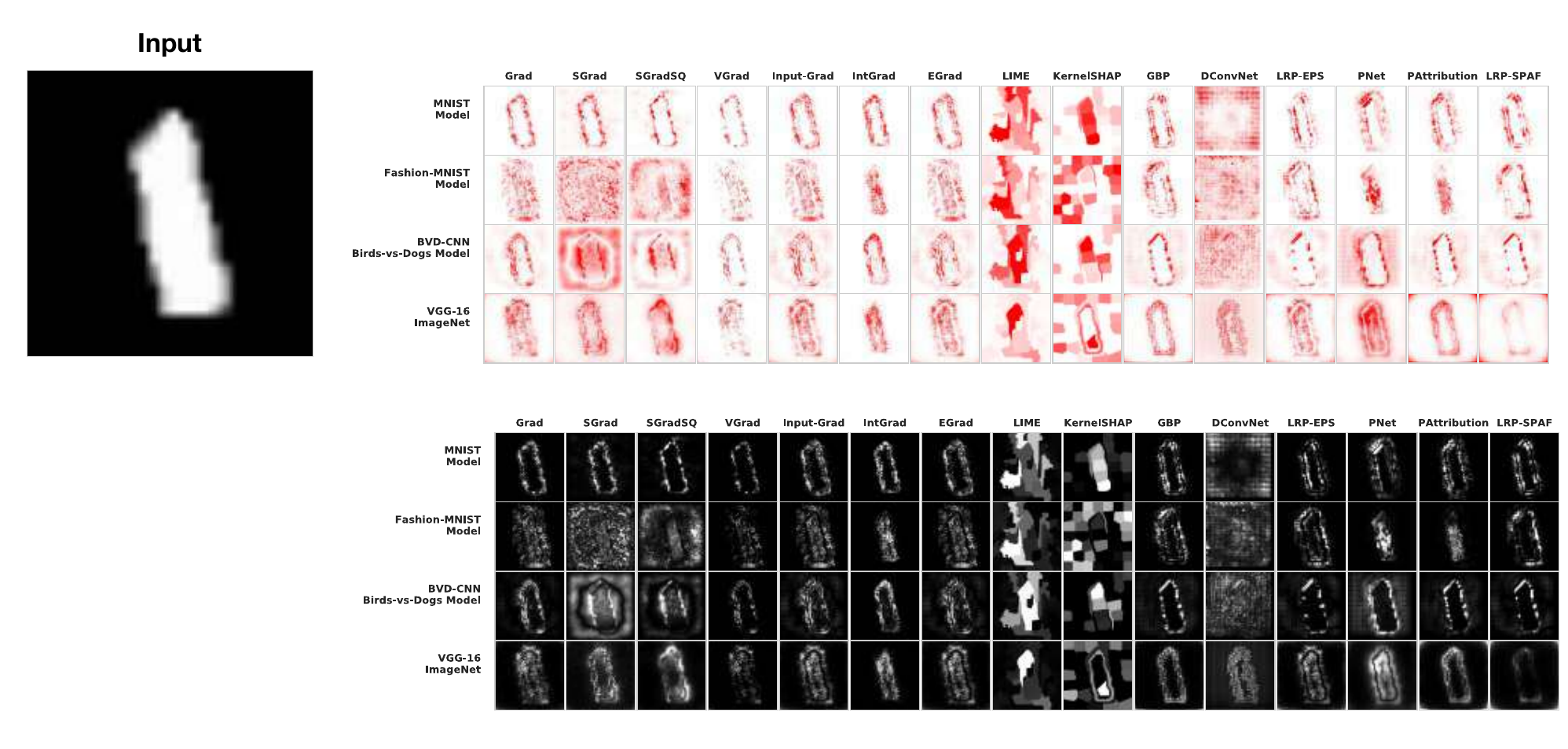}
\caption{\textbf{Test-time Randomization Example for MNIST.} We show MNIST attributions for $4$ different models: A fashion MNIST model (trained on fashion MNIST), an MNIST Model (trained on MNIST), a birds-vs-dogs model (trained on birds-vs-dogs dataset), and a VGG-16 model trained on IMAGENET.}
\label{fig:appendixtesttime2}
\end{figure}

\begin{figure}[!h]
\centering
\includegraphics[scale=0.4, page=26]{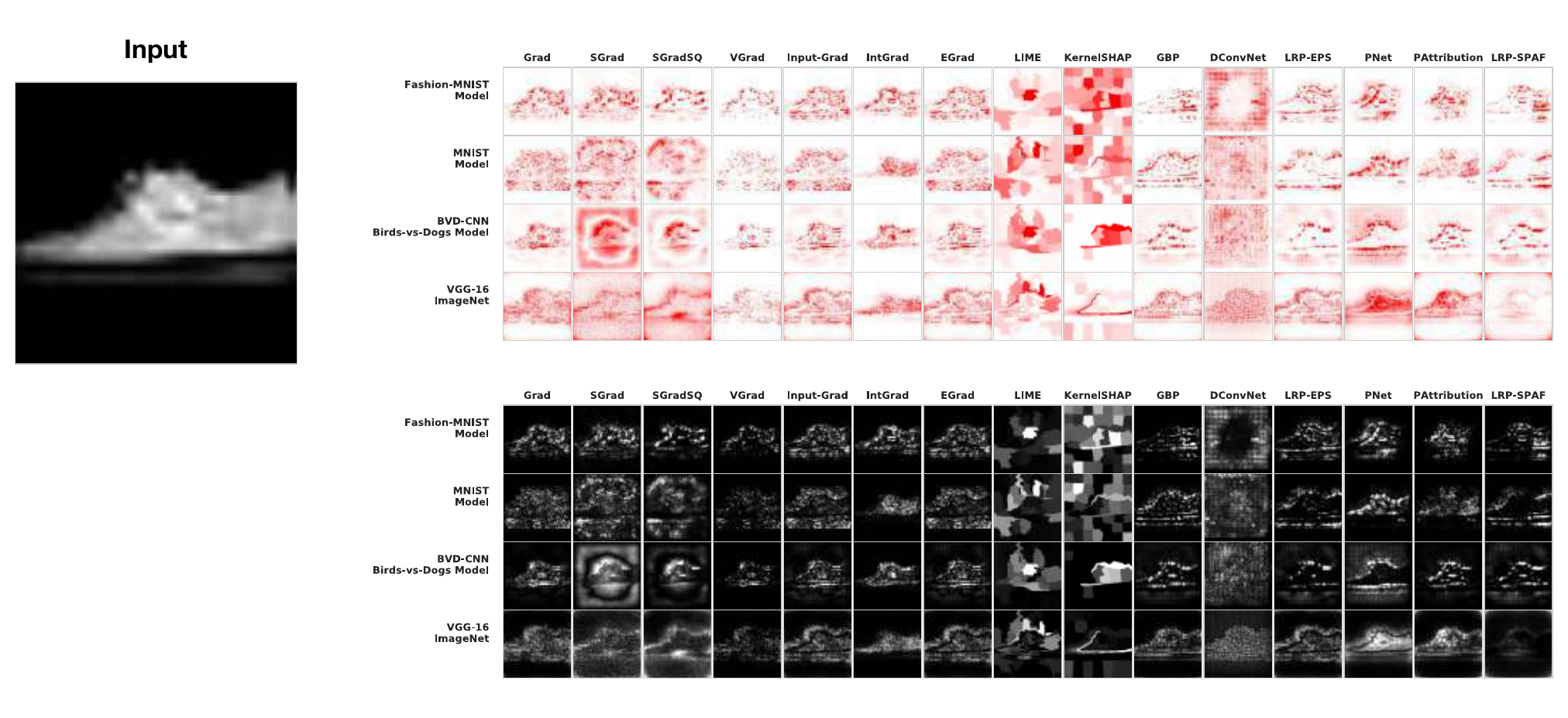}
\caption{\textbf{Test-time Randomization Example for Fashion MNIST.} We show Fashion MNIST attributions for $4$ different models: A fashion MNIST model (trained on fashion MNIST), an MNIST Model (trained on MNIST), a birds-vs-dogs model (trained on birds-vs-dogs dataset), and a VGG-16 model trained on IMAGENET.}
\label{fig:appendixtesttime3}
\end{figure}

\begin{figure}[!h]
\centering
\includegraphics[scale=0.4, page=27]{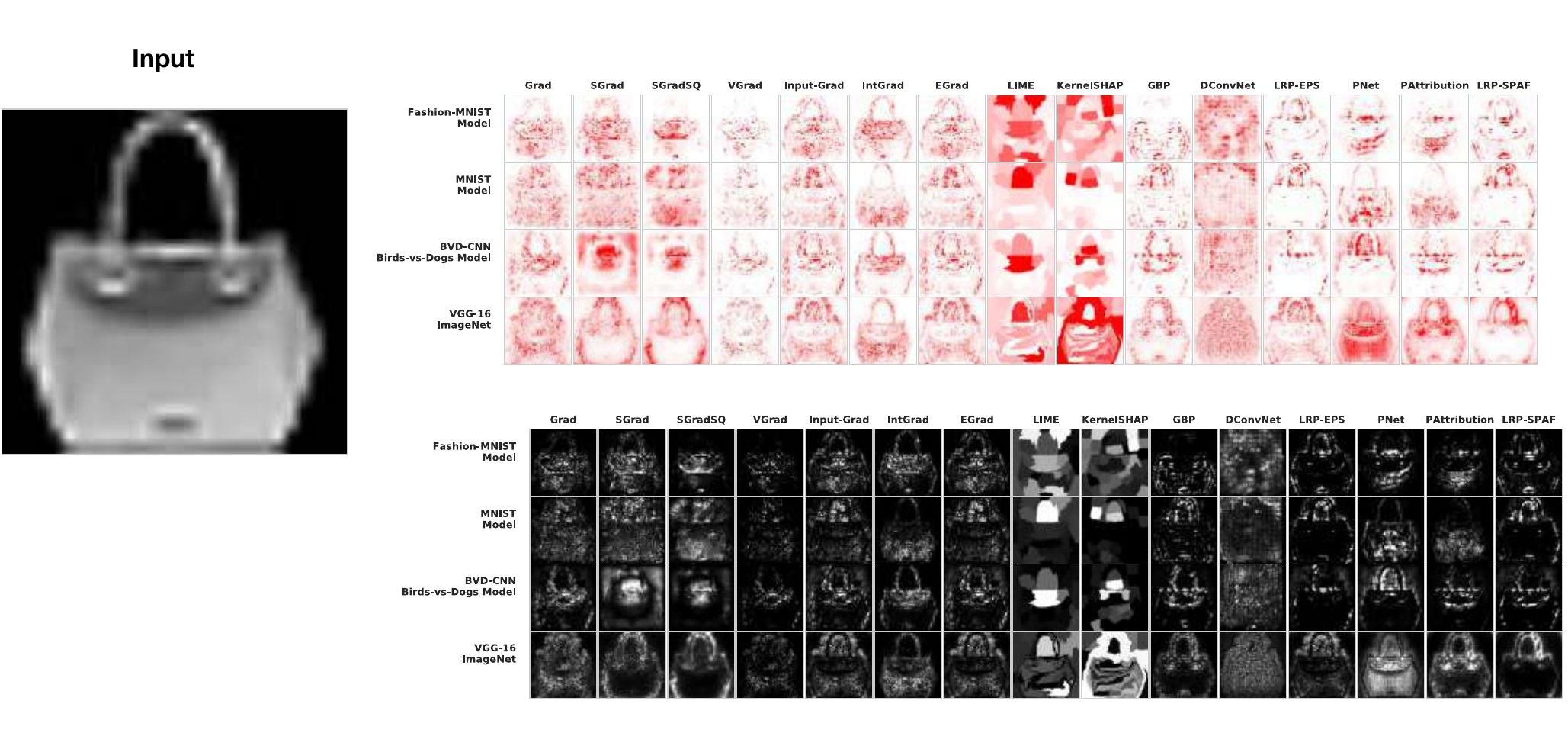}
\caption{\textbf{Test-time Randomization Example for Fashion MNIST.} We show Fashion MNIST attributions for $4$ different models: A fashion MNIST model (trained on fashion MNIST), an MNIST Model (trained on MNIST), a birds-vs-dogs model (trained on birds-vs-dogs dataset), and a VGG-16 model trained on IMAGENET.}
\label{fig:appendixtesttime3}
\end{figure}

\begin{figure}[!h]
\centering
\includegraphics[scale=0.4, page=28]{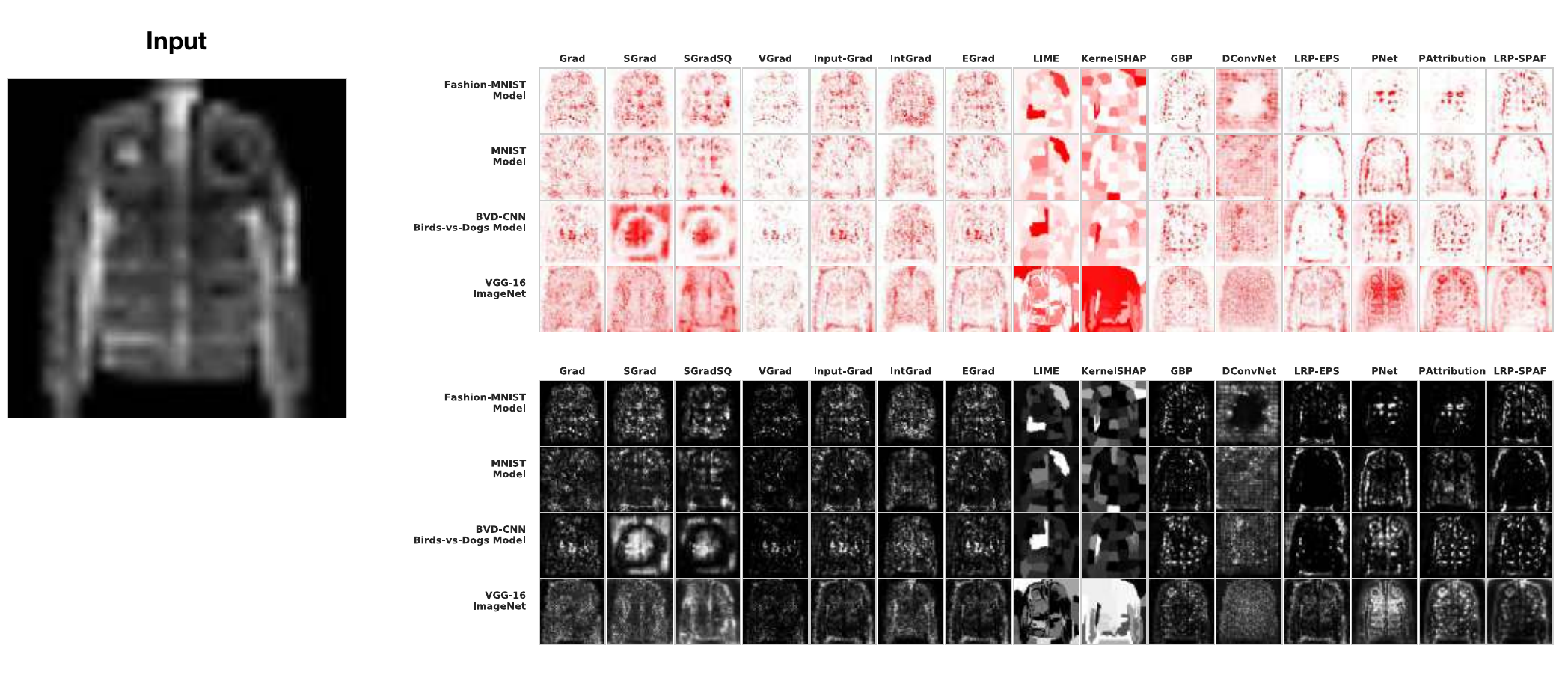}
\caption{\textbf{Test-time Randomization Example for Fashion MNIST.} We show Fashion MNIST attributions for $4$ different models: A fashion MNIST model (trained on fashion MNIST), an MNIST Model (trained on MNIST), a birds-vs-dogs model (trained on birds-vs-dogs dataset), and a VGG-16 model trained on IMAGENET.}
\label{fig:appendixtesttime3}
\end{figure}

\begin{figure}[!h]
\centering
\includegraphics[scale=0.4, page=29]{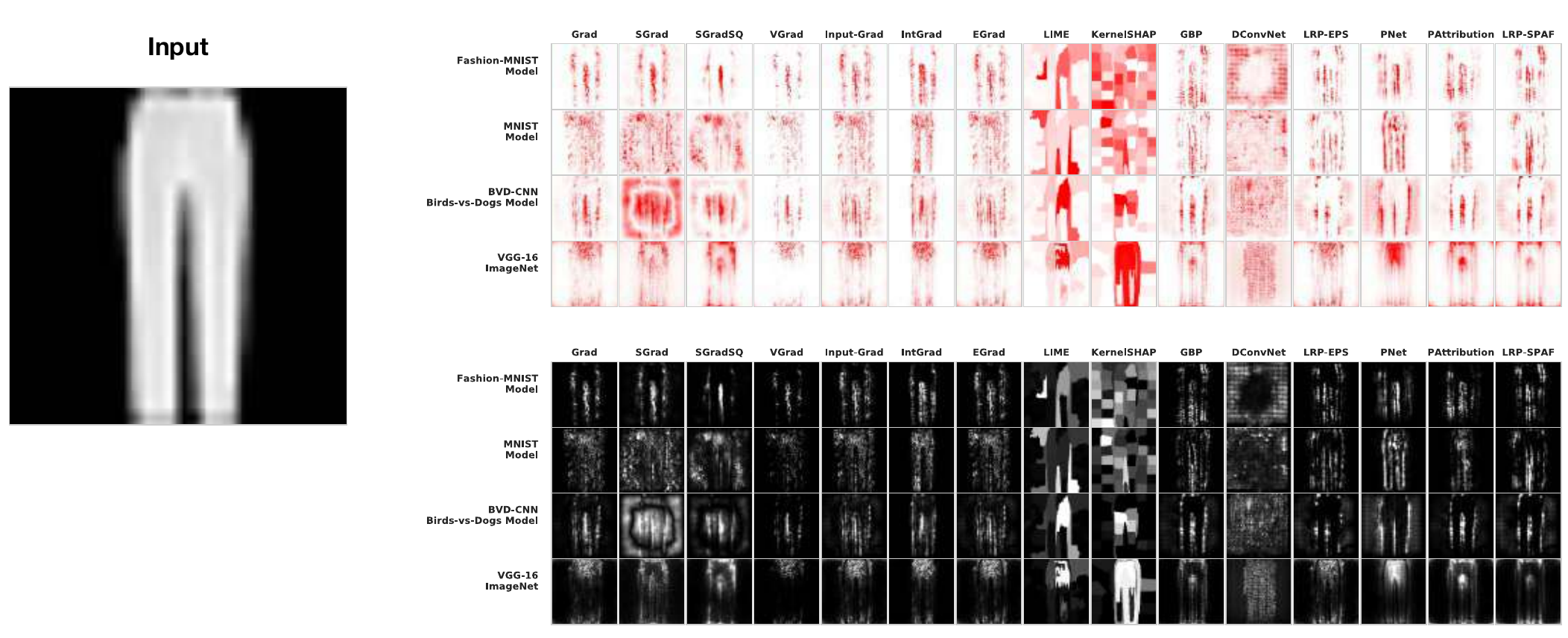}
\caption{\textbf{Test-time Randomization Example for Fashion MNIST.} We show Fashion MNIST attributions for $4$ different models: A fashion MNIST model (trained on fashion MNIST), an MNIST Model (trained on MNIST), a birds-vs-dogs model (trained on birds-vs-dogs dataset), and a VGG-16 model trained on IMAGENET.}
\label{fig:appendixtesttime3}
\end{figure}

\clearpage

\begin{figure*}[!h]
\centering
\includegraphics[width=1.\textwidth,  page=65]{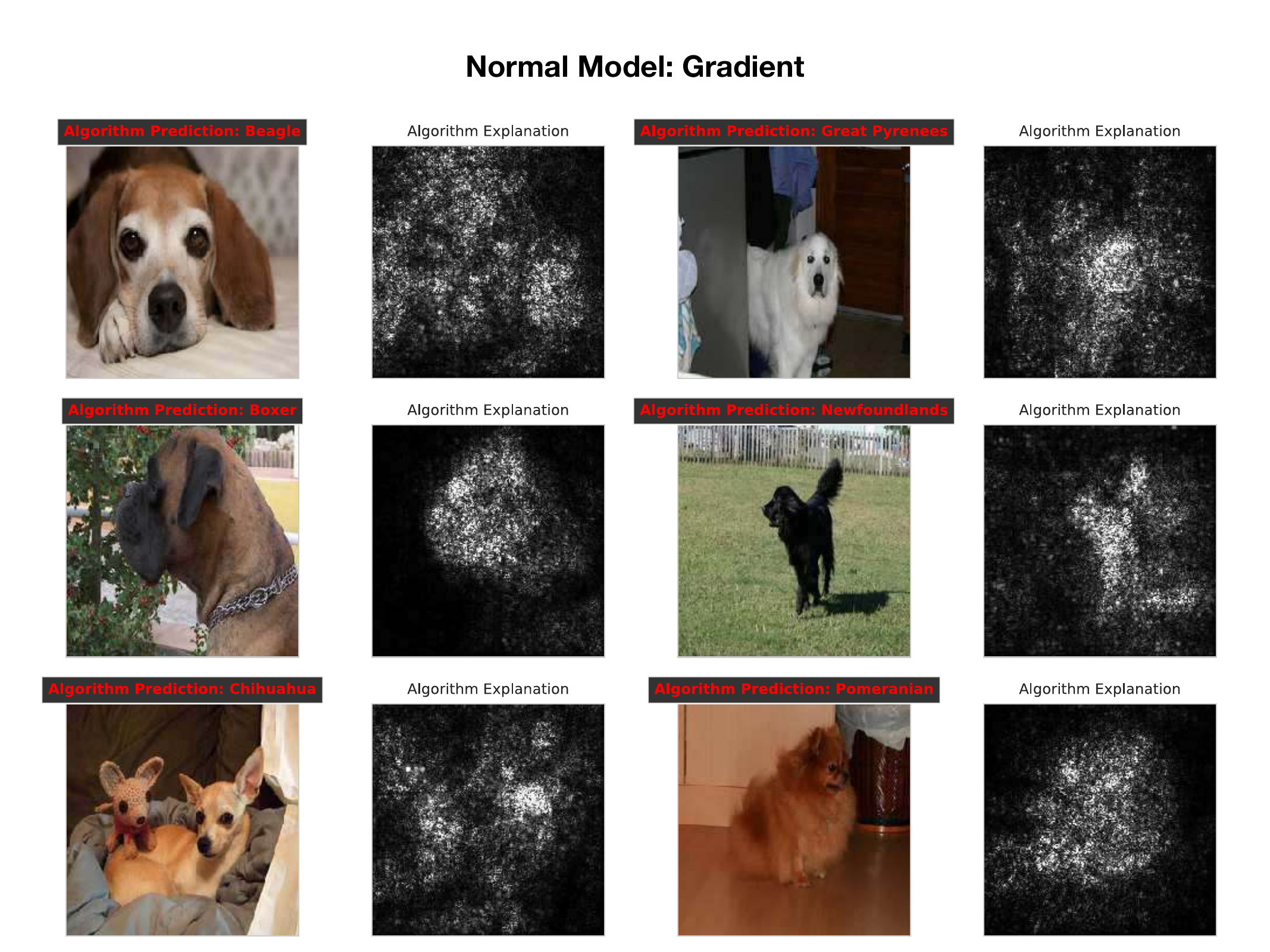}
\caption{Images used as part of the user study.}
\label{fig:userstudy1}
\end{figure*}

\begin{figure*}[!h]
\centering
\includegraphics[width=1.\textwidth,  page=66]{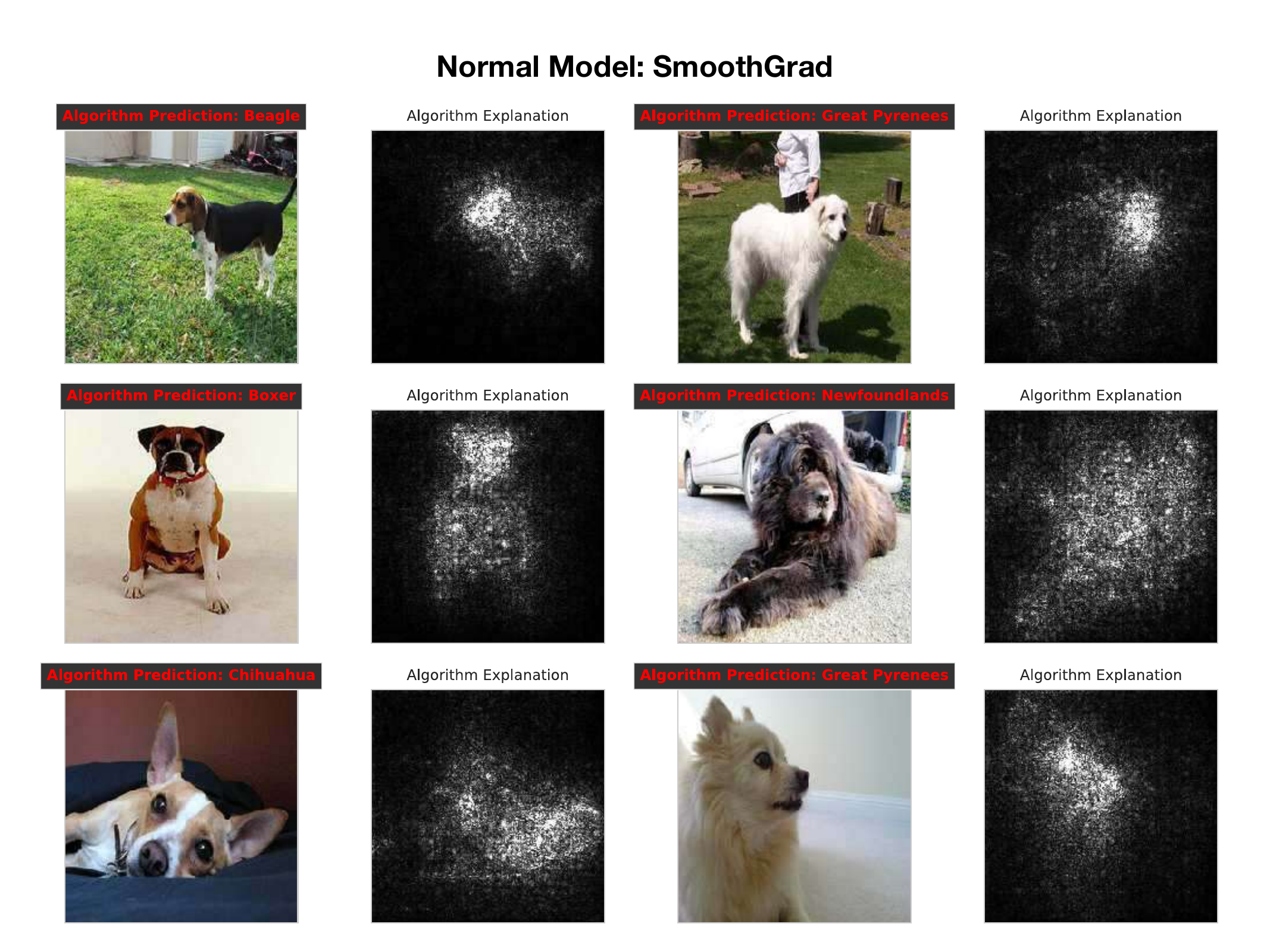}
\caption{Images used as part of the user study.}
\label{fig:userstudy2}
\end{figure*}

\begin{figure*}[!h]
\centering
\includegraphics[width=1.\textwidth,  page=67]{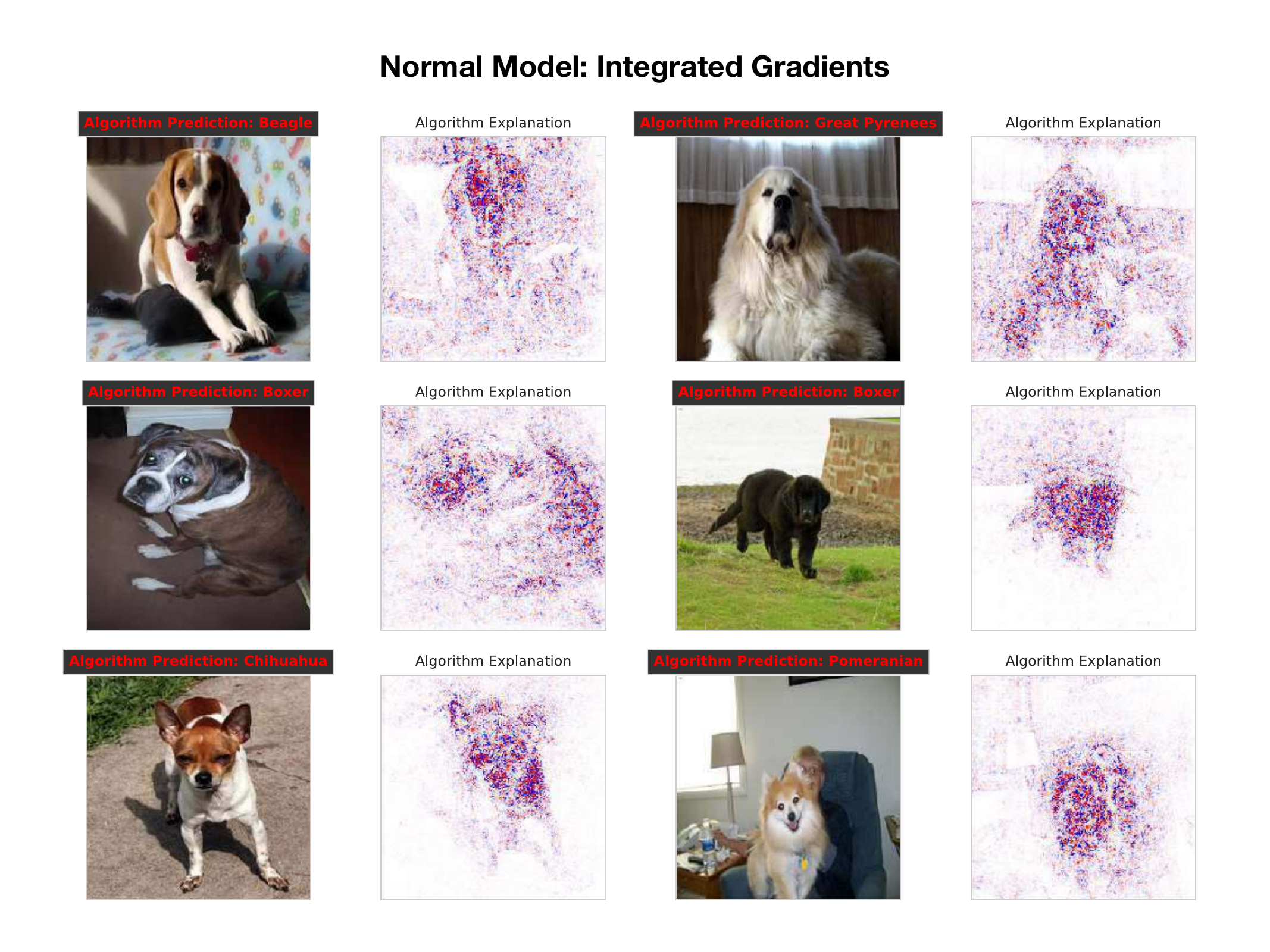}
\caption{Images used as part of the user study.}
\label{fig:userstudy3}
\end{figure*}

\begin{figure*}[!h]
\centering
\includegraphics[width=1.\textwidth,  page=68]{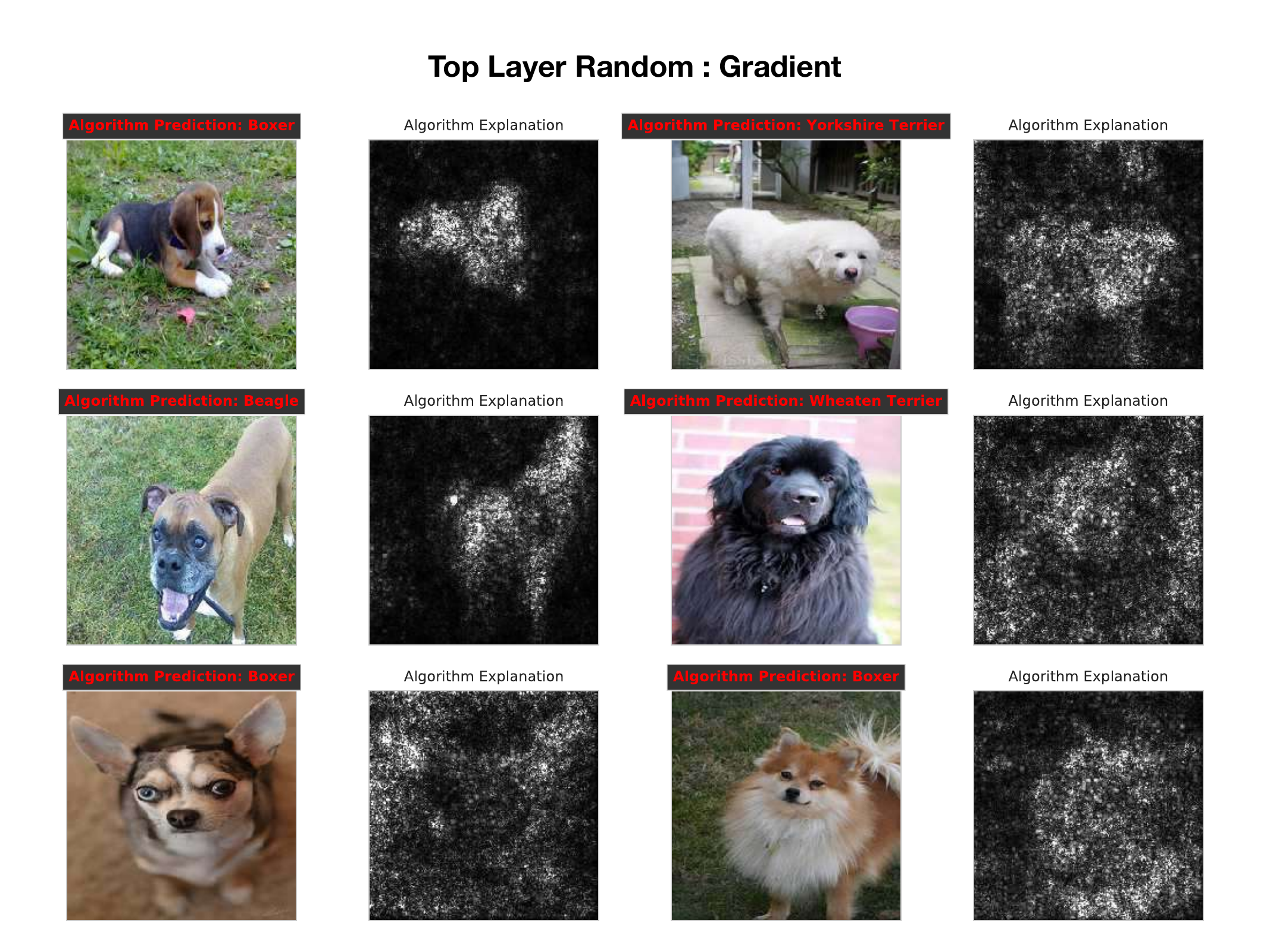}
\caption{Images used as part of the user study.}
\label{fig:userstudy4}
\end{figure*}

\begin{figure*}[!h]
\centering
\includegraphics[width=1.\textwidth,  page=69]{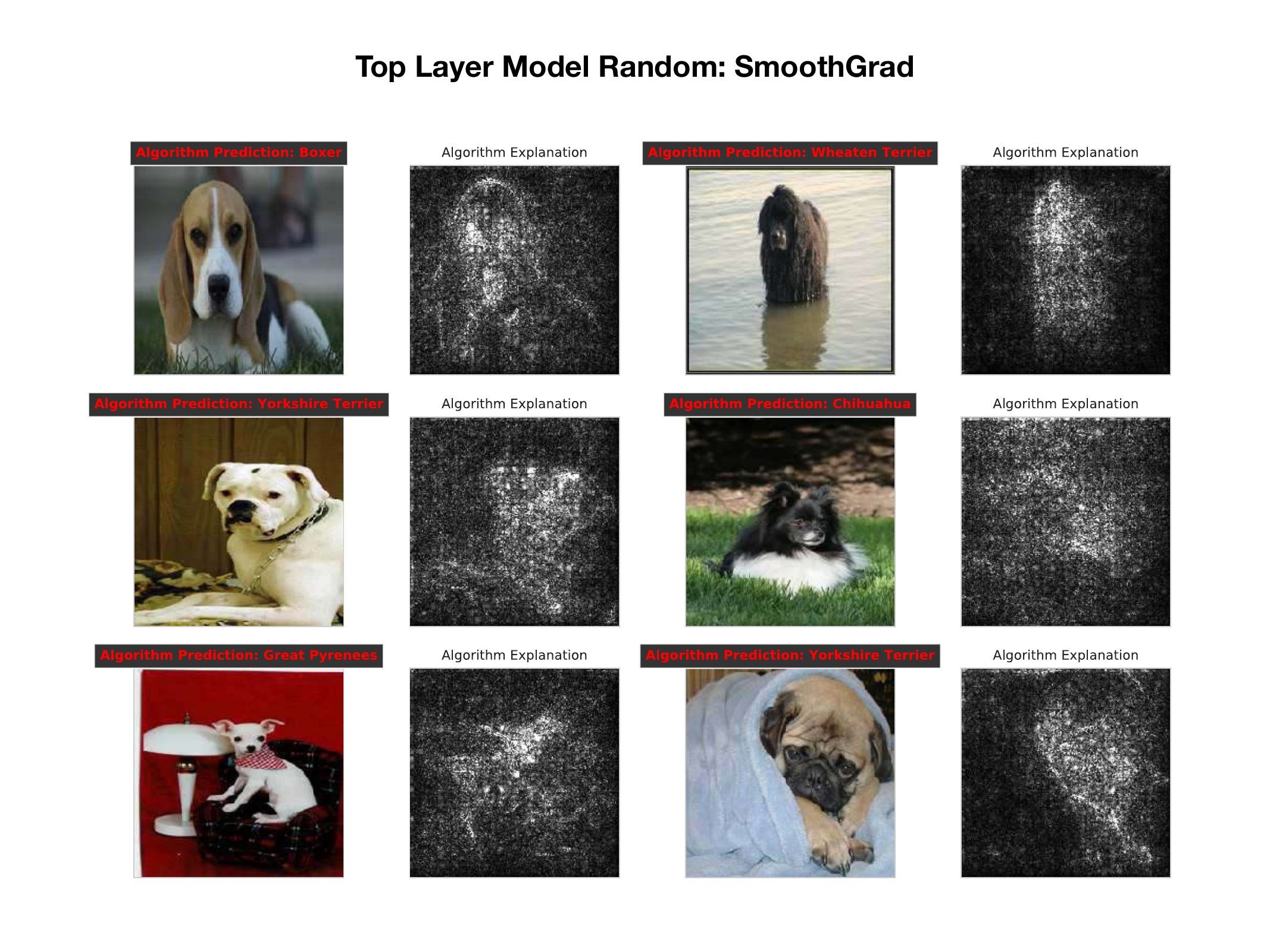}
\caption{Images used as part of the user study.}
\label{fig:userstudy5}
\end{figure*}

\begin{figure*}[!h]
\centering
\includegraphics[width=1.\textwidth,  page=70]{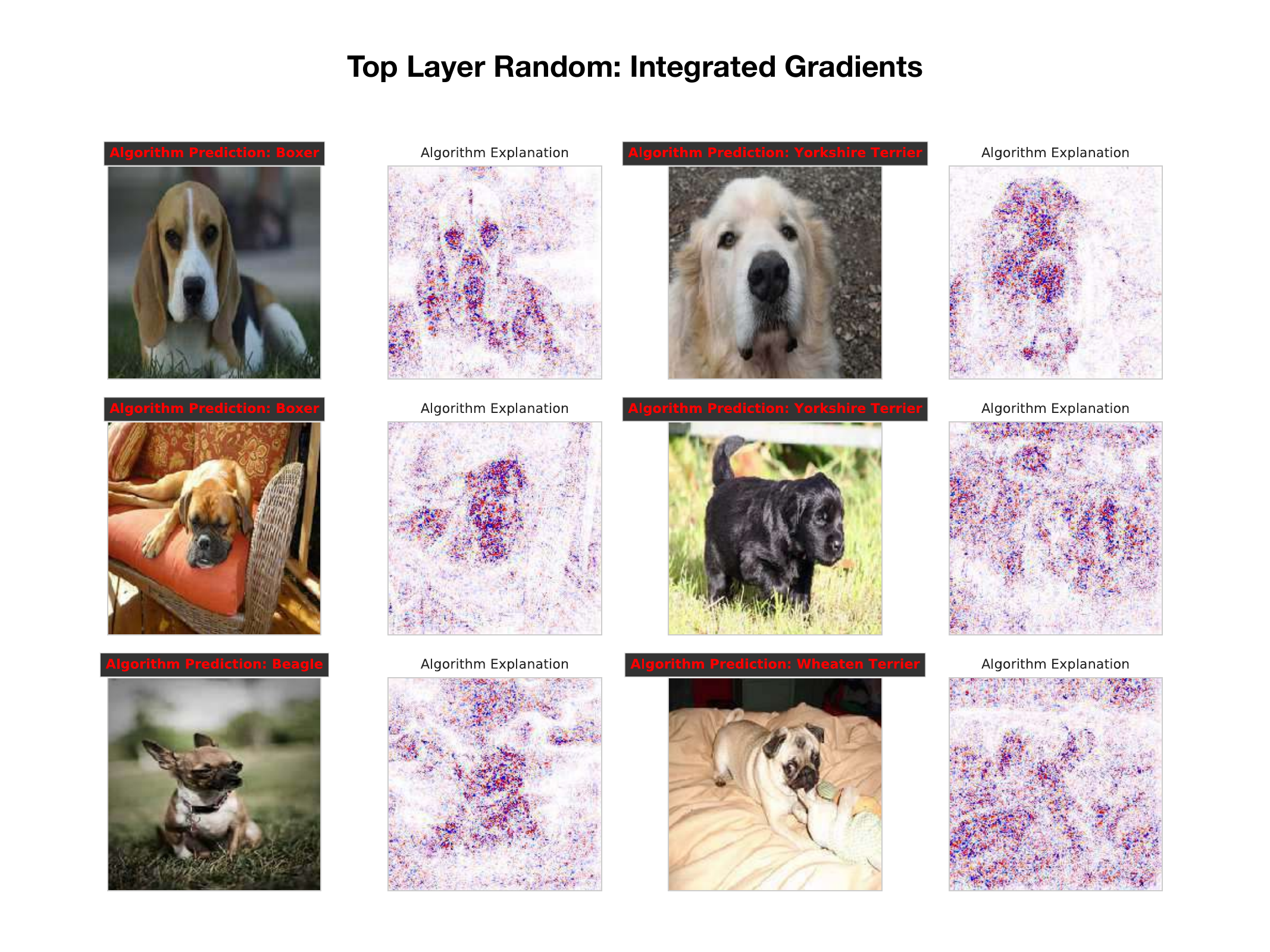}
\caption{Images used as part of the user study.}
\label{fig:userstudy6}
\end{figure*}

\begin{figure*}[!h]
\centering
\includegraphics[width=1.\textwidth,  page=74]{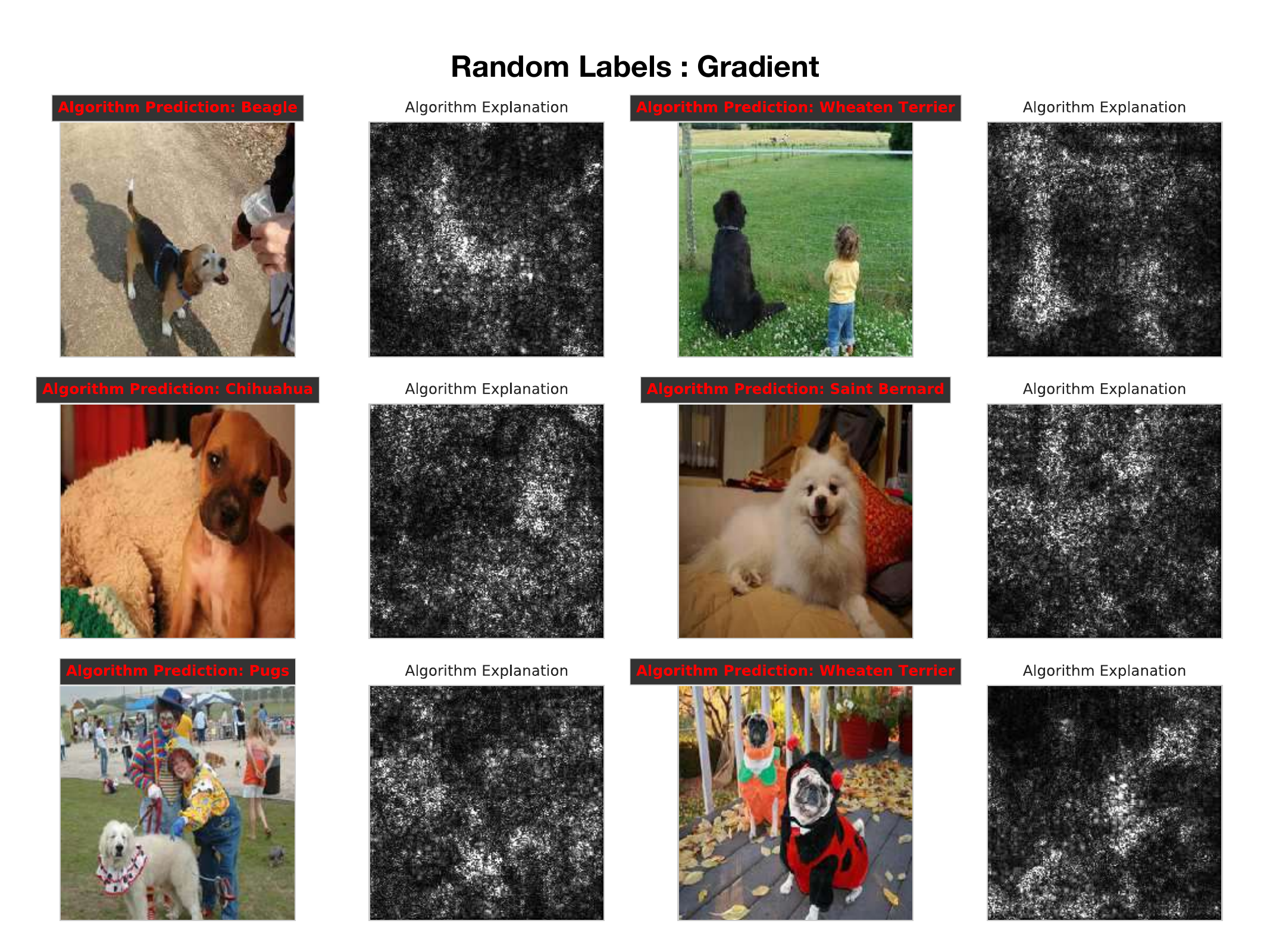}
\caption{Images used as part of the user study.}
\label{fig:userstudy7}
\end{figure*}

\begin{figure*}[!h]
\centering
\includegraphics[width=1.\textwidth,  page=75]{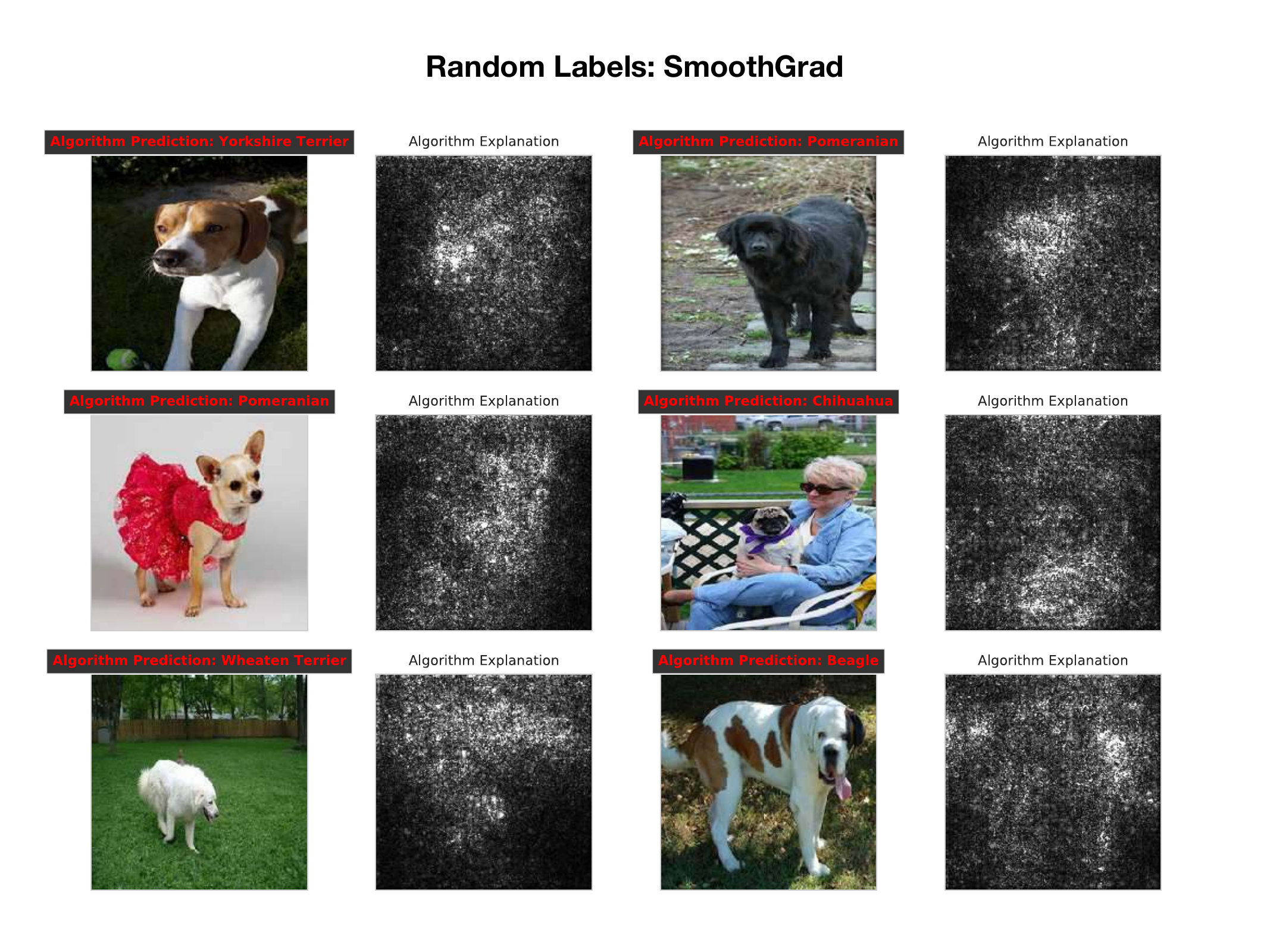}
\caption{Images used as part of the user study.}
\label{fig:userstudy8}
\end{figure*}

\begin{figure*}[!h]
\centering
\includegraphics[width=1.\textwidth,  page=76]{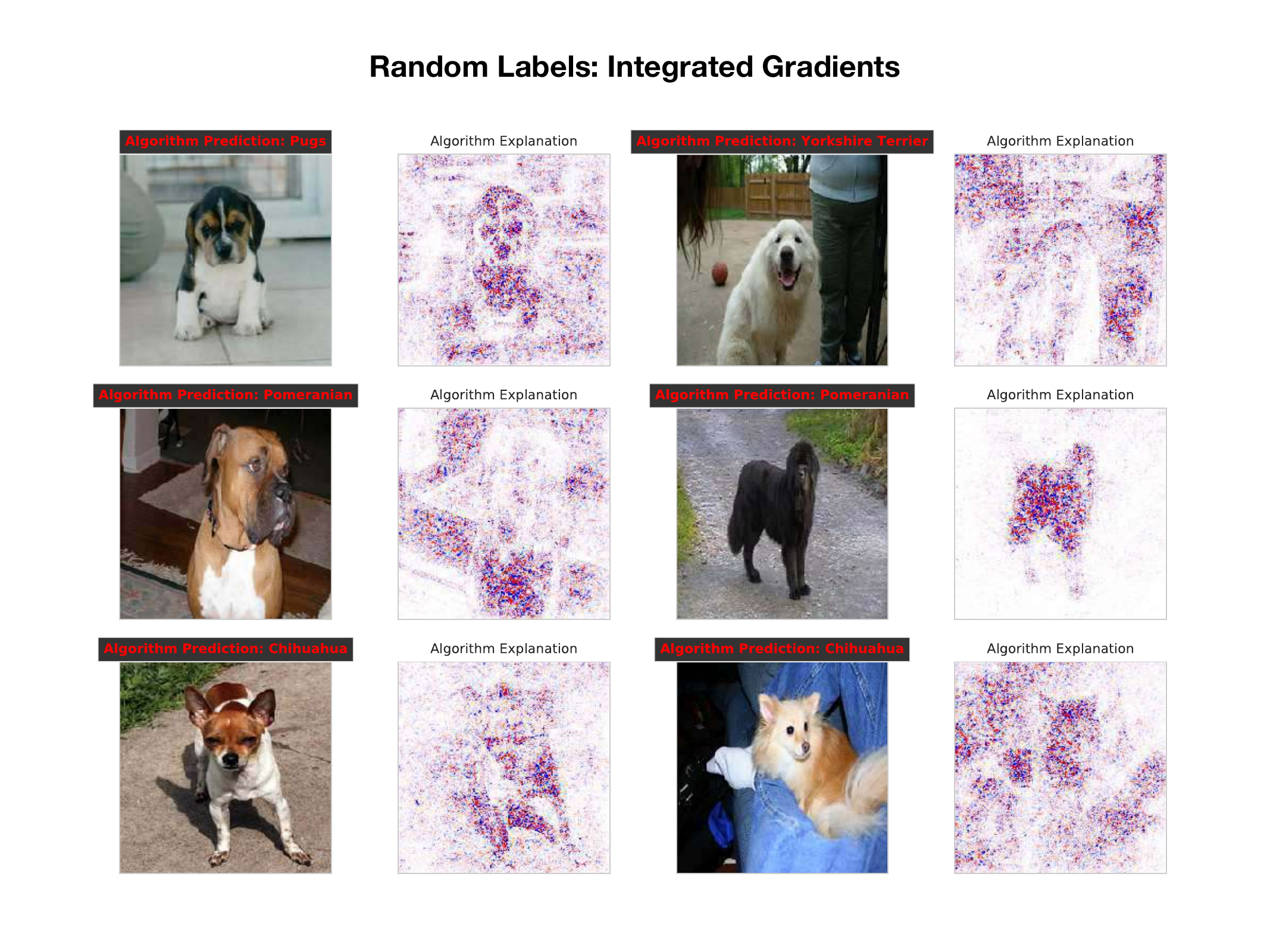}
\caption{Images used as part of the user study.}
\label{fig:userstudy9}
\end{figure*}

\begin{figure*}[!h]
\centering
\includegraphics[width=1.\textwidth,  page=77]{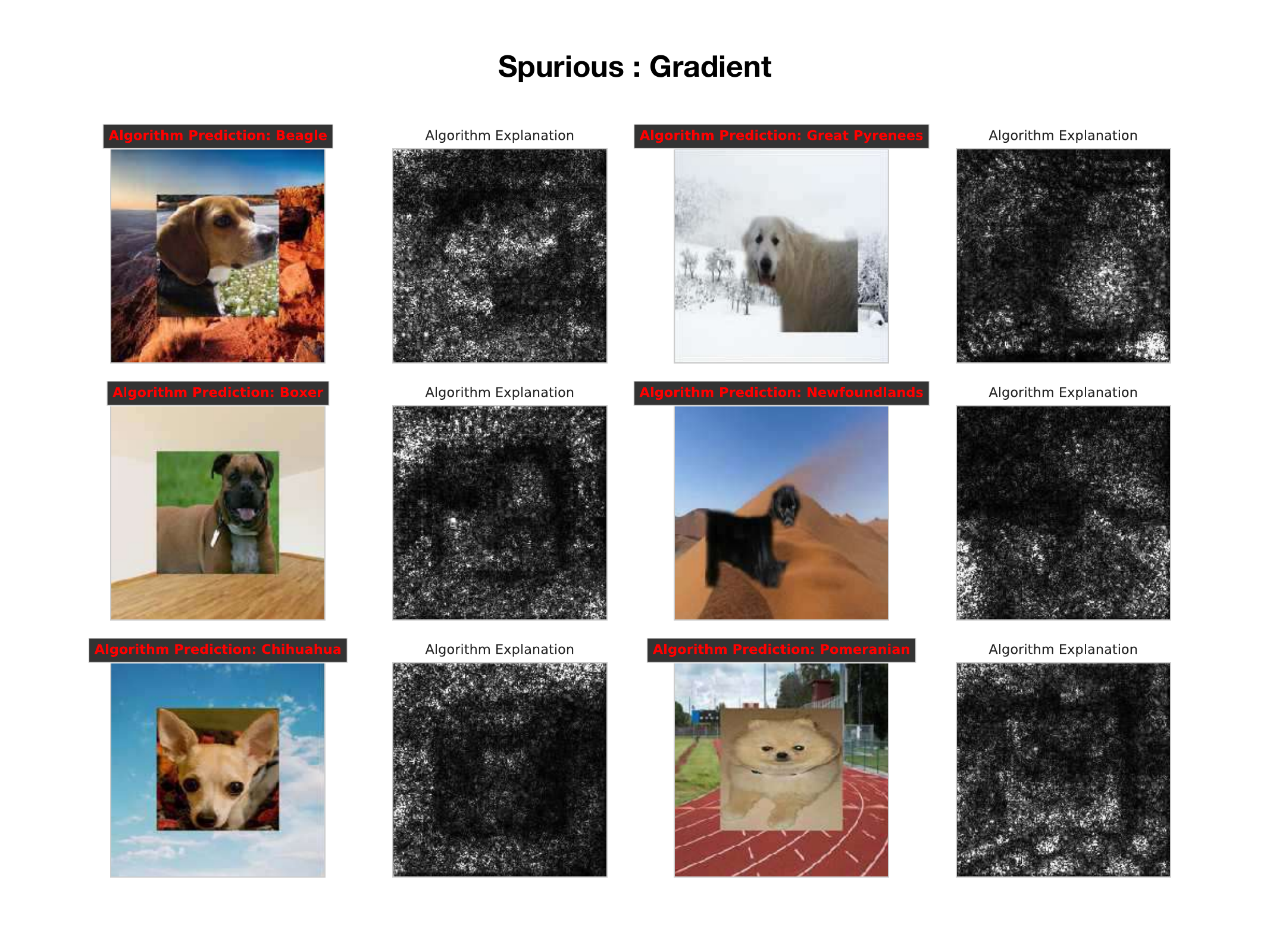}
\caption{Images used as part of the user study.}
\label{fig:userstudy10}
\end{figure*}

\begin{figure*}[!h]
\centering
\includegraphics[width=1.\textwidth,  page=78]{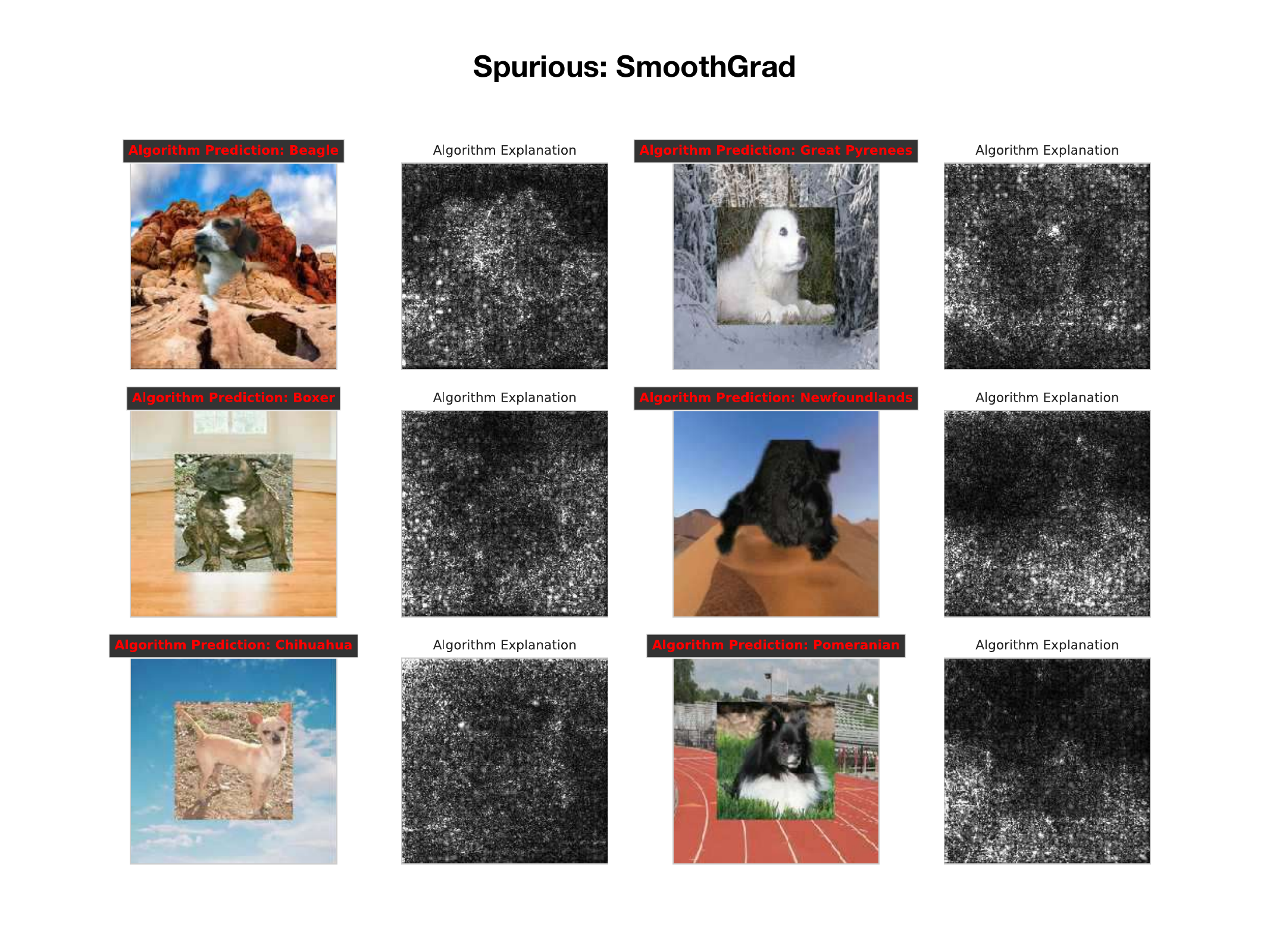}
\caption{Images used as part of the user study.}
\label{fig:userstudy11}
\end{figure*}

\begin{figure*}[!h]
\centering
\includegraphics[width=1.\textwidth,  page=79]{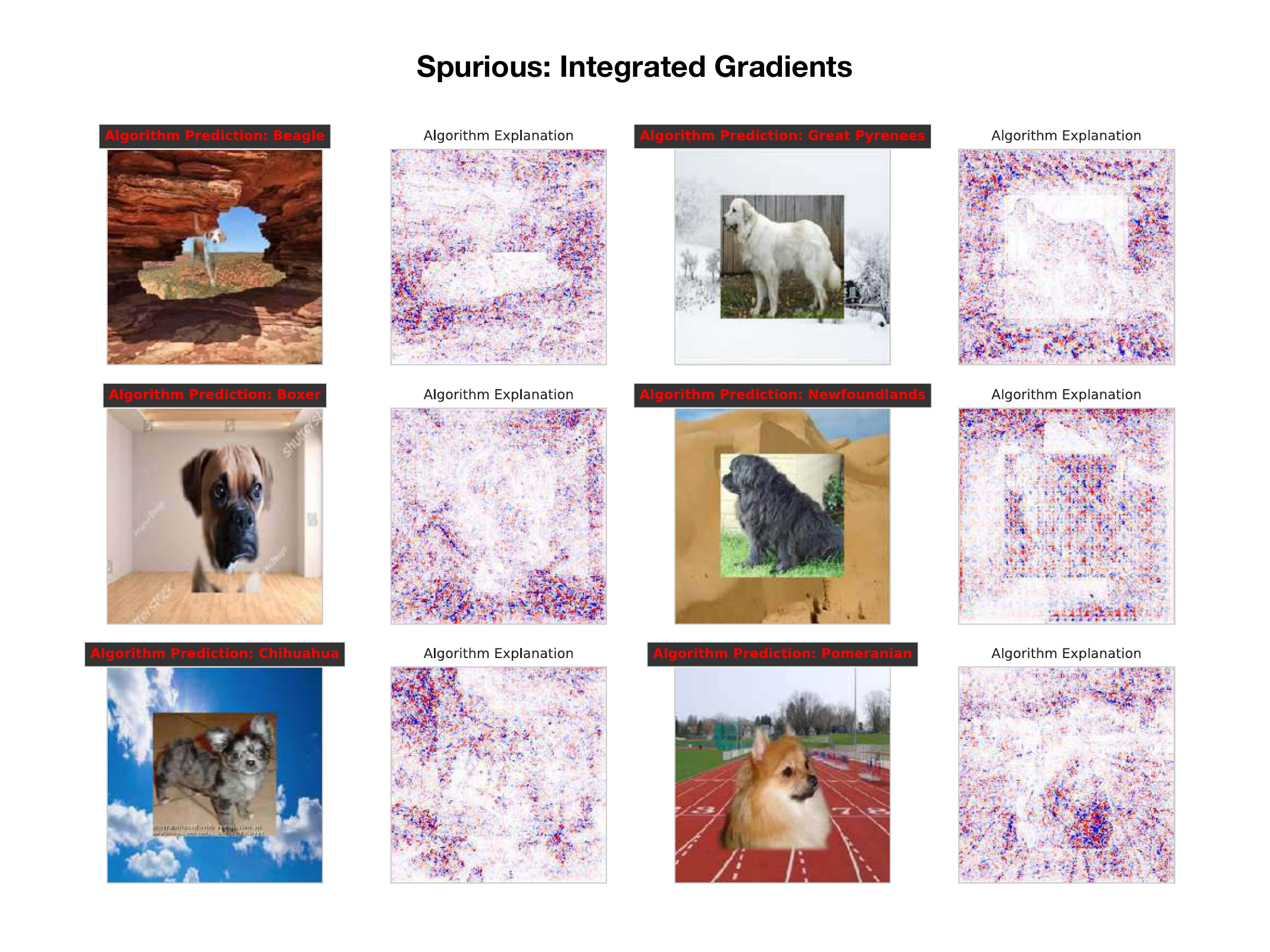}
\caption{Images used as part of the user study.}
\label{fig:userstudy12}
\end{figure*}

\begin{figure*}[!h]
\centering
\includegraphics[width=1.\textwidth,  page=83]{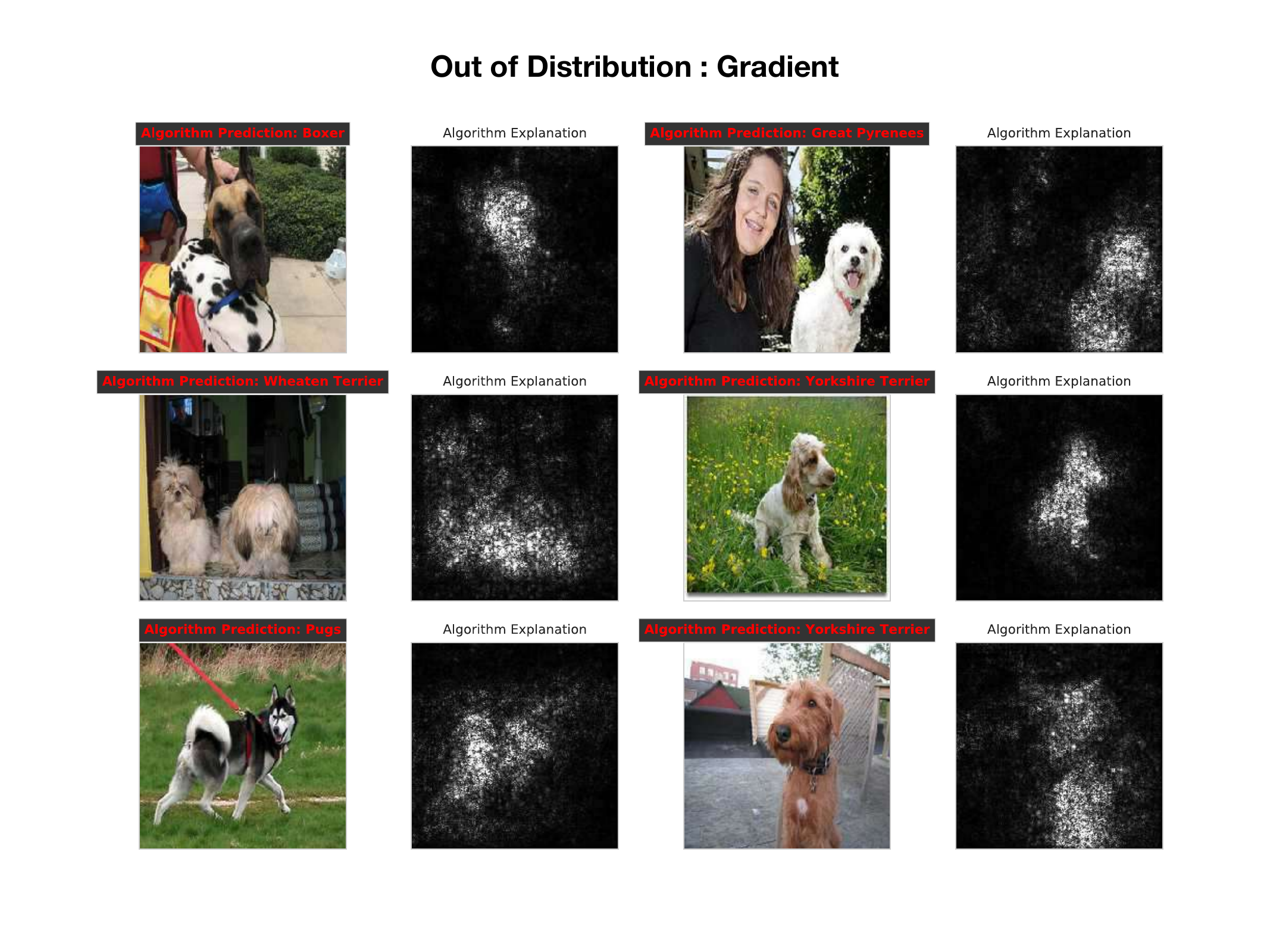}
\caption{Images used as part of the user study.}
\label{fig:userstudy13}
\end{figure*}

\begin{figure*}[!h]
\centering
\includegraphics[width=1.\textwidth,  page=84]{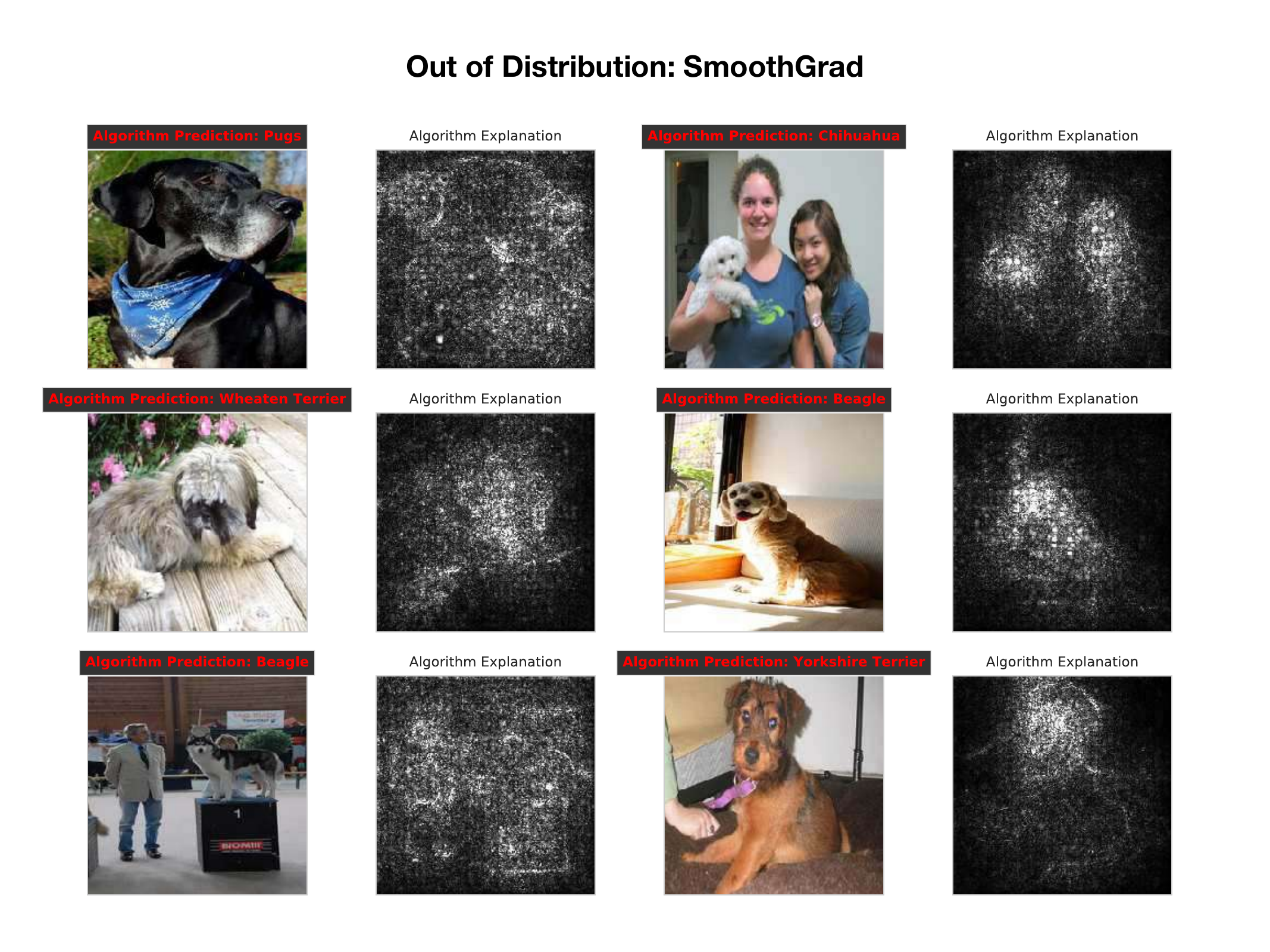}
\caption{Images used as part of the user study.}
\label{fig:userstudy14}
\end{figure*}

\begin{figure*}[!h]
\centering
\includegraphics[width=1.\textwidth,  page=85]{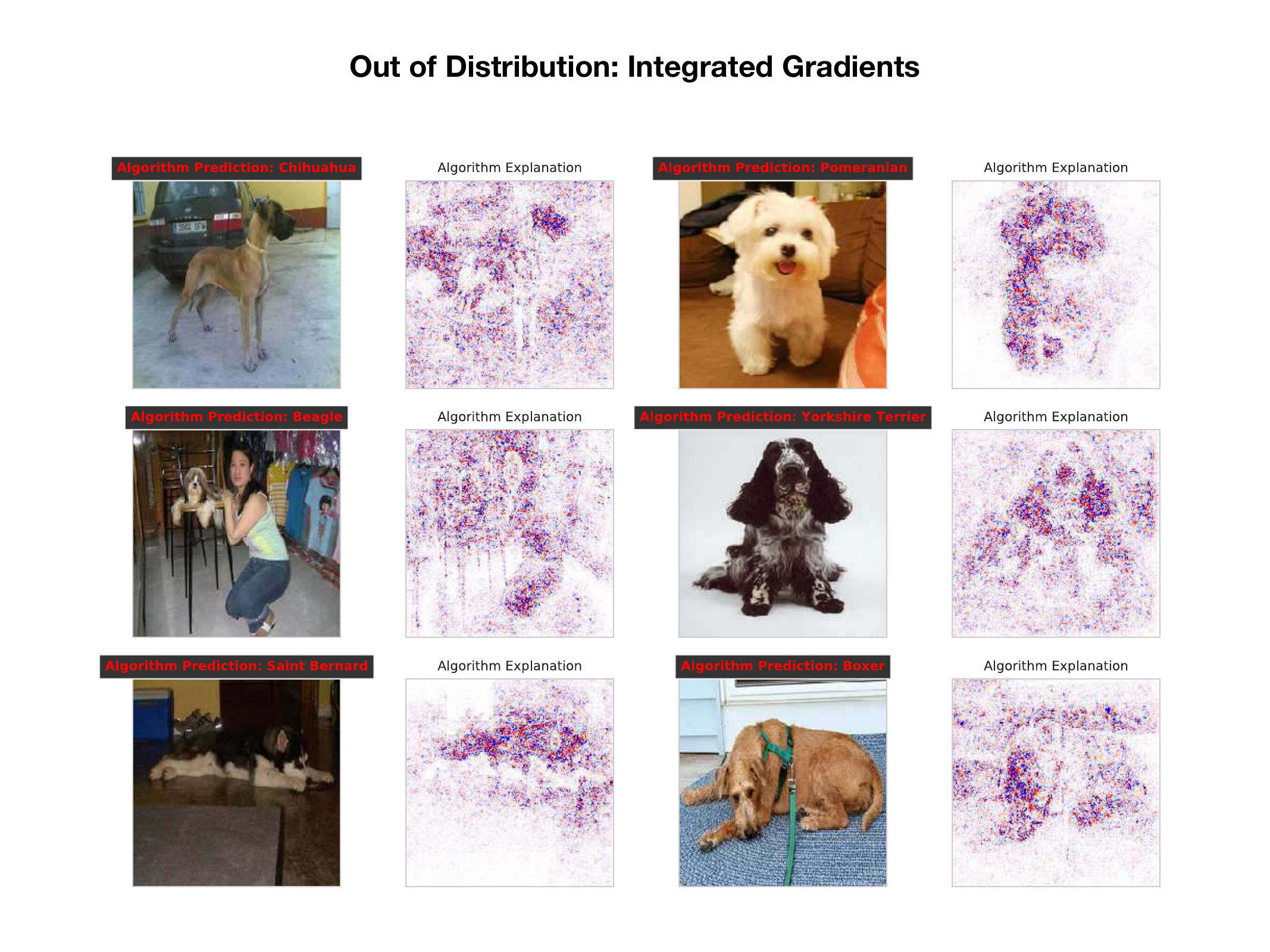}
\caption{Images used as part of the user study.}
\label{fig:userstudy15}
\end{figure*}
\end{document}